\documentclass[imslayout,preprint]{imsart} 
\def\journal@name{} 
\usepackage[margin=1.5in]{geometry}

\RequirePackage[breaklinks]{hyperref}
\RequirePackage{amsthm,amsmath,amsfonts,amssymb}
\RequirePackage[authoryear,sort]{natbib}%
\newcommand{\law}[1]{\mathfrak{L}_{#1}}
\newcommand{\R}{\mathbb{R}}
\newcommand{\on}[1]{\operatorname{#1}}

\newcommand{\samplesize}{n}
\newcommand{\iter}{k}
\newcommand{\swapiter}{K}

\newcommand{\exparm}{\eta'}

\newcommand{\testfun}{g}
\newcommand{\steintestfun}{{f}}

\newcommand{\trueparm}{{\theta_{\star}}}
\newcommand{\MLEparm}[1]{{\theta_{\star}\upper{#1}}}

\newcommand{\driftcoeff}{a}
\newcommand{\noisefun}{q}
\newcommand{\RiccatiSolution}{v}

\newcommand{\skspace}{D([0,1])}
\newcommand{\weinerprocess}{W}

\newcommand{\skelement}{u}
\newcommand{\skothelement}{v}

\newcommand{\exmatrix}[1]{\Lambda_{#1}}

\newcommand{\error}[1]{{\epsilon_{\mathrm{#1}}}}

\newcommand{\idbatch}[2]{I(#1,#2)}
\newcommand{\stepsize}{h}
\newcommand{\batchsize}{b}
\newcommand{\invtemp}{\beta}
\newcommand{\timescale}{\alpha}
\newcommand{\spacescale}{w}
\newcommand{\noise}{\xi}
\newcommand{\Gradbatch}[2]{\psi_{\idbatch{#1}{#2}}}
\newcommand{\Heisbatch}[2]{\sigma_{\idbatch{#1}{#2}}}
\newcommand{\rmdbatch}[2]{R_{\idbatch{#1}{#2}}}
\newcommand{\swapidbatch}[2]{I'(#1,#2)}
\newcommand{\swapGradbatch}[2]{\psi_{\swapidbatch{#1}{#2}}}
\newcommand{\swapHeisbatch}[2]{\sigma_{\swapidbatch{#1}{#2}}}
\newcommand{\newy}{\mathcal{Y}}
\newcommand{\newz}{\mathcal{Z}}
\newcommand{\exy}{\mathcal{Y}'}
\newcommand{\newyt}{\mathcal{Y}_t}
\newcommand{\newzt}{\mathcal{Z}_t}
\newcommand{\exyt}{\mathcal{Y}'_t}
\newcommand{\pmt}{\theta}
\newcommand{\newpmt}{\eta}

\newcommand{\pmtn}{\pmt^{(n)}}
\newcommand{\stepsizen}{h^{(n)}}
\newcommand{\batchsizen}{b^{(n)}}
\newcommand{\invtempn}{\beta^{(n)}}
\newcommand{\timescalen}{\alpha^{(n)}}
\newcommand{\spacescalen}{w^{(n)}}

\usepackage[cleverefIMSART,commenters={XW,MK,JN,JH,SB}]{jnmaxpax}
\usepackage{mathrsfs}
\usepackage{autonum} 
\usepackage{mathtools}
\usepackage[shortlabels]{enumitem}

\mathtoolsset{centercolon}

\AtBeginDocument{ %
\def\[#1\]{\begin{align}#1\end{align}}
\def\+[#1\+]{\begin{equation+}\begin{aligned}#1\end{aligned}\end{equation+}}
}

\usepackage{chngcntr}

\allowdisplaybreaks[3]

\begin{document}

\begin{frontmatter}
\title{Quantitative Error Bounds for Scaling Limits of Stochastic Iterative Algorithms}
\runtitle{~Quantitative Scaling Limits of Stochastic Iterative Algorithms}
\runauthor{Wang et al.~}

\begin{aug}
\author[MS]{\fnms{Xiaoyu}~\snm{Wang}\ead[label=xw]{shawnwxy@bu.edu}},
\author[X]{\fnms{Miko{\l}aj J.}~\snm{Kasprzak}\ead[label=mk]{kasprzak@essec.edu}},
\author[UW]{\fnms{Jeffrey}~\snm{Negrea}\ead[label=jn]{jnegrea@uwaterloo.ca}},
\author[MS]{\fnms{Solesne}~\snm{Bourguin}\ead[label=sb]{bourguin@bu.edu}},
\and
\author[MS,CDS]{\fnms{Jonathan H.}~\snm{Huggins}\ead[label=jh]{huggins@bu.edu}}

\address[MS]{Department of Mathematics \& Statistics, Boston University\printead[presep={,\ }]{xw,sb,jh}}

\address[X]{Department of Information Systems, Data Analytics \& Operations, ESSEC Business School\printead[presep={,\ }]{mk}}
\address[UW]{Department of Statistics \& Actuarial Science, University of Waterloo\printead[presep={,\ }]{jn}}
\address[CDS]{Faculty of Computing \& Data Sciences, Boston University}
\end{aug}

\begin{abstract}
Stochastic iterative algorithms, including stochastic gradient descent (SGD) and stochastic gradient Langevin dynamics (SGLD), are widely utilized for optimization and sampling in large-scale and high-dimensional problems
in machine learning, statistics, and engineering. 
Numerous works have bounded the parameter error in, and characterized the uncertainty of, these approximations. 
One common approach has been to use scaling limit analyses to relate the distribution of algorithm sample paths to a continuous-time stochastic process approximation, particularly in asymptotic setups. 
Focusing on the univariate setting, in this paper, we build on previous work to derive non-asymptotic functional approximation error bounds between the algorithm sample paths and the Ornstein--Uhlenbeck approximation using an infinite-dimensional version of Stein's method of exchangeable pairs. 
We show that this bound implies weak convergence under modest additional assumptions and leads to a bound on the error  
of the variance of the iterate averages of the algorithm. 
Furthermore, we use our main result to construct error bounds in terms of two common metrics: the L\'evy-Prokhorov and bounded Wasserstein distances. 
Our results provide a foundation for developing similar error bounds for the multivariate setting and for more sophisticated stochastic approximation algorithms. 
\end{abstract}



\end{frontmatter}

\section{Introduction}

We often wish to find an (approximate) solution to a deterministic equation $\gradloss(\pmt) = 0$, with one notable special case being to find an (approximate) minimum of a function $\loss$ with gradient $\gradloss =\grad\loss$.
Iterative stochastic approximation methods are employed in a wide variety of disciplines to solve such problems in cases where it is convenient to use random, unbiased estimates of $\gradloss(\pmt)$ \citep{Robbins1951,pdc1997,lai2003}. 
For example, these methods are applicable when it is only feasible to compute this approximation and scale well to high-dimensional problems, notably the training of deep neural networks \citep{NIPS2007_bottou,Goodfellow-et-al-2016,Li_Chen_Carlson_Carin_2016}.

A canonical example of stochastic approximation is provided by stochastic gradient descent \citep[SGD;][]{Nemirovski2009,NIPS2011_40008b9a,bottou2016} for optimizing functions with the form of a finite sum 
$
\loss(\pmt) = \samplesize^{-1} \sum_{i=1}^\samplesize \obsloss{i}(\pmt),
$
where $\pmt \in \R^d$ and $\obsloss{i}$ is a loss function, typically associated with the $i$th observation in a dataset. 
Given a batch size $\batchsize \in \N_+$ and independent, uniformly distributed random
variables $I(k, i) \in \{1,\dots,\samplesize\}$ (for $\iter \in \N, i \in \{1,\dots,\batchsize\}$), an unbiased approximation to $\loss(\pmt)$ is given by 
\[
\stochloss{\iter}(\pmt) \defas \batchsize^{-1} \sum_{i=1}^\batchsize \obsloss{I(\iter, i)}(\pmt),
\]
with corresponding unbiased gradient approximation $\gradstochloss{\iter} \defas \grad\stochloss{\iter}$. 
The SGD algorithm constructs a sequence of estimates $\pmt_1, \pmt_2, \dots$
of a local critical point $\mle$ that satisfies $\gradloss(\mle) = 0$.
Given an initial value $\pmt_0$ and step size sequence $\stepsize_1, \stepsize_2, \dots$, the estimates are defined recursively by the update equation 
\[ \label{eq:SGD}
\pmt_{\iter+1} = \pmt_{\iter} + \stepsize_\iter \gradstochloss{\iter}(\pmt_\iter).
\]

More recently, SGD and variants such as stochastic gradient Langevin dynamics (SGLD) have been used in machine learning for approximate sampling to quantify uncertainty about model parameters \citep{conf/icml/WellingT11,sgmcmc2021}. 
SGLD adds isotropic Gaussian noise to each stochastic gradient step, where this noise is sometimes 
preconditioned by a positive-definite matrix. 
Given inverse temperature parameter $\invtemp \in (0,\infty]$ (with $\infty^{-1} \defas 0$) and independent standard Gaussian noise $\noise_\iter \distas N(0,I)$, 
the SGLD update equation is 
\[ \label{eq:SGLD}
\pmt_{\iter+1} = \pmt_{\iter} + \stepsize_\iter \gradstochloss{\iter}(\pmt_\iter) + \sqrt{\frac{2\stepsize_\iter}{\invtemp}}\noise_{\iter}.
\]
The canonical choice for the temperature is $\invtemp = \samplesize$;
however, other choices have been explored \citep{Wenzel:2020:DNN-Bayes,Raginsky:2017}.
Since we can recover \cref{eq:SGD} from \cref{eq:SGLD} by taking 
$\invtemp = \infty$, we will focus on the more general case, which we refer to 
as SG(L)D to remind the reader both SGD and SGLD are included.

\subsection{Related Work}

Given its widespread use, SG(L)D has been studied from numerous perspectives, including bounding
the parameter error, characterizing its stationary distribution, studying its convergence properties, and its escape time from local minima \citep[e.g.,][]{Mcleish1976,ruppert1988,doi:10.1137/0330046,kushner2003stochastic,NIPS2011_40008b9a,negrea2022statistical,rakhlin2011making,10.1214/19-AOS1850,pmlr-v125-mou20a,cheng2020stochastic,srikant2024rates,pmlr-v99-anastasiou19a,ge2015escaping,jin2017escape}.
One fruitful approach to understanding SG(L)D has been using scaling limit analyses, which relate SG(L)D sample paths to a continuous-time stochastic process. 
Given a spatial scaling $\spacescale$ and temporal scaling $\timescale$,
define the rescaled, continuous-time SG(L)D process $Y = (Y_t)_{t \in [0,1]}$ by
\[ \label{eq:SGLD-process}
Y_t \defas \spacescale\rbra*{\pmt_{\floor{\timescale t}}-\mle}. %
\]
We focus on the fixed step size case where $\stepsize_k = \stepsize$ for all $k \in \N$, since it is practically relevant to both optimization and sampling \citep{10.1214/19-AOS1850,NIPS2011_40008b9a,conf/icml/WellingT11,sgmcmc2021,baker2019,pmlr-v48-hardt16}.
With appropriate choices of $\spacescale = \spacescale(\stepsize)$ and $\timescale = \timescale(\stepsize)$,
it can be shown that $Y$ converges in distribution to
an Ornstein--Uhlenbeck (OU) process when $\stepsize \to 0$. 
For example, \citet{kushner2003stochastic} show that if
($i$) $\spacescale = \sqrt{\stepsize}$ and $\timescale=\stepsize$, 
($ii$) the Jacobian
$B \defas \grad \gradloss(\mle)$ exists, and ($iii$) the limit 
\[
A \defas \lim_{\stepsize \to 0} h^2 \, \EE\cbra*{\rbra*{\gradstochloss{\iter}(\pmt_\iter)+\sqrt{\frac{2}{h\invtemp}}\noise_k - \gradloss(\pmt_\iter)}^{\otimes 2}}
\] 
exists, then as $\stepsize \to 0$, $Y$ converges weakly to the OU process $Z = (Z_t)_{t \in [0,1]}$ that satisfies the SDE 
\[
    \d Z_t
       & = -B Z_t \,\d t + \sqrt{A}\,\d W_{t}, 
\]
where $(W_t)_{t \in [0,1]}$ is a standard Brownian motion. 
Taking a statistical perspective, \citet{negrea2022statistical} instead 
consider a sequence of random loss functions $\loss = \loss^{(\samplesize)}$ that depend on the the sample size $n$ and allow  $\stepsize$, $\batchsize$, $\invtemp$, $\spacescale$, and $\timescale$ to depend on $\samplesize$; they prove a similar type of result when $\stepsize \to 0$ as $\samplesize \to \infty.$

Characterization of the sample-path distribution facilitate analysis of the asymptotic accuracy of averaged SGD iterates \citep{kushner2003stochastic,negrea2022statistical}, which is introduced by \citet{doi:10.1137/0330046}, and is widely used in practice \citep{laszlo1996,bach2013nonstronglyconvexsmoothstochasticapproximation,pmlr-v75-tripuraneni18a}.
Furthermore, scaling limits enable us to evaluate the efficiency of using SG(L)D as a Markov chain Monte Carlo (MCMC) sampler by analyzing its mixing time.
In addition, they can be used to determine the asymptotic stationary distribution of the sampler and establish Bernstein-von Mises-type theorems for the stationary distribution \citep{kushner2003stochastic,mandt2017stochastic,negrea2022statistical}.

\subsection{Our Contribution}
Despite the value of existing scaling limit results, they are all asymptotic in nature, which leaves it unclear in what regimes the results hold in practice.\footnote{A partial exception is \citet{pmlr-v99-anastasiou19a}, which gives error bounds between the marginal distributions 
of the iterates and a Gaussian approximation. However, they do not study path-wise convergence.}
We begin to fill this gap by using functional Stein's method,
which was first proposed by \citet{barbour1990diffusion} and later extended and further developed by several authors, including \citet{kasprzak2020brownian,decreusefond2016functional,coutin2020donsker,barbour2023,besancon2024diffusive,balasubramanian2024}. 
The exchangeable-pair approach to Stein's method was introduced by \citet{stein1986approximate}, extended to the multivariate context by \citet{chatterjee2008multivariate, meckes2009multivariate, reinert2010multivariate}, 
and finally combined with the functional version of Stein's method in \citet{kasprzak2020exchangeable,dobler2021stein}. 
So far, the functional exchangeable-pair method has been successfully applied to a limited array of specific models, including weighted U-statistics and a certain graph-valued stochastic process \citep{dobler2021stein}, as well as a functional combinatorial central limit theorem \citep{kasprzak2020exchangeable}. As far as the authors are aware, \textit{all previous works on the functional Stein's method of exchangeable pairs have concentrated on its applicability to target processes which have independent increments}. 
While \cite{barbour2023} have successfully studied processes with non-independent increments using Stein's method, their focus was on the cases in which the approximated process takes the form of an integral with respect to a point process. 
Their are not easily applicable to the setup of the present paper they rely on tools specific for point processes, such as Palm measures, and do not use exchangeable pairs.
We substantially extend the applicability of the functional Stein's method of exchangeable pairs by using it to target a discretized Ornstein-Uhlenbeck process. \textit{We thus illustrate how functional Stein's method of exchangeable pairs can be used to upper-bound a distance from a process which does not have independent increments}. 
By combining it with judiciously chosen Taylor expansions, we also demonstrate that it is well-suited to the analysis of processes with complex covariance structures, like SG(L)D.

Specifically, we establish bounds for the approximation error between the rescaled SG(L)D process $Y = (Y_t)_{t \in [0,1]}$ and the limiting OU process $Z = (Z_t)_{t \in [0,1]}$, in the univariate case -- i.e., when the processes $Z$ and $Y$ are real-valued. For a certain collection of test functions $g$, we establish a uniform bound on $\abs{\EE g(Z)- \EE g(Y)}$.
To achieve this, we construct a new scaled SG(L)D process $\newy$ based on a linear approximation to $\grad \obsloss{i}(\pmt)$ and compare it with a discretized OU process $\newz$. We decompose the original error into three components: the error between $\newy$ and $\newz$, the linear approximation error, and the discretization error, with the latter two being straightforward to bound. To bound the first error, we construct an exchangeable pair $(\newy, \exy)$ by randomly replacing the batch in one of the SG(L)D stochastic gradients. This approach enables us to apply the abstract approximation theorem of \citet{dobler2021stein}. We comment in \cref{rmk:multivariate-extension} on a possible extension of our bound to the multivariate case.

\subsection{Paper Structure}
After describing our approach in detail in \cref{sec:approach}, 
in \cref{sec:main}, we present our main result under two setups that cover common use cases. 
The resulting error bounds depend on the step size $\stepsize$ and various algorithm tuning 
parameters such as batch size. 
Our most general error bound, \cref{MainThm} in the supplementary material \citep{supp2025quantscalinglimit}, of which the bounds in \cref{sec:main} are corollaries, is given in \cref{app:additional proofs} of the supplementary material.

In \cref{sec:boundforouprocess}, we provide a quantitative bound on the difference between two OU processes, which we use to obtain a quantitative version of the statistical scaling limit from \citet{negrea2022statistical}.
Our OU error bound is based on a novel $L^p$ maximal inequality for OU processes that may be of independent interest.  

We present a number of implications of our main results in \cref{sec:implications}. 
First, we show how to use our main result to prove the weak convergence of $Y$ to $Z$. 
Next, we provide an error bound on the uncertainty quantification of iterate averages of SG(L)D. 
Finally, we use our main result together with recent results from \citet{barbour2024stein} to bound the distance between SG(L)D and its scaling limit with respect to two standard probability metrics, the Levy--Prokhorov distance and bounded Wasserstein distance. 

\Cref{sec:main-theorem-proof} contains a simplified version of our most general result (\cref{MainThm} in the supplementary material), together with a sketch proof. Full technical proofs of all the main results are postponed to the appendix, included as supplementary material.

\section{Approach}\label{sec:approach}

We bound the difference between an appropriately chosen OU process
and the iterates of the SG(L)D algorithm with constant step size $\stepsize_\iter = \stepsize$.
Following standard practice \citep{barbour1990diffusion,kasprzak2020brownian}, we compare the centered, continuous-time interpolation $Y \defas (Y_t)_{t \in [0,1]}$ defined in \cref{eq:SGLD-process} to the OU process $Z \defas (Z_t)_{t \in [0,1]}$ satisfying
the SDE
\[\label{formulaz}
    \begin{cases}
        \d Z_t = -\timescale\stepsize\FI Z_t \,\d t + w\sqrt{\frac{\alpha h^2}{b} \Omega  + \frac{ 2 \alpha h}{\invtemp}I}\, \d\weinerprocess_{t} \\
        Z_0 = Y_0
    \end{cases},
\]
where $\FI \defas -\grad^{\otimes 2}\loss(\mle)$, $\foFI \defas \samplesize^{-1}\sum_{i=1}^{\samplesize} \grad \obsloss{i}(\mle)^{\otimes 2}$ and $W$ is a Wiener process.

\subsection{Functional Method of Exchangeable Pairs}\label{sec:exchangeable_pairs}
To quantify the difference between $Y$ and $Z$, we want to bound $|\EE\{\testfun(Y)\} -\EE\{\testfun(Z)\}|$ for a possibly large class of real-valued functionals $\testfun$ acting on the {Skorokhod space} $\skspace$ of all c\`adl\`ag functions on $[0,1]$. 
Our approach is to use a functional version of Stein's method \citep{barbour1990diffusion,dobler2021stein}.
So that we can define the relevant class of functionals we consider, denote the Frech\'{e}t derivative by $\FD$ and the Euclidean norm by $\abs{\cdot}$.
And, for $\skelement \in \skspace$ and $\testfun \colon \skspace \to \R$, define the norms 
\[
 \norm{\skelement}
    & \defas\sup_{t\in[0,1]}\abs{\skelement(t)}, \\
\norm{\FD\testfun(\skelement)}
    &\defas\sup_{\norm{\skothelement}\leq 1} \abs{\FD\testfun(\skelement)[\skothelement]}, \\
    \ \andT \\
\norm{\FD^2\testfun(\skelement)}
    &\defas\sup_{\substack{\norm{\skothelement_1}\leq 1\\ \norm{\skothelement_2}\leq 1}}\abs{\FD^2\testfun(\skelement)[\skothelement_1,\skothelement_2]},
\]
where $\skothelement, \skothelement_1, \skothelement_2 \in \skspace$.
Following \citet{barbour1990diffusion}, we concentrate on the class of test functions 
$M = \{\testfun \colon \skspace\to\R\,:\,\|\testfun\|_M<\infty\}$, where
\[
\norm{\testfun}_M
    & \defas
        \sup_{\skelement\in \skspace}\frac{\norm{\testfun(\skelement)}}{1+\norm{\skelement}^3}
        +\sup_{\skelement\in \skspace}\frac{\norm{\FD\testfun(\skelement)}}{1+\norm{\skelement}^2} \\
    &\qquad
        +\sup_{\skelement\in \skspace}\frac{\norm{\FD^2 \testfun(\skelement)}}{1+\norm{w}}
        +\sup_{\skelement,\skothelement\in \skspace}\frac{\norm{\FD^2 \testfun(\skelement+\skothelement)-\FD^2 \testfun(\skelement)}}{\norm{\skothelement}}.\label{m_space}
\]

In this work we consider the one-dimensional case $d = 1$;
we discuss extending our analysis to $d > 1$ in \cref{rmk:multivariate-extension} below. 
In the univariate case, we can use the following functional version of the method of exchangeable pairs.
We recall that $(\newy, \newy')$ is an \emph{exchangeable pair} of $\skspace$-valued random variables if $(\newy, \newy')$ and $(\newy', \newy)$ are equal in distribution.
\begin{proposition}[{\citet[Theorem 4.1 and Proposition 3.2]{dobler2021stein}}] \label{ThmM}
    Assume that $(\newy, \newy')$ is an exchangeable pair of $\skspace$-valued random variables, fix some $\exmatrix{} > 0$, and, 
    for all $f \in M$, let the random variable $R_f = R_f(\newy)$ be defined to satisfy
    \[ \label{eq:regression-eqn-for-exch-pair}
    \FD f(\newy)[\newy]= 2\,\cEE{\newy}\cbra*{\FD f(\newy)[(\newy - \newy')\exmatrix{}]} + R_f,
    \]
    where $\cEE{\newy}\defas \EE\cbra*{\cdot \mid \newy}$.
    Let $(z_1,\dots,z_{\samplesize})$ be a Gaussian vector with a positive definite covariance and, for $t \in [0,1]$, let $\newz_t \defas \sum_{i=1}^{\lfloor\samplesize t\rfloor}z_i$.
    For the process $\newz \defas (\newz_t)_{t\in [0,1]}$
    and any $g\in M$, set $f_g(w) \defas \int_0^{\infty}\EE[g(we^{-u}+\sqrt{1-e^{-2u}}\newz)]\d u$. Then we have the bound
    \[
    \abs{\EE[g(\newy)] - \EE[g(\newz)]} \le \error{exch} + \error{cov} + \error{rem},
    \]
    where, for $f = f_g$,
    \[
    \error{exch} &= \frac{\|g\|_M}{6}\EE\left\{\|(\newy - \newy')\exmatrix{}\|\,\|\newy - \newy'\|^2\right\}, \\
    \error{cov} &=\left|\EE\cbra*{\FD^2 f(\newy)[(\newy - \newy')\exmatrix{},\newy - \newy']} - \EE\cbra*{\FD^2f(\newy)[\newz,\newz]}\right|, \\
        \andT\\
    \error{rem} &=|\EE(R_f)|.
    \] 
    Moreover, the following bounds on $f_g$ hold for all $w,h\in\skspace$:
\[
    &\|\FD f_g(w)\|\leq \|g\|_M\left(1+\frac{2}{3}\|w\|^2+\frac{4}{3}\EE\|\newz_t\|^2\right), \\
    &\|\FD^2 f_g(w)\|\leq \|g\|_M\left(\frac{1}{2}+\frac{1}{3}\|w\|+\frac{1}{3}\EE\|\newz_t\|^2\right),\\
       \andT\\
    &\frac{\|\FD^2 f_g(w+h)-\FD^2f_g(w)\|}{\|h\|}\leq \|g\|_M\left(\frac{1}{2}+\frac{1}{3}\|w\|+\frac{1}{3}\EE\|\newz\|^2\right).\label{f_properties}
\]
\end{proposition}
\begin{remark}[Interpretation of the error bound]
We can interpret $\error{exch}$ as measuring the difference between the elements of the exchangeable pair,
$\error{cov}$ as measuring the difference in the covariance of $\newy$ and $\newz$, 
and $\error{rem}$ as measuring the magnitude of the remainder term in the ``regression equation'' \cref{eq:regression-eqn-for-exch-pair}.    
\end{remark}

\subsection{Comparison Processes and the Exchangeable Pair}\label{sec:comparison_exchangeable}
The forms of $Y$ and $Z$ are such that we cannot directly apply \cref{ThmM}. 
Therefore, we introduce two intermediate processes, $\newy$ and $\newz$, and bound
\[
\abs{\EE g(Y)-\EE g(Z)} 
&\le \abs{\EE g(Y)-\EE g(\newy)}+\abs{\EE g(\newy)-\EE g(\newz)}+\abs{\EE g(\newz)-\EE g(Z)},
\]
where we use direct methods to bound the first and third terms (see \cref{BDforR,BDforZ} in \cref{sec:main-theorem-proof}).

We take $\newz$ to be a discretized version of $Z$ that can be written as the sum of Gaussian increments: 
\[ \label{formulanewz}
\newzt 
&\defas Z_{\floor{\timescale t}/\timescale} 
=\sqrt{\frac{ \spacescale^2 \stepsize^2\alpha}{b} \Omega  + \frac{2\spacescale^2 \stepsize\alpha}{\invtemp}}\int_{0}^{\floor{\timescale t}/\timescale} e^{-\stepsize \FI (\floor{\timescale t}-\timescale s)}\d W_s.
\]
We define $\newy$ as the rescaled version of different SG(L)D iterates $(\newpmt_\iter)_{\iter \in \N}$ 
that use a linear approximation to the stochastic gradient. 
For the remainder of the paper, whenever a single source of stochastic gradients is considered, without loss of generality we assume that $\mle = 0$. (When we will consider a sequence of possible sources, as in \cref{sec:weak-conv}, it will be necessary to keep track of all of their different critical points.)
We write the linear approximation to $\grad \obsloss{i}(\pmt)$ using a Taylor expansion at $\mle = 0$
as $\widehat\grad\obsloss{i}(\pmt) \defas \psi_i - \sigma_i\pmt$, where 
$\psi _i \defas \grad\obsloss{i}(\mle)$ and $\sigma_i \defas -\grad^{\otimes 2}\obsloss{i}(\mle)$. 
Hence the linear approximation to the stochastic gradient $\gradstochloss{\iter}(\pmt)$ 
is given by 
\[
\gradstochlinloss{\iter}(\pmt) 
\defas \batchsize^{-1} \sum_{i=1}^\batchsize \widehat\grad\obsloss{I(\iter, i)}(\pmt)
\]
and the corresponding SG(L)D update equation is 
\[\label{sgdnewpmt}
\newpmt_{\iter+1} = \newpmt_{\iter} + \stepsize\,\gradstochlinloss{\iter}(\newpmt_\iter) + \sqrt{\frac{2\stepsize}{\invtemp}}\noise_{\iter}.
\]
Letting 
\[ \label{formulanewy}
\newyt \defas \spacescale \newpmt_{\floor{\timescale t}},
\]
we define the intermediate process $\newy \defas (\newyt)_{t \in [0,1]}$.

We construct the exchangeable pair for $\newy$ by randomly replacing one stochastic gradient with an independent and identically distributed copy.
Let $\swapiter \distas \unifdist\{0,\cdots,n-1\}$ independent and define the stochastic gradients for the exchangeable pair as 
\[
\gradstochlinloss{\iter}'= 
\begin{cases}
    \gradstochlinloss{\iter} & \text{if $\iter \ne K$} \\
    \batchsize^{-1} \sum_{i=1}^\batchsize \widehat\grad\obsloss{I'(K, i)} & \text{if $\iter = K$},
\end{cases}
\]
where $I'(K,i)$ is an independent copy of $I(K,i)$. 
Define an exchangeable copy of the original SG(L)D iterates by
\[
\exparm_{\iter+1} &= \exparm_{\iter} +\stepsize\gradstochlinloss{\iter}'(\exparm_{\iter})+\sqrt{\frac{2h}{\invtemp}}\xi'_k, &
\exparm_{0} & = \pmt_{0}.
\]
Now, letting $\exyt \defas w\exparm_{\floor{\alpha t}}$, we can define the second process in the exchangeable pair as $\exy \defas (\exyt)_{t\in[0,1]}$. 

\section{Functional Error Bounds between Stochastic Iterative Algorithms and Ornstein--Uhlenbeck Processes}\label{sec:main}

\subsection{Assumptions}
Our main result relies on the following assumptions, all of which are standard or weaker than those typically used in non-asymptotic analyses of SG(L)D \citep{10.1214/19-AOS1850,NIPS2011_40008b9a,brosse2017,kushner2003stochastic,pflug1986}

\begin{assumption}[Nonnegative curvature]   \label{asmp:curvature} 
For all $i \in \{1,\dots,n\}$, $\sigma_i \ge 0$. 
\end{assumption}
\begin{assumption}[Second-order smoothness] \label{asmp:hosmooth}
The second-order smoothness constant 
\[
C_R \defas \sup_\pmt \max_{i = 1,\dots, \samplesize} \frac{|\grad\obsloss{i}(\pmt) - \widehat\grad\obsloss{i}(\pmt)|}{(\pmt - \mle)^2}
\]
is finite. 
\end{assumption}
\begin{assumption}[Step size upper bound] \label{asmp:stepsize} 
The step size satisfies $\stepsize \in (0, 1/(2L))$, where $L \defas \max\{1, \sigma_1,\sigma_2,\dots,\sigma_n\}$.
\end{assumption}
\begin{assumption}[Iterate moment bounds] \label{asmp:bounded-iterates} 
There exist constants $K_1, K_3, \overline{\stepsize} > 0$ such that for $\stepsize \in (0, \overline{\stepsize})$ and $p \in \{1, 3\}$,
\[
\EE\{|\theta_k|^{2p}\}^{1/p} &\le K_p h \tau^2_{2p}, 
\]
where for $q > 0$, $\tau_q \defas \EE\cbra{\abs{b^{-1}\sum_{i=1}^{b}\psi_{I(1,i)} + \rbra{\frac{2}{h\invtemp}}^{1/2}\noise_1}^q}^{1/q}$.
\end{assumption}
\begin{remark}[Assumptions] \label{rmk:assumptions}
While \cref{asmp:curvature} is not strictly necessary, it substantially simplifies the statement
of the main result. 
The factor of 2 in \cref{asmp:stepsize} could be replaced by any constant $> 1$, which would only change
some of the constant factors in our error bounds.
Under the assumptions used in \citet[][Lemma 13]{10.1214/19-AOS1850}, and using their notation,
\cref{asmp:bounded-iterates} holds with $\overline{\stepsize} = 1/(L C_3)$ and $K_p = D_p / \mu$; however, we note that this $L$ is not the same as ours. 
\end{remark}

\Cref{asmp:curvature,asmp:hosmooth,asmp:bounded-iterates} hold for many common statistical models, as the following example illustrates. 
\begin{example}\label{ex:glm}
To illustrate conditions under which our assumptions hold, we consider applying SGD to a canonical generalized linear model with observations $\rbra*{x_i,y_i}_{i\in[n]}$. 
In this case,
\[
    \grad\obsloss{i}(\theta) = \cbra*{y_i-q(\theta x_i + \beta_0)}x_i,
\] 
where $\beta_0$ is a (fixed) intercept term and $q$ is the log partition function. \\

\noindent  \textbf{Linear regression:} In this case $q(u)=u$, so
    $\sigma_i = x_i^2 \ge 0$ and  $|\grad\obsloss{i}(\pmt) - \widehat\grad\obsloss{i}(\pmt)| = 0$. 
    Hence, \cref{asmp:curvature,asmp:hosmooth} hold. 
    For \cref{asmp:bounded-iterates}, see to \citet[Example 1]{10.1214/19-AOS1850}. \\

\noindent    \textbf{Logistic regression:}
    In this case $q(u)=e^u/(1+e^u)$, so $\sigma_i = x_i^2 {e^{\beta_0}}/{(1+e^{\beta_0})^2}\ge 0$
    and 
    \[ 
    |\grad\obsloss{i}(\pmt) - \widehat\grad\obsloss{i}(\pmt)|
    \le \frac{e^{-(|x_i \pmt+\beta_0|+|\beta_0|)}(1-e^{-2(|x_i \pmt+\beta_0|+|\beta_0|)})}{2(1+e^{-(|x_i \pmt+\beta_0|+|\beta_0|)})^4}x_i^2 \pmt^2
    \le \frac{x_i^2}{27}\pmt^2.
    \]
    So \cref{asmp:hosmooth} holds with $C_R \le \max_{i\in[n]} x_i^2/27$.  For \cref{asmp:bounded-iterates}, see \citet[Example 1]{10.1214/19-AOS1850}. \\

\noindent  \textbf{Poisson regression:} $q(u)=e^u$. In this case, $\sigma_i = e^{\beta_0}x_i^2 \ge 0$ and  $|\grad\obsloss{i}(\pmt) - \widehat\grad\obsloss{i}(\pmt)|\le \frac{e^{|x_i\pmt+\beta_0|+|\beta_0|}}{2}x^2_i\pmt^2$. Thus for bounded parameter domains, \cref{asmp:hosmooth} holds. 
\Cref{asmp:bounded-iterates} also holds with suitably constrained parameter domain since it will satisfy all the assumptions in \citet{10.1214/19-AOS1850}.
Notably, the $L$-co-coercivity assumption will be true with convex 
and $L$-smooth loss function, which is true with a parameter parameter domain). \\

\noindent \textbf{Exponential families:} These examples also hold for SGD applied to an exponential family with density $f(z \mid \pmt)=\exp\cbra*{\pmt T(z) -A(\pmt)+B(z)}$ and observations $(z_i)_{i\in[n]}$ by taking $\beta_0 = 0$, $y_i=T(z_i)$, and $x_i=1$. 
\end{example}

\subsection{Error Bounds}
We give bounds for the functional approximation error $\abs{\EE \testfun(Y)-\EE \testfun(Z)}$  under two parameter setups which cover common use cases
and ensure the scaled iterates will have a stable and non-trivial limit.

\paragraph*{\textbf{Statistical setting}} 
For positive constants $w_1$ and $w_2$, batch size $\batchsize = \batchsizen$, and number of epochs 
$m = m^{(\samplesize)}$,  set  $\stepsize = \frac{2 w_1 \batchsize}{\samplesize}$, $\invtemp= \frac{n}{w_2}$,
$\timescale = \frac{m\samplesize}{b}$, and $\spacescale = \samplesize^{1/2}$.
Assume that for some positive constant $c$, $m \le c n^{1/2}$.
By \emph{number of epochs} we mean the expected number of times each loss function $\obsloss{i}$ 
is used to compute the stochastic gradient. 
This setup covers cases where the goal is to provide uncertainty quantification for the parameter, 
which can be obtained with appropriate choices of the weights $w_1$ and $w_2$ \citep{negrea2022statistical}. 
The step size $\stepsize$ is set such that the stationary distribution of the OU process is independent
of the choice of $\batchsize$. 
We are interested in the dependence on the sample size $\samplesize$, batch size $\batchsize$,
and number of epochs $m$. 
In particular, for the batch size, we are interested in selecting $\batchsize$ to minimize the error bound.

\paragraph*{\textbf{Numerical setting}}
For positive constants $c_1$, $c_2$ and $c_3$, set $\timescale = c_1 \stepsize^{-1}$, $\spacescale = c_2\{(\batchsize/\stepsize)^{1/2} \wedge \invtemp^{1/2}\}$ and assume $\batchsize\le c_3\stepsize^{-4}$. Assume $\stepsize\leq 1$, $\invtemp\geq 1$.
This setup covers cases where the goal is to obtain a good estimate to $\mle$.
We are interested in the dependence on $\stepsize$, $\batchsize$, and $\invtemp$. 
Unlike the statistical setup, the resulting bounds do not have any explicit dependence 
on $\samplesize$ or the number of epochs.

To state our main result in these two setups, for $\testfun\in M$, let $\bar\testfun \defas \testfun/\norm{\testfun}_M$. 

\begin{theorem}\label{thm:main-theorem-simple}\label{THM:MAIN-THEOREM-SIMPLE}
Retain the notation of \cref{sec:approach}. Suppose that processes $Y$ and $Z$ are univariate (real-valued). For all $\testfun\in M$, under \crefrange{asmp:curvature}{asmp:bounded-iterates}, if $Z_0 = \pmt_0 = 0$, then the following holds:
\begin{enumerate}
    \item 
In the statistical setting, there exists a constant $C_{\mathrm{stat}} \in O\left(C_R^3+C_RL^{6}+L^7\right)$
independent of $m$, $\samplesize$, and $\batchsize$ such that 
\[
    \abs{\EE \bar\testfun(Y)-\EE \bar\testfun(Z)} 
        &\le C_{\mathrm{stat}} \sqrt{\frac{m^{6}\batchsize\cbra*{\log(\samplesize/\batchsize) + m^{6}}}{\samplesize}}.
\]
\item
In the numerical setting, there exists a constant $C_{\mathrm{num}} \in O\left(C_R^3+C_RL^{6}+L^7\right)$ independent of $\stepsize$, $\batchsize$, and $\invtemp$ such that
\[
    \abs{\EE \bar\testfun(Y)-\EE \bar\testfun(Z)} 
        &\le C_{\mathrm{num}} \sqrt{\stepsize\log(h^{-1}) + \invtemp^{-1}}.
\]
\end{enumerate}
\end{theorem}

We prove each part of \cref{thm:main-theorem-simple} as a corollary to the \emph{much more general result} \cref{MainThm} in the supplementary material.
The proof of \cref{thm:main-theorem-simple} can be found in \cref{sec:proof-of-main-theorem-simple} of the supplementary material.
The proof of a simplified version of \cref{MainThm} in the supplementary material is sketched in \cref{sec:main-theorem-proof}.

\begin{remark}[Interpretation of \cref{thm:main-theorem-simple}]
In the statistical setting, the error bound is minimized by selecting $b = 1$, which is equivalent to choosing
the smallest possible step size that leads to the desired asymptotic stationary distribution. 
Furthermore, using the relationship between $h$ and $b$, we can see that the bound in the statistical setting is of order 
$
\sqrt{\stepsize m^{6}\cbra*{\log(1/\stepsize) + m^{6}}}.
$
Hence, in both the statistical and numerical settings, the error bound is of order $\sqrt{\stepsize \log(1/\stepsize)}$ for $\stepsize \to 0$. 
So, in the numerical setting, the error bound is minimized by choosing $h$ as small as possible, as we would expect.
However, since $\invtemp$ is now a free parameter, the Gaussian noise must be scaled down sufficiently quickly
as well (namely, at an $\stepsize \log(1/\stepsize)$ rate or faster). We note that the order of our bound is the same as the order of the tightest bound available in the literature \citep{barbour1990diffusion} for the approximation of a random walk with Brownian motion with respect to the metric generated by the class of test functions $M$ defined in \cref{sec:exchangeable_pairs}. 
Finally, we note that the statistical setting error bound is of order $m^6$.
An interesting question for future work is whether this dependence on the number of epochs is optimal.
It seems plausible the dependence should be linear or close-to-linear in $m$. 
\end{remark}

\begin{remark}[Extension to multivariate setting] \label{rmk:multivariate-extension}
\Cref{thm:main-theorem-simple} holds for univariate $\pmt$. Investigating the distance between $Y$ and $Z$ for $\pmt \in \R^d, d> 1$, would require an additional lengthy and technically intricate analysis.
A particular challenge is that the gradient steps of a stochastic approximation algorithm do not commute. 
In the case of the exchangeable pairs method for convergence of a random walk to Brownian motion, the original path and its exchangeable pair move perfectly in parallel after the random substitution point. 
Non-commutativity of gradient steps arises from two sources: that the second derivative matrices, $\sigma_i$, that appear in the Taylor expansion 
\[
    \grad \ell_i(\theta) 
        & = \psi_i + \sigma_i (\theta - \trueparm) + R_i(\theta)
\]
do not commute; furthermore, the remainder term $R_i(\theta)$ is non-constant.
In the present work, we handle the remainder term, while deferring analysis of non-commutativity of the second derivative matrices.
In order to use proof techniques similar to ours for $d > 1$, we would first need to establish approximate commutativity of the matrix multiplication operation on the collection of matrices $(I-\frac{h}{b}\sum_{i=1}^{b}\Heisbatch{m}{i})_{m=1}^{\timescale}$ since
exact commutativity only holds in the most trivial of examples; for instance, it fails to hold for the generalized linear models discussed in \Cref{ex:glm}.
Thus, we believe a multi-dimensional extension of our main result likely holds -- and would be worthwhile and useful.
However, we consider it out-of-scope of the present paper to keep
the length manageable.
\end{remark}

\begin{remark}[Comparison to other applications of functional Stein's method]
That the order of gradient steps matter (even in one dimension), unlike the order of random walk steps, implies that the paths of the exchangeable pair do not move perfectly in parallel in the setting of the present work. 
Compare to the case of convergence of a random walk to Brownian motion, where this parallel shift property had been used to find that the difference between the exchangeable pair is, in some average sense, equivalent to a scaled-down path of the same process \citep{kasprzak2020exchangeable, dobler2021stein, barbour1990diffusion}.
In this case, the self-similarity property has been heavily exploited to obtain tight quantitative bounds. 
In our case, without an equivalent parallel shift structure, more care must be taken to correctly handle the resulting difference in paths of the exchangeable pair. 
The methods we use for this are novel and are an important step in extending exchangeable pair, and functional Stein's method more generally, to a wider class of approximated and approximating processes. 
Demonstrating this possible utility is one of the present work's broader technical contributions. This corresponds to the fact that the approximating OU process does not have \emph{independent increments}, unlike in previous applications of functional Stein's method of exchangeable pairs.  \label{rmk:comparison_stein} 
\end{remark}

\section{Functional Error Bound for Ornstein--Uhlenbeck processes}
\label{sec:boundforouprocess}

\subsection{Motivation and Setup}\label{sec:motivation_and_setup}
In the statistical setting, we must consider a sequence of problem instances with different underlying sample sizes for the empirical distribution, indexed by the sample size $\samplesize$.
In this case, we must index the rescaled SG(L)D iterates $Y$ and the limiting OU process $Z$ by the sample size as well.
Specifically, the SG(L)D update equation becomes
\[ 
\pmtn_{\iter+1} = \pmtn_{\iter} + \stepsizen \gradstochloss{\iter}^{(\samplesize)} + \sqrt{\frac{2\stepsizen}{\invtempn}}\noise_{\iter},
\]
where $\gradstochloss{\iter}^{(\samplesize)}\defas \frac{1}{\batchsizen}\sum_{i=1}^{\batchsizen} \grad \obsloss{I^{(\samplesize)}(\iter, i)}(\pmt)$ with independent, uniformly distributed random
variables $I^{(\samplesize)}(k, i) \in \{1,\dots,\samplesize\}$ (for $\iter \in \N, i \in \{1,\dots,\batchsize\}$).
Thus, the rescaled iterate process $Y^{(\samplesize)} = (Y^{(\samplesize)}_t)_{t \in [0,1]}$ is defined by
\[ 
Y^{(\samplesize)}_t \defas \spacescalen\rbra*{\pmt_{\floor{\timescalen t}}-\MLEparm{\samplesize}}, %
\]
where $\MLEparm{\samplesize} \in \argmin_{\theta} \frac{1}{\samplesize}\sum_{i=1}^\samplesize \ell_i(\theta)$ is the loss-minimizer when the sample size is $\samplesize$.
The corresponding OU process $Z^{(\samplesize)} \defas (Z^{(\samplesize)}_t)_{t \in [0,1]}$ satisfies
the SDE
\[
    \begin{cases}
        \d Z^{(\samplesize)}_t = -2w_1 m\FI^{(\samplesize)} Z^{(\samplesize)}_t \,\d t + 2\sqrt{w_1 m \rbra{w_1\foFI^{(\samplesize)}  +  w_2  I}}\, \d\weinerprocess_{t} \\
        Z^{(\samplesize)}_0 = Y^{(\samplesize)}_0
    \end{cases},
\]
where $\FI^{(\samplesize)} \defas -\samplesize^{-1}\sum_{i=1}^{\samplesize}\grad^{\otimes 2}\obsloss{i}(\MLEparm{\samplesize})$ and $\foFI^{(\samplesize)} \defas \samplesize^{-1}\sum_{i=1}^{\samplesize} \grad \obsloss{i}(\MLEparm{\samplesize})^{\otimes 2}$.

While \cref{thm:main-theorem-simple} bounds the difference between $Y^{(\samplesize)}$ and $Z^{(\samplesize)}$, it remains to show that $Z^{(\samplesize)}$ converges to a \emph{population process} $Z^{(\infty)}$ as $\samplesize \to \infty$.
In fact, \citet{negrea2022statistical} show that $Y^{(\samplesize)}$ convergences
to $Z^{(\infty)} = (Z^{(\infty)}_t)_{t \in [0,1]}$ defined by the SDE 
\[
    \begin{cases}
        \d Z^{(\infty)}_t = -2w_1 m\FI^{(\infty)} Z^{(\infty)}_t \,\d t + 2\sqrt{w_1 m \rbra{w_1\foFI^{(\infty)}  +  w_2  I}}\, \d\weinerprocess_{t} \\
        Z^{(\infty)}_0 = \trueparm
    \end{cases},
\]
where $\FI^{(\infty)} \defas -\EE\cbra*{\grad^{\otimes 2}\obsloss{1}(\mle)}$ and $\foFI^{(\infty)} \defas \EE \cbra*{\grad \obsloss{1}(\mle)^{\otimes 2}}$, with  $\trueparm\in\argmin_{\theta} \EE[\ell_1(\theta)]$ being the population optimum. 
Hence, in the statistical setting, we wish to bound 
\[
\abs{\EE g(Y^{(\samplesize)}) - \EE g(Z^{(\infty)})} \le \abs{\EE g(Y^{(\samplesize)}) - \EE g(Z^{(\samplesize)})} + \abs{\EE g(Z^{(\samplesize)})- \EE g(Z^{(\infty)})}.
\] 
The first term is bounded in \cref{thm:main-theorem-simple}, so all that remains is to bound $\abs{\EE g(Z^{(\samplesize)})- \EE g(Z^{(\infty)})}$. 

\subsection{Error Bound for OU processes}
 To this end, in this section we obtain a general quantitative functional error bound for two arbitrary OU processes starting at the same location.  

\begin{theorem}\label{thm:OU-error-bound}
For two univariate OU processes $Z$ and $\tilde{Z}$ where 
\[\d Z_t = -B Z_t \,\d t + \sqrt{A}\, \d\weinerprocess_{t},\quad\d \tilde{Z}_t = -\tilde{B} \tilde{Z}_t \,\d t + \sqrt{\tilde{A}}\, \d\weinerprocess_{t} \]
and $Z_0 = \tilde{Z_0}$, 
there exists a positive constant $\mathcal{K} = \mathcal{K}(A,\tilde{A},B,\tilde{B})$ such
that
\[
|\EE \bar g(Z)-\EE \bar g(\tilde{Z})| \leq \mathcal{K} \left[ \abs{A-\tilde{A}} + \abs{B-\tilde{B}} +\abs{A-\tilde{A}}^3 + \abs{B-\tilde{B}}^3  \right].
\]
\end{theorem}
\begin{remark}
The positive constant in \cref{thm:OU-error-bound} is given by 
  \[
\mathcal{K}(A,\tilde{A},B,\tilde{B}) = \max \left\{\mathcal{K}_1 +
  2^{5/2}C_3A^{3/2}\mathcal{K}_3^{1/3}, 2\mathcal{K}_3 \right\}.
\]
For any $p \ge 1$, the constant $\mathcal{K}_p = \mathcal{K}_p(A,\tilde{A},B,\tilde{B})$ appearing in the
above expression is given by
\[
\mathcal{K}_p(A,\tilde{A},B,\tilde{B}) = C_p\max \left\{\frac{2^{\frac{3p-4}{2}}}{ \min \{A,\tilde{A}\}^{p/2}},  2^{2(p-1)}\left(   C(\tilde{A},B,\tilde{B})^{p/2}
  +2^{p/2}\tilde{A}^{p/2}e^{p\abs{\tilde{B}-B}}\right)\right\},
\]
where $C_p$ is the constant appearing in \cref{theoremmaxineqouprocesses} and
$C(\tilde{A},B,\tilde{B})$ is the constant appearing in \cref{applicationtooursituation}, which are both stated later in this section. 
\end{remark}

Plugging in the values for $Z^{(\samplesize)}$ and $Z^{(\infty)}$ allows us to obtain a bound for quantity $\abs{\EE g(Z^{(\samplesize)})- \EE g(Z^{(\infty)})}$, as follows.

\begin{corollary}
\label{cor:bd_for_ou}
Recall the notation of \Cref{sec:motivation_and_setup}. We have 
\[\abs{\EE \bar g(Z^{(\samplesize)})- \EE \bar g(Z^{(\infty)})} 
&\leq \mathcal{K}\Big[ 4w_1^2 m\abs{\foFI^{(\samplesize)}-\foFI^{(\infty)}} + 2w_1 m\abs{\FI^{(\samplesize)}-\FI^{(\infty)}} \\
&\phantom{\leq~\mathcal{K}\big[} +64w_1^6 m^3\abs{\foFI^{(\samplesize)}-\foFI^{(\infty)}}^3 + 8w_1^3 m^3\abs{\FI^{(\samplesize)}-\FI^{(\infty)}}^3 \Big],
\]
where constant $\mathcal{K}$ is given by \cref{thm:OU-error-bound}.
\end{corollary}
\begin{proof}
Apply \cref{thm:OU-error-bound} with $B= 2w_1 m\FI^{(\samplesize)}$,
$A=4w_1 m \rbra{w_1\foFI^{(\samplesize)}  +  w_2 I}$, $\tilde{B}=2w_1 m\FI^{(\infty)}$ and
$\tilde{A}=4w_1 m \rbra{w_1\foFI^{(\infty)}  +  w_2  I}$
\end{proof}

\subsection{Proof of the Error Bound for OU Processes}
\begin{proof}[Proof of {\cref{thm:OU-error-bound}}]
The proof of \cref{thm:OU-error-bound} relies on the following $L^p$ maximal inequality for time-inhomogeneous Ornstein--Uhlenbeck processes of the form 
\[\label{ourOUprocessfamily}
  \begin{cases}
    \displaystyle \d X_t = -\driftcoeff X_t \d t + e^{-\driftcoeff t}\noisefun(t)\d W_t\\
    \displaystyle X_0 = 0
  \end{cases},
\]
where $\noisefun : [0,1] \to \R$ is a function that is square-integrable with respect to the Lebesgue measure. 
\begin{theorem}[$L^p$ maximal inequality for Ornstein--Uhlenbeck processes]\label{theoremmaxineqouprocesses}
Let $X_t$ be an Ornstein--Uhlenbeck process of the form \cref{ourOUprocessfamily}. Then, for any $p >0$ and $\gamma > 0$, it holds that
\+[
\retainlabel{eq:mainlpbound}
\EE \left( \sup_{0\leq s \leq t} \abs{X_s}^p \right) &\leq
  C_p\sup_{0\leq s \leq t} \left\{ \left( e^{-2\driftcoeff s} \left[\gamma+\int_0^s
  \noisefun(u)^2\d u\right]\right)^{\frac{p}{2}} \right\} \\
  &\quad \times \log^{\frac{p}{2}} \left(1+ \frac{1}{\gamma}\int_0^t\noisefun(s)^2\d s + \log
    \left( 1+ \frac{1}{\gamma}\int_0^t\noisefun(s)^2\d s \right)\right),
\+]
where $C_p$ is a constant depending solely on $p$. \setbox0\vbox{\cref{eq:mainlpbound}}
\end{theorem}

We will use the following corollary to \cref{theoremmaxineqouprocesses}. 
Both results are proved in \cref{sec:proof_maxineqou} of the supplementary material.
\begin{corollary}\label{applicationtooursituation}
Let $A, \tilde{A}, B, \tilde{B} >0$ be positive constants. Then, for 
any $p >0$, it
holds that
\begin{equation}\label{applicationbound1}
\EE \left[ \sup_{0\leq t \leq 1} \left\vert e^{-Bt}\int_0^t e^{Bs}\left(
      \sqrt{A}-\sqrt{\tilde{A}} \right)\d W_s\right\rvert ^p \right] \leq
\frac{C_p}{2^{p/2} \min \{A,\tilde{A}\}^{p/2}}\abs{A-\tilde{A}}^p,\tag{$i$}
\end{equation}
\begin{equation}\label{applicationbound2}
\EE \left[ \sup_{0\leq t \leq 1} \left\vert e^{-Bt}\int_0^t
    \sqrt{\tilde{A}}\left(e^{Bs} - e^{\tilde{B}s}
     \right)\d W_s\right\rvert ^p \right] \leq
C_p C(\tilde{A},B,\tilde{B})^{p/2}\abs{B-\tilde{B}}^p, \tag{$ii$}
\end{equation}
and
\begin{equation}\label{applicationbound3}
\EE \left[ \sup_{0\leq t \leq 1} \left\vert e^{-\tilde{B}t}\int_0^t
    \sqrt{\tilde{A}} e^{\tilde{B}s}\d W_s\right\rvert ^p \right] \leq
2^{p/2}C_p\tilde{A}^{p/2},\tag{$iii$}
\end{equation}
where $C_p$ is a constant depending only on $p$, and
\[
C(\tilde{A},B,\tilde{B}) = 
  \frac{\tilde{A}(1+2B(1+4\tilde{B}))(3+4\tilde{B})e^{4 \abs{\tilde{B}-B}}}{\tilde{B}(B+\tilde{B})}.
\]
\end{corollary}

We now proceed to our proof of \cref{thm:OU-error-bound}.
Using similar arguments as in \cite{dobler2021stein}, we can write
\[\label{dobkaspboundtouse}
|\EE g(Z)-\EE g(\tilde{Z})| \le \|g\|_{M} \left[\EE  |Z-\tilde{Z}|+2\EE |Z-\tilde{Z}|^3+2\EE |Z|^3\EE (|Z-\tilde{Z}|^3)^{1/3}\right].
\]
In order the control the terms of the form $\EE  |Z-\tilde{Z}|^p$,
we can start by writing
\[
Z_t -\tilde{Z}_t &= \int_{0}^{t}\sqrt{A}e^{-B(t-s)}-\sqrt{\tilde{A}}e^{-\tilde{B}(t-s)}\d W_s\\
&= e^{-Bt}\int_{0}^{t}\left(\sqrt{A}-\sqrt{\tilde{A}}\right)e^{Bs}\d W_s + e^{-Bt}\int_{0}^{t}\sqrt{\tilde{A}}\left(e^{Bs}-e^{\tilde{B}s}\right)\d W_s\\
&\quad +
\left(e^{-Bt}-e^{-\tilde{B}t}\right)\int_{0}^{t}\sqrt{\tilde{A}}e^{\tilde{B}s}\d
W_s,
\]
so that, for any $p \geq 1$,  
\[
\EE |Z-\tilde{Z}|^p &\leq 2^{2(p-1)} \EE \left( \sup_{0\leq t \leq 1} \left\vert e^{-Bt}\int_0^t e^{Bs}\left(
      \sqrt{A}-\sqrt{\tilde{A}} \right)\d W_s\right\rvert ^p \right)\\
&\quad + 2^{2(p-1)}\EE \left( \sup_{0\leq t \leq 1} \left\vert e^{-Bt}\int_0^t
    \sqrt{\tilde{A}}\left(e^{Bs} - e^{\tilde{B}s}
    \right)\d W_s\right\rvert ^p \right)\\
&\quad + 2^{2(p-1)}\EE \left( \sup_{0\leq t \leq 1} \left\vert \left(e^{-Bt}-e^{-\tilde{B}t}\right)\int_0^t
    \sqrt{\tilde{A}} e^{\tilde{B}s}\d W_s\right\rvert ^p \right).
\]
We can bound the first two summands directly by using items
\eqref{applicationbound1} and \eqref{applicationbound2} in
\cref{applicationtooursituation}. For the last summand, observe that
as
\[
\left\vert e^{-Bt}-e^{-\tilde{B}t}\right\rvert  = e^{-\tilde{B}t} \left\vert
  e^{(\tilde{B}-B)t}-1 \right\rvert \leq e^{-\tilde{B}t}
\abs{\tilde{B}-B}e^{\abs{\tilde{B}-B}}
\]
by the mean value theorem and
the fact that $t \leq 1$, we have
\[
&\EE \left( \sup_{0\leq t \leq 1} \left\vert \left(e^{-Bt}-e^{-\tilde{B}t}\right)\int_0^t
    \sqrt{\tilde{A}} e^{\tilde{B}s}\d W_s\right\rvert ^p \right) \\
&\qquad\qquad\qquad\qquad\qquad\qquad\quad \leq \abs{\tilde{B}-B}^pe^{p\abs{\tilde{B}-B}}\EE \left( \sup_{0\leq t \leq 1} \left\vert e^{-\tilde{B}t}\int_0^t
    \sqrt{\tilde{A}} e^{\tilde{B}s}\d W_s\right\rvert ^p \right). 
\]
The right-hand side can now be dealt with using item
\eqref{applicationbound3} in
\cref{applicationtooursituation}. In total, we obtain
\[
\EE |Z-\tilde{Z}|^p &\leq \frac{2^{\frac{3p-4}{2}}C_p}{ \min \{A,\tilde{A}\}^{p/2}}\abs{A-\tilde{A}}^p \\
&\quad + 2^{2(p-1)}C_p \left(   C(\tilde{A},B,\tilde{B})^{p/2}
  +2^{p/2}\tilde{A}^{p/2}e^{p\abs{\tilde{B}-B}}\right)\abs{\tilde{B}-B}^p\\
&\leq \mathcal{K}_p(A,\tilde{A},B,\tilde{B}) \left( \abs{A-\tilde{A}}^p + \abs{B-\tilde{B}}^p \right).
\]
By item
\eqref{applicationbound3} in
\cref{applicationtooursituation}, note that
\[
\EE \|Z\|^3 \leq 2^{3/2}C_3A^{3/2},
\]
so that using \cref{dobkaspboundtouse}, we finally obtain
\[
|\EE g(Z)-\EE g(\tilde{Z})| \leq \|g\|_{M}\Bigg[&
\left(\mathcal{K}_1 +
  2^{5/2}C_3A^{3/2}\mathcal{K}_3^{1/3}\right)\left(
  \abs{A-\tilde{A}} + \abs{B-\tilde{B}}  \right)\\
& +2\mathcal{K}_3\left(
  \abs{A-\tilde{A}}^3 + \abs{B-\tilde{B}}^3  \right)\Bigg],
\]
which concludes the proof.
\end{proof}
\section{Implications of the Error Bounds} \label{sec:implications}
The quantitative bounds provided by \cref{thm:main-theorem-simple,thm:OU-error-bound} are useful for a number of reasons, as we illustrate in the following three implications.

\subsection{Variance of Iterate Averages}

As a first example, our main result allows us to understand the finite-sample, non-zero step-size behavior of these algorithms. 
For tuning stochastic gradient algorithms using the functional Bernstein--von Mises theorem of \citet{negrea2022statistical}, asymptotic (qualitative) results only technically allow one to falsify 
a bad tuning in the sense that, for sufficiently large sample sizes, the finite-sample-size 
non-zero-step-size behavior will be close to the OU process limit. 
Qualitative limit results fail to signal if any particular sample-size--step-size pair is close, 
so we cannot determine if a tuning is actually ``good’’ in this way. 
Quantitative bounds, on the other hand, do provide results that directly reflect the performance 
of the actual set-up a practitioner might be using with real pre-limiting values for these quantities.
For example, one implication of practical relevance \citep{kushner2003stochastic,negrea2022statistical} is
that we can bound the difference between the variance of iterate averages and the variance of its scaling limit.

\begin{corollary}[Error bound for the variance of iterate averages]\label{coritav} Recall the notation of \cref{sec:approach,sec:main} and assume that $Y$ and $Z$ are univariate processes. Let $\epsilon>0$ and define $\bar{Y} = \int_{0}^{1}Y_t \,\d t$ and $\bar{Z} = \int_{0}^{1}Z_t \,\d t$. Then $\abs{\EE \bar\testfun(Y)-\nobreak\EE \bar\testfun(Z)}\le \epsilon$ implies that 
\[\label{eq:cor_claim}
\abs{\Var(\bar{Y})-\Var(\bar{Z})} \le \rbra*{1.53\abs{\EE\bar{Y}}+3.53}\epsilon.
\]
Furthermore, under the hypotheses of \cref{thm:main-theorem-simple}, 
\[
\abs{\EE\bar{Y}} \le K_1 C_R \spacescale\timescale\stepsize \rbra*{\frac{\stepsize\Omega}{\batchsize}+\frac{2}{\invtemp}}.
\]
Therefore, in both the statistical and numerical settings, under the hypotheses of \cref{thm:main-theorem-simple}, $\abs{\EE\bar{Y}}\in o(1)$ and so $\abs{\Var(\bar{Y})-\Var(\bar{Z})} \le C \epsilon$,
where $C$ is constant. 
\end{corollary}
\begin{proof}%
  Since $\EE Z_t = 0$,
\[
\abs{\Var(\bar{Y})-\Var(\bar{Z})}
&=\abs{\EE\bar{Y}^2 -\EE\bar{Z}^2 - \EE\bar{Y}\rbra*{\EE\bar{Y} -\EE\bar{Z}}} \le\abs{\EE\bar{Y}^2 -\EE\bar{Z}^2}+\abs{\EE\bar{Y}}\abs{\EE\bar{Y} -\EE\bar{Z}}.
\]
Letting $g_2(X) \defas \rbra*{\int_{0}^{1}X_t \d t}^2$ and $g_1(X) \defas \int_{0}^{1}X_t \d t$, it follows
that 
\[ \label{eq:cor_first_eq}
\abs{\Var(\bar{Y})-\Var(\bar{Z})} \le \rbra*{\norm{g_1}_M \abs{\EE\bar{Y}}+\norm{g_2}_M}\epsilon.
\]

Now, fix $\skelement,\skothelement,\skothelement^1,\skothelement^2\in\skspace$ and note that
\[
\lim_{\norm{\skothelement}\to 0}\frac{g_1(\skelement+\skothelement) - g_1(\skelement) - g_1(\skothelement)}{\norm{\skothelement}}=0.
\]
Moreover, note that $g_1: \left(\skspace,\norm{\cdot}\right)\to\left(\mathbb{R},|\cdot|\right) $ is a bounded linear operator. Therefore,
\[
\FD g_1(\skelement)[\skothelement]=g_1(\skothelement)=\int_0^1 \skothelement_t \,\d t 
\quad\text{and}\quad
\FD^2g_1(\skelement)[\skothelement^1,\skothelement^2]=0.
\]
It follows that
\[\label{eq:g_1_norm}
\norm{g_1}_M 
    &= \sup_{\skelement\in \skspace}\frac{|g_1(\skelement)|}{1+\norm{\skelement}^3}
        +\sup_{\skelement\in \skspace}\frac{\norm{\FD g_1(\skelement)}}{1+\norm{\skelement}^2} \\
    &\leq \sup_{\skelement\in \skspace} \frac{\|\skelement\|}{1+\norm{\skelement}^3}+\sup_{\skelement\in \skspace}\sup_{\norm{\skothelement}\leq 1} \frac{\left|\int_0^1 \skothelement_t\d t\right|}{1+\norm{\skelement}^2}\\
    &\leq  \frac{2^{2/3}}{3} + 1 < 1.53 .
\]

Again fixing $\skelement,\skothelement,\tilde\skothelement \in\skspace$, by the chain rule,
\[
\FD g_2(\skelement)[\skothelement] = \FD(g_1^2)(\skelement)[\skothelement]=2 g_1(\skelement)\FD g_1(\skelement)[\skothelement]=2\left(\int_0^1 \skelement_t\,\d t\right)\left(\int_0^1 \skothelement_t\,\d t\right). 
\]
It follows that
\[
\FD^2g_2(\skelement)[\skothelement,\tilde\skothelement] 
= 2\left(\int_0^1 \skothelement_t\,\d t\right)\FD g_1(\skelement)[\tilde\skothelement] 
= 2 \left(\int_0^1 \skothelement_t\,\d t\right)\left(\int_0^1 \tilde\skothelement_t\,\d t\right).
\]
Therefore,
\[
&\sup_{\skelement\in \skspace}\frac{|g_2(\skelement)|}{1+\norm{\skelement}^3} \leq \sup_{\skelement\in \skspace}\frac{\norm{\skelement}^2}{1+\norm{\skelement}^3} < 0.53, \\
&\sup_{\skelement\in \skspace} \frac{\norm{\FD g_2(\skelement)}}{1+\norm{\skelement}^2}\leq  \sup_{\skelement\in \skspace}\frac{2\norm{\skelement}}{1+\norm{\skelement}^2} = 1, \\
&\sup_{\skelement\in \skspace} \frac{\norm{\FD^2 g_2(\skelement)}}{1+\norm{\skelement}}\leq \sup_{\skelement\in \skspace}\frac{2}{1+\norm{\skelement}} = 2,\\
\]
and $\FD^2 g_2 (\skelement+\skothelement) - \FD^2g_2(\skelement) = 0$ for all $\skelement,\skothelement\in\skspace$, which implies that 
\[\label{eq:g_2_norm}
\norm{g_2}_M\leq 3.53.
\]
Now, \cref{eq:cor_claim} follows from \cref{eq:cor_first_eq,eq:g_1_norm,eq:g_2_norm}.

In order to control $\abs{\EE\bar{Y}}$, we note that by \cref{lm:pmt} below, only the remainder part of $Y_t$ is not mean zero. Using the following notation
\[
Q(j,k)= 
\begin{cases}
    \prod_{m=j}^{k-1}\rbra*{1-\frac{\stepsize}{b}\sum_{i=1}^{b}\Heisbatch{m}{i}} & \text{if $j <k$} \\
    1 & \text{if $j\ge k$}
\end{cases}
\] 
(as in \cref{eq:Q_def} below), we note that $0\le Q(j, k) \le 1$, by \cref{asmp:curvature,asmp:stepsize}.
Now, by \cref{asmp:hosmooth}, and \cref{asmp:bounded-iterates}, we have
\[
\abs{\EE\bar{Y}} 
&= \abs*{\frac{1}{\timescale}\sum_{k=1}^{\timescale}\spacescale\stepsize\sum_{j=0}^{k-1} \EE\cbra*{Q(j+1,k)\rbra*{\gradstochloss{j}(\pmt_j)-\gradstochlinloss{j}(\pmt_j)}}}  \\
&\le wh K_1 C_R \sum_{j=0}^{\timescale}\EE(\pmt_j^2) \\
&\le w\timescale K_1C_R\rbra*{\frac{h^2\Omega}{b}+\frac{2h}{\invtemp}}.
\]
So, under the hypotheses of \cref{thm:main-theorem-simple}, in the statistical setting
\[
\abs{\EE\bar{Y}}\le 4K_1 C_R w_1 m(w_1\Omega+w_2)n^{-1/2} \in o(1),
\]
while in numerical setting
\[
\abs{\EE\bar{Y}}\le K_1 C_R c_1c_2^2(\Omega+2)w^{-1} \in o(1).
\]
\end{proof}

\subsection{Weak Convergence}\label{sec:weak-conv}

A second reason \cref{thm:main-theorem-simple} is useful in that quantitative bounds provide a more modular tool for probing the qualitative asymptotics---especially in cases where the target distribution is itself random.
Quantitative bounds allow us to leverage (implicitly) fixed target distribution results to obtain results 
even when the target distribution is changing. 
This is of particular importance when the target undergoes stochastic convergence. 
The main example we consider below is the empirical distribution of a data set changing with its sample size, 
and converging to the underlying sampling distribution. 
Classical results on the OU limit of stochastic approximations,  such as those of \citet{kushner2003stochastic},
are derived in a setting where the gradients are sampled from a fixed distribution. 
However, in the statistical settings of interest, the distribution from which gradients are sampled is stochastic and changing.
Accounting for stochastic convergence of the target, we aim to derive a functional Bernstein--von Mises theorem, as in \citet{negrea2022statistical} and required by \citet{mandt2017stochastic}. 

Quantitative bounds allow us to determine when it is valid to interchange or combine the limits as 
the step size vanishes and the sample size increases, while qualitative bounds do not. 
In previous work, the qualitative versions of such joint limits had to be derived separately or in a bespoke way for each regime in which the step size and sample size vary together.
But quantitative bounds allow this to be done in a modular way: a quantitative bound can be derived first, and 
then---after checking for appropriate uniformity of the quantitative bounds---qualitative limiting results 
can be derived and limits safely interchanged.
To illustrate how this can be done, we start by showing that our error bounds can guarantee weak convergence.

\begin{corollary}\label{prop:weak-convergence} ~
    \begin{enumerate}
    \item  In the statistical setting, 
        retain the setup and assumptions of \Cref{thm:main-theorem-simple} for each sample size $n$ with corresponding parameters $C_R=C_R\upper{n}, L=L\upper{n}$. 
        Assume that $K_1$ and $K_3$ of \cref{asmp:bounded-iterates} are almost surely bounded in $n$. Recall the notation of \Cref{sec:motivation_and_setup}.
        Suppose that
    \[
    \log^2\left({m\samplesize}/{\batchsize}  \right)\cbra*{\rbra*{C_R\upper{n}}^3 + C_R\upper{n}\rbra*{L\upper{n}}^{6}+\rbra*{L\upper{n}}^7}\sqrt{\frac{m^{6}\batchsize\cbra*{\log(\samplesize/\batchsize) + m^{6}}}{\samplesize}}\xrightarrow{\samplesize\to\infty} 0 
    \]
    and
    \[
        \abs*{\Omega\upper{n}-\Omega\upper{\infty}}&\xrightarrow{\samplesize\to\infty} 0 &  \text{for } \Omega\upper{\infty}>0, \\
        \abs*{\Sigma\upper{n}-\Sigma\upper{\infty}}&\xrightarrow{\samplesize\to\infty} 0 &  \text{for }  \Sigma\upper{\infty}>0.      
    \]
    Then, the law of $Y\upper{n}$ converges weakly to the law of $Z\upper{\infty}$ as $n\to\infty$, in \skspace, with respect to both the uniform and the Skorokhod topologies.
    \item Similarly, in the numerical setting, retaining the setup and assumptions of \Cref{thm:main-theorem-simple}, suppose that 
    \[
    \log^2(\stepsize^{-1})\rbra*{C_R^3+C_RL^{6}+L^7}\sqrt{\stepsize\log(\stepsize^{-1}) + \invtemp^{-1}}\xrightarrow{\stepsize\to 0}0.
    \]
    Moreover, assume that the law of $Z$ converges as $h\to 0$ weakly in \skspace, with respect to the Skorokhod topology, to the law of a random element $Z^{(\lim)}$ of $C[0,1]$. Then, the law of $Y$ converges weakly to the law of $Z^{(\lim)}$ as $h\to 0$, in \skspace, with respect to both the uniform and the Skorokhod topologies.
    \end{enumerate}
\end{corollary}
\begin{proof}
    \Cref{prop:weak-convergence} follows directly from \cref{thm:main-theorem-simple,thm:OU-error-bound} and \citet[Proposition 3.1]{barbour2009functional}.  
\end{proof}

\begin{remark}
    Verifying that $K_1$ and $K_3$ are bounded is problem-dependent. 
    However, we would typically expect this condition to hold.
    For example, consider using  \citet[][Lemma 13]{10.1214/19-AOS1850} (see \cref{rmk:assumptions}). 
    Then it suffices to verify the following: 
    (1) $\obsloss{i}$ is almost surely convex, 
    (2) $M^{(\samplesize)} \defas \sup_{j \in \{2,\dots,5\}} \EE\left[\sup_\pmt |\frac{\d^j}{\d \pmt^j} \obsloss{1}^{(\samplesize)}(\pmt)|\right] < \infty$, and
    (3) for the strong convexity constant of 
    $\loss^{(\samplesize)} = \samplesize^{-1} \sum_{i=1}^\samplesize \obsloss{i}(\pmt)$,
    which we denote by  $\mu^{(\samplesize)}$, to satisfy
    $\liminf_{\samplesize \to \infty} \mu^{(\samplesize)} > 0$.
\end{remark}

The next result provides practically verifiable conditions under which the abstract conditions required by \cref{prop:weak-convergence}
hold in the statistical setting. 
In light of the previous remark, we still assume $K_1$ and $K_3$ to be bounded. 
\begin{corollary}\label{cor:statistical-weak-convergence}
    Assume that $K_1$ and $K_3$ of \cref{asmp:bounded-iterates} are almost surely bounded in $n$.
    If the loss functions $\ell_i$ are sampled i.i.d.\ such that for positive constants $c_1,\ c_2,\ c_3$ such that $\EE[\norm{\grad\ell_i(\cdot)}_{L^\infty}^{c_1} ]<\infty$,  $\EE[\norm{\hess\ell_i(\cdot)}_{L^\infty}^{c_2} ]<\infty$, and
    $\EE[\norm{\grad^{\otimes 3}\ell_i(\cdot)}_{L^\infty}^{c_3}]<\infty$, with $c_1\geq 8$, $c_2\geq 18$ and $c_3>6$, and
    $\norm{\theta\upper{n}_\star-\theta_\star}\leq C_\theta n^{-1/2}$ all but finitely often,
    then the premises of \cref{prop:weak-convergence} in the statistical setting hold almost surely.
\end{corollary}

\subsection{Proof of \cref{cor:statistical-weak-convergence}}

\Cref{cor:statistical-weak-convergence} follows directly from
\cref{prop:statistical-weak-convergence-1,prop:sigma-bound,prop:omega-bound}, which state and prove subsequently.

\begin{proposition}\label{prop:statistical-weak-convergence-1}
    If the loss functions $\ell_i$ are sampled i.i.d.\ such that for positive constants $c_2$ and $c_3$  it holds that $\EE[\norm{\hess\ell_i(\cdot)}_{L^\infty}^{c_2} ]<\infty$, and
    $\EE[\norm{\grad^{\otimes 3}\ell_i(\cdot)}_{L^\infty}^{c_3}]<\infty$,
    then 
    \[
        \rbra[\big]{C_R\upper{n}}^3 + C_R\upper{n}\rbra[\big]{L\upper{n}}^{6}+\rbra[\big]{L\upper{n}}^7 \leq \samplesize^{\frac{3}{c_3}} + \samplesize^{\frac{1}{c_3}+\frac{6}{c_2}} +  \samplesize^{\frac{7}{c_2}}
    \]
    all but finitely often with probability 1.
\end{proposition}

Generalizing our previous notation, let $\hat\grad\ell_i\upper{\samplesize}(\pmt)
        \defas \psi_i\upper{\samplesize}+ \sigma_i\upper{\samplesize} (\pmt-\MLEparm{\samplesize})$,
where $\psi_i\upper{\samplesize} \defas \grad\ell_i(\MLEparm{\samplesize})$
and $\sigma_i\upper{\samplesize} \defas \hess\ell_i(\MLEparm{\samplesize})$. 
Then, letting 
$\grad R_i\upper{\samplesize}(\pmt) \defas \grad\ell_i(\pmt) - \hat\grad\ell_i\upper{\samplesize}(\pmt)$, we can write 
the second-order smoothness constant as 
\[
    C_R\upper{\samplesize}
        &= \max_{i\in\range{\samplesize}} \max_{\pmt} \frac{\abs{\grad R_i\upper{\samplesize}(\pmt)} }{(\pmt-\MLEparm{\samplesize})^2}.
\]
\Cref{prop:statistical-weak-convergence-1} follows immediately from the two propositions below. 

\begin{proposition} \label{prop:L-bound}
    Recall that $L\upper{\samplesize} = \max \{1, \sigma_1, \dots, \sigma_\samplesize\}$.
    If loss functions $\ell_i$ are sampled i.i.d.
    and for some $c > 0$, it holds that $\EE[\norm{\hess\ell_i(\cdot)}_{L^\infty}^c ]<\infty$,
    then $L\upper{\samplesize} \leq \samplesize^{\frac{1}{c}}$ all but finitely often with probability 1.
\end{proposition}

\begin{proposition} \label{prop:3-tensor-moment-bound}
    If the $\ell_i$ are i.i.d.\ and there exists a positive constant $c_3$ such that $\EE[\norm{\grad^{\otimes 3}\ell_i(\cdot)}_{L^\infty}^{c_3} ]<\infty$, then 
    $C_R\upper{\samplesize} \leq \samplesize^{1/c_3}$ all but finitely often with probability 1.
\end{proposition}

The proofs of both propositions rely on the following technical lemma:
\begin{lemma}[Lemma~2 of \citet{negrea2022statistical}]
    \label{lem:as-maximal-ineq}
    Let $\alpha:\PosReals \to \PosReals$ be non-decreasing, right continuous with left limits, with $\alpha(0)=0$, and $\lim_{t\to\infty}\alpha_{t} = \infty$.
    Let $\mu$ be a probability measure on $\Reals$ such that if $Z\distas \mu$, then $\Pr(Z\geq 0)=1$ and $\EE\sbra*{\alpha(Z)}<\infty$.
    Let $Z_i \distas \mu$ for all $i\in\Nats$ (possibly not independent) and  
    let $\alpha^+: u \mapsto \inf\cbra*{t\geq 0 : \alpha_{t}\geq u}$ be the generalized inverse of $\alpha$.
    Then
    \[
        \Pr\cbra*{\max_{i \in \range{\samplesize}} Z_i \geq \alpha^{+}(n) \quad \textup{i.o.}} = 0.
    \]
\end{lemma}
\begin{proof}[Proof of \cref{prop:L-bound}]
The result follows immediately from \cref{lem:as-maximal-ineq}. 
\end{proof}
\begin{proof}[Proof of \cref{prop:3-tensor-moment-bound}]
Using the integral form of Taylor's remainder theorem,
    \[
        \grad \ell_i(\pmt) 
            & = \psi_i\upper{\samplesize} + \sigma_i\upper{\samplesize} (\pmt - \MLEparm{\samplesize}) + \int_{\MLEparm{\samplesize}}^\pmt \grad^{\otimes 3}\ell_i(\vartheta)(\pmt-\vartheta) \dee\vartheta.
    \]
    Thus, 
    \[
    \abs{\grad R_i\upper{\samplesize}(\pmt)} 
        & \leq \frac{\abs{\pmt-\MLEparm{\samplesize}}^2}{2} \norm{\grad^{\otimes 3}\ell_i(\cdot)}_{L^\infty}
    \]    
    and the result follows from \cref{lem:as-maximal-ineq}.
\end{proof}
\begin{remark}
The conditions for \cref{prop:L-bound,prop:3-tensor-moment-bound}  can be slightly relaxed to require $\grad\ell_i$ and $\hess\ell_i$ to be Lipschitz for each $i$, with a random Lipschitz constant that has finite $c_2$th and $c_3$th moment respectively.
\end{remark}

The following propositions, proved in \cref{proof_of_bdonfi} of the supplementary material, give the bounds for $\abs{\FI^{(\samplesize)}- \FI^{(\infty)}}$ and $\abs{\foFI^{(\samplesize)}- \foFI^{(\infty)}}$. 

\begin{proposition}
\label{prop:sigma-bound}
    If the loss functions $\ell_i$ are sampled i.i.d.\ such that
    \begin{enumerate}
        \item $v_4 = \EE[\abs{\hess\ell_1(\trueparm)-\FI^{(\infty)}}^4]<\infty$,
        \item $\EE[\norm{\grad^{\otimes 3}\ell_1(\cdot)}_{L^\infty}^c ]<\infty$,
        \item $\norm{\MLEparm{\samplesize}-\trueparm} \leq C_{\theta} \samplesize^{-b}$ all but finitely often with probability 1, and 
        \item $c>1/b$, 
    \end{enumerate}
    then  $\abs{\FI^{(\samplesize)}- \FI^{(\infty)}} \leq C_{\theta} \samplesize^{\frac{1}{c}-b}+\frac{\log(n)^2}{n^{1/4}}$ all but finitely often with probability 1. 
\end{proposition}
\begin{proposition}
\label{prop:omega-bound}
    If the loss functions $\ell_i$ are sampled i.i.d.\ such that
    \begin{enumerate}
        \item $\EE[\abs{\grad\ell_1(\trueparm)^{\otimes 2}-\foFI^{(\infty)}}^4]<\infty$,
        \item $\EE[\norm{\hess\ell_1(\cdot)}_{L^\infty}^d ]<\infty$, and
        \item $\norm{\MLEparm{\samplesize}-\trueparm} \leq C_{\theta} \samplesize^{-b}$ all but finitely often with probability 1,
    \end{enumerate}
    then $\abs{\foFI^{(\samplesize)}- \foFI^{(\infty)}}\leq 2 C_{\theta} n^{\frac{1}{8} +\frac{1}{d} -b} + C_{\theta}^2 n^{\frac{2}{d}-2b} + \frac{\log(n)^2}{n^{1/4}}$ all but finitely often with probability 1.
\end{proposition}

\subsection{Bounds for Common Metrics}\label{sec:common_metrics}
Next we will show how to use the ideas of \citet{barbour2024stein} to develop, as a corollary to our main result, bounds on standard and commonly used notions of distances between probability measures. 
The first distance we consider is the \emph{L\'evy--Prokhorov metric} \citep{zolotarev1994}. 
Let the function $\law{\cdot}$ return the \textit{law} of its argument. For $\lambda > 0$, a Skorokhod-measurable set $K\in\skspace$, and $\operatorname{dist}(w,K)\defas \inf\{\|w-v\|:v\in K\}$, define the \emph{$\lambda$-enlargement 
of $K$} as 
\[
    K^{\lambda}\defas \left\{w\,:\,\operatorname{dist}(w,K)<\lambda\right\}\supseteq K.
\]
The L\'evy--Prokhorov metric between the laws of $Y$ and $Z$ is then given by
\[\label{d_lp}
    d_\mathrm{LP}\left(\law{Y},\law{Z}\right)\defas \inf\left\{\epsilon>0\,:\, \Pr[Y\in K]\leq \Pr[Z\in K^{\epsilon}]+\epsilon \text{ $\forall$ Skorokhod measurable sets $K$}\right\}.
\]
On any metric space, convergence of a sequence of probability measures with respect to the L\'evy--Prokhorov distance implies weak convergence of that sequence; see, e.g. \citet[Section 6, page 72]{billingsley1999}. On separable metric spaces, it precisely metrizes the topology of weak convergence of probability measures, which has led to its widespread use.

The second distance we consider is  the \emph{bounded Wasserstein distance}, which is also known in the literature under several other names, including the \textit{Kantorovich-Rubinstein metric}, the \textit{Dudley metric} or the \textit{bounded Lipschitz distance}; see \citet[pages 12--13]{janson2001} and \citet{zolotarev1984}.
Let $\mathcal{H}$ be the set of all $\testfun :\skspace\to\R$ that are bounded, Skorokhod-measurable and Lipschitz with respect to the sup-norm $\|\cdot\|$, satisfying 
\[
\sup\left\{|\testfun(w)| :\,w\in\skspace\right\}\leq 1\quad \text{and}\quad \|\FD\testfun\|\leq 1.
\]
Then the bounded Wasserstein distance between the laws of $Y$ and $Z$ is given by
\[\label{d_bw}
    d_\mathrm{BW}\left(\law{Y},\law{Z}\right)\defas \sup_{\testfun \in\mathcal{H}}\left|\EE \testfun(Y) - \EE \testfun(Z)\right|.
\]
Convergence with respect to the bounded Wasserstein distance also implies weak convergence \citet[Section 8.3]{bogachev2007}. If the underlying metric space is separable and complete then the reverse implication also holds. Those properties and the fact that the bounded Wasserstein distance is an integral probability metric make it another common choice of metric. 

\begin{corollary}[Bound on the L\'evy--Prokhorov and bounded Wasserstein distances]\label{cor_levy_wasserstein}
    Retain the assumptions and setup of \Cref{thm:main-theorem-simple} and assume additionally that $\invtemp^{-1} \in O(1)$ as $\stepsize\to 0$. Let $r>1$ be such that $\EE\left|\psi_{I(1,1)}\right|^{4r}<\infty$ and suppose that \Cref{asmp:bounded-iterates} holds for $p=4r$. Assume that $C_R\geq 1$ and that $\frac{(L^2 + C_R^2)\stepsize}{b} \vee \frac{L^6 + C_R^2}{\invtemp} \in O(1)$. 
    Then, in the numerical setting, for $\stepsize\to 0$,  
    \[
    d_\mathrm{LP}\left(\law{Y},\law{Z}\right) &\in O\left(\stepsize^{\frac{1}{20}-\frac{9}{200r-20}}\left\{\rbra*{C_RL^7+C_R^3}\rbra*{\sqrt{\log(\stepsize^{-1})} + \stepsize^{-\frac{1}{2}}\invtemp^{-\frac{1}{2}}}+L^{2r}\right\}\right) \\
    \andT \\
    d_\mathrm{BW}\left(\law{Y},\law{Z}\right) &\in O\left(\stepsize^{\frac{1}{14}-\frac{2}{49r-21}}\left\{\rbra*{C_RL^7+C_R^3}\rbra*{\sqrt{\log(\stepsize^{-1})} + \stepsize^{-\frac{1}{2}}\invtemp^{-\frac{1}{2}}}+L^{2r}\right\}\right). 
    \]
\end{corollary}

The proof can be found in \cref{sec:proof_levy_wass} of the supplementary material.

\begin{remark}
    Suppose that the assumptions of \Cref{thm:main-theorem-simple} hold. 
    Assume additionally that $n/b \to \infty$ and $m\in O(1)$ for $n\to\infty$, and that $r>1$ is such that $\EE\left|\Psi_X\right|^{4r}<\infty$. Moreover, suppose that \cref{asmp:bounded-iterates} also holds for $p=4r$. Assume that $C_R\geq 1$ and that $ \frac{L^6 + C_R^2}{n} \in O(1)$. Then \Cref{cor_levy_wasserstein} implies that, in the statistical setting, for $n \to \infty$, 
    \[
    d_\mathrm{LP}\left(\law{Y},\law{Z}\right) &\in O\left(\left(\frac{b}{n}\right)^{\frac{1}{20}-\frac{9}{200r-20}}\left\{\rbra*{C_RL^7+C_R^3}\log^{1/2}\left(\frac{n}{b}\right) + L^{2r}\right\} \right)\\
        \andT \\
    d_\mathrm{BW}\left(\law{Y},\law{Z}\right) &\in O\left(\left(\frac{b}{n}\right)^{\frac{1}{14}-\frac{2}{49r-21}}\left\{\rbra*{C_RL^7+C_R^3}\log^{1/2}\left(\frac{n}{b}\right)+L^{2r}\right\}\right).
    \]
\end{remark}

\section{A More General Functional Bound for Stochastic Iterative Algorithms} \label{sec:main-theorem-proof}
Our aim in this section is to derive a bound analogous to the bounds of \Cref{thm:main-theorem-simple} yet holding under significantly weaker assumptions. We retain the setup of \cref{sec:approach} and \Cref{asmp:bounded-iterates,asmp:curvature,asmp:hosmooth,asmp:stepsize} but we do not focus on the numerical or the statistical setup. 
Our bound will be derived through a series of lemmas presented below. We present a simplified version of our bound in \cref{thm:general_simplified}, stated in \cref{sec:simplified_bound}, and our most general bound in \cref{MainThm} stated in \cref{app:additional proofs} of the supplementary material.

For the rest of this section, the notation ``$\lesssim$'' will mean ``less or equal up to a numerical constant'', i.e., ``$\leq\, C\, \times$'', where $C$ is a numerical constant independent of $\timescale, \stepsize, \invtemp, L, \Omega, \Sigma, \spacescale$, and the moments of $\Gradbatch{1}{1}$.

\subsection{Initial Bound Decomposition}
Recall from \cref{sec:comparison_exchangeable} that to bound $\abs{\EE g(Y)-\nobreak\EE g(Z)}$, for $g\in M$, we use the decomposition
\[
\abs{\EE g(Y)-\EE g(Z)} 
&\le \abs{\EE g(Y)-\EE g(\newy)}+\abs{\EE g(\newy)-\EE g(\newz)}+\abs{\EE g(\newz)-\EE g(Z)}.
\]
The following two lemmas give the bound for 
$\abs{\EE g(Y)-\EE g(\newy)}$ and $\abs{\EE g(Z)-\EE g(\newz)}$.
The proofs are given in \cref{app:diffY,app:diffZ} of the supplementary material.
\begin{lemma} For $Y$ and $\newy$ defined in \cref{eq:SGLD-process,formulanewy}, if \cref{asmp:hosmooth,asmp:bounded-iterates} hold, then 
\label{BDforR}
\[
&\abs{\EE \bar g(\newy) -\EE \bar g(Y)}\\ &\lesssim \Bigg\{w\timescale K_1C_R\rbra*{\frac{h^2\Omega}{b}+\frac{h}{\invtemp}} \\
&\qquad\quad+ w^3 \timescale^3 C_R^3 K_3^3 \rbra*{\frac{h^6}{b^3}\rbra*{\Omega^3+(\EE \Gradbatch{1}{1}^4)^{3/2}+\EE \Gradbatch{1}{1}^4\Omega+\EE\Gradbatch{1}{1}^6} +\frac{h^3}{\invtemp^3}}\\
&\quad +w^3\Bigg\{\frac{\timescale^2 \stepsize^4}{\batchsize^2}\sbra*{\frac{\timescale\stepsize^2L^2}{b}E_1+\timescale^2\stepsize^4L^4E_1+\frac{\EE \Gradbatch{1}{1}^4}{b}+\Omega^2+\Omega}\\
&\qquad+\frac{\timescale^2\stepsize^2}{\invtemp^2}\sbra*{1+\frac{\timescale\stepsize^2L^2}{\batchsize}+\timescale^2\stepsize^4L^4}+\rbra*{\frac{1}{\batchsize}+\timescale \stepsize^2L^2}\sqrt{\Omega}(1+L^3) \frac{\timescale^3h^{11/2}L^2}{\invtemp^{3/2}}\Bigg\}^{3/4} \\
&\qquad  \times w\timescale C_R K_3 \sbra*{\frac{h^6}{b^3}\rbra*{\Omega^3+(\EE \Gradbatch{1}{1}^4)^{3/2}+\EE \Gradbatch{1}{1}^4\Omega+\EE\Gradbatch{1}{1}^6} +\frac{h^3}{\invtemp^3}}^{1/3}\Bigg\},
\]
where
\[E_1=\EE \Gradbatch{1}{1}^4+\Omega^2+\sqrt{\Omega}(1+L^3)\sbra*{\rbra*{\EE \Gradbatch{1}{1}^4+\Omega^2}^{3/4}+1}+\Omega L(1+L^4)\sbra*{\Omega\frac{\timescale h^2}{b}+\frac{\timescale h}{\invtemp}},
\]
as in \cref{def:e1} below.
\end{lemma}

\begin{lemma}\label{BDforZ} For $Z$ and $\newz$ defined in \cref{formulaz,formulanewz}, 
\[&|\EE g(Z)-\EE g(\newz)|\\
\lesssim &\|g\|_M\Bigg\{\frac{\spacescale\timescale^{1/2}\stepsize^2\Omega^{1/2}\Sigma}{\batchsize^{1/2}}+\frac{\spacescale\timescale^{1/2}\stepsize^{3/2}\Sigma}{\invtemp^{1/2}}+\frac{\spacescale\stepsize\Omega^{1/2}\sqrt{\log(2\timescale)}}{\batchsize^{1/2}}+\frac{\spacescale\stepsize^{1/2}\sqrt{\log(2\timescale)}}{\invtemp^{1/2}}\\
&+\frac{\spacescale\timescale^{3/2}\stepsize^6\Omega^{3/2}\Sigma^3}{\batchsize^{3/2}}+\frac{\spacescale^3\timescale^{3/2}\stepsize^{9/2}\Sigma^3}{\invtemp^{3/2}}+\frac{\spacescale^3\stepsize^3\Omega^{3/2}\log^{3/2}(2\timescale)}{\batchsize^{3/2}}+\frac{\spacescale^3\stepsize^{3/2}\log^{3/2}(2\timescale)}{\invtemp^{3/2}}\\
&+\frac{\spacescale^4\timescale^{2}\stepsize^{5}\Omega^2\Sigma}{\batchsize^2}+\frac{\spacescale^4\timescale^{3/2}\stepsize^{4}\Omega^2\sqrt{\log(2\timescale)}}{\batchsize^2}+\frac{\spacescale^4\timescale^{2}\stepsize^{3}\Sigma}{\invtemp^2}+\frac{\spacescale^4\timescale^{3/2}\stepsize^{2}\sqrt{\log(2\timescale)}}{\invtemp^2}\Bigg\}.\]
\end{lemma}

In order to bound $|\EE g(\newz)-\EE g(\newy)|$ we apply \cref{ThmM} (based on Stein's method) to get 
\[ \label{eq:steins-method-error-decomp}
|\EE g(\newz)-\EE g(\newy)| \le \error{exch}+\error{cov}+\error{rem},
\]
where 
\begin{align}
&\error{rem}=\abs{\EE\cbra*{\FD f(\newy)[\newy]-2\EE^{\newy}\cbra*{\FD f(\newy)[(\newy-\exy)\Lambda]}}},\\
&\error{exch}=\frac{\|g\|_M\Lambda }{6} \EE\|\newy-\exy\|^3,\\
&\error{cov}=\abs{\EE\cbra*{\FD^2 f(\newy)[(\newy-\exy)\Lambda,\newy-\exy]}-\EE \FD^2 f(\newy)[\newz,\newz]}.
\end{align}
After some preliminary calculations in \cref{sec:prelim-calculations}, we bound each error term in turn in \cref{sec:error-term-bounds}.
The final bound is given in \cref{sec:simplified_bound}.

\subsection{Preliminary Calculations} \label{sec:prelim-calculations}

We will start by analyzing the behavior of SG(L)D iterates $\newpmt_k$, which we will then use to bound the error terms in \cref{eq:steins-method-error-decomp}.
First, the following lemma gives the explicit formula for $\newpmt_k$, which we write in terms of the quantities
\[\label{eq:Q_def}
Q(j,k)= 
\begin{cases}
    \prod_{m=j}^{k-1}\rbra*{1-\frac{\stepsize}{b}\sum_{i=1}^{b}\Heisbatch{m}{i}} & \text{if $j <k$} \\
    1 & \text{if $j\ge k$}.
\end{cases}
\] 
Note that $0\le Q(j, k) \le 1$, by \cref{asmp:curvature,asmp:stepsize}.
\begin{lemma}\label{lm:pmt}
For $\cbra*{\newpmt_k}_{k=1}^{\timescale}$ defined by \cref{sgdnewpmt}, assuming $\newpmt_0=0$, we have
\[
\newpmt_k= \sum_{j=0}^{k-1} Q(j+1,k)\rbra*{\frac{h}{b}\sum_{i=1}^{b}\Gradbatch{j}{i}+\sqrt{\frac{2h}{\invtemp}}\noise_{j}}.
\]
\end{lemma}
The proof of \cref{lm:pmt} is a straightforward induction, so we defer it to \cref{app:proof_lemma2} in the supplementary material.
Under \cref{asmp:curvature,asmp:stepsize}, the following three lemmas give the bound for moments of $\eta_k$ and moments of the sup-norms of $\eta_k$.
These bounds are useful in bounding the error terms $\error{exch},\error{cov},\error{rem}$ arising from Stein's method.
\begin{lemma}\label{BDforM} For $\iter =1,\cdots,\timescale$, if \cref{asmp:curvature,asmp:stepsize} hold,
\[\EE\newpmt_k^2 \le \timescale \rbra*{\frac{h^2 \Omega}{b} + \frac{2h}{\invtemp}}\] and 
\[\EE\newpmt_k^4 \lesssim & \frac{\timescale^2h^4}{b^2} \Bigg\{\EE \Gradbatch{1}{1}^4+\Omega^2+\sqrt{\Omega}(1+L^3)\sbra*{\rbra*{\EE \Gradbatch{1}{1}^4+\Omega^2}^{3/4}+1}\\
&+\Omega L(1+L^4)\sbra*{\Omega\frac{\timescale h^2}{b}+\frac{\timescale h}{\invtemp}}\Bigg\}+\frac{\timescale^2h^2}{\invtemp^2}+\sqrt{\Omega}(1+L^3) \frac{\timescale^2h^{7/2}}{\invtemp^{3/2}}.\]
\end{lemma}

\begin{lemma}\label{BDforSM} For $\iter =1,\dots,\timescale$, if \cref{asmp:curvature,asmp:stepsize} hold, then

\[
\EE\sup_{k\in\{1,\cdots,\timescale\}}\newpmt_k^2
\lesssim &\frac{\timescale \stepsize^2}{\batchsize}\sbra*{\frac{E_1^{1/2}\timescale^{1/2}\stepsize L}{\batchsize^{1/2}}+\sqrt{\Omega}L(1+L^3)\frac{\timescale^{1/2}\stepsize^{3/4}}{\invtemp^{3/4}}+L^2\Omega\timescale\stepsize^2+\Omega}\\
&+\frac{\timescale\stepsize}{\invtemp}\sbra*{\frac{\timescale^{1/2}\stepsize L}{\batchsize^{1/2}}+1+L^2\timescale\stepsize^2}\\
\EE\sup_{k\in\{1,\cdots,\timescale\}}\newpmt_k^4
\lesssim & \frac{\timescale^2 \stepsize^4}{\batchsize^2}\sbra*{\frac{\timescale\stepsize^2L^2}{b}E_1+\timescale^2\stepsize^4L^4E_1+\frac{\EE \Gradbatch{1}{1}^4}{b}+\Omega^2+\Omega}\\
&+\frac{\timescale^2\stepsize^2}{\invtemp^2}\sbra*{1+\frac{\timescale\stepsize^2L^2}{\batchsize}+\timescale^2\stepsize^4L^4}+\rbra*{\frac{1}{\batchsize}+\timescale \stepsize^2L^2}\sqrt{\Omega}(1+L^3) \frac{\timescale^3h^{11/2}L^2}{\invtemp^{3/2}},
\]
where 
\[E_1=\EE \Gradbatch{1}{1}^4+\Omega^2+\sqrt{\Omega}(1+L^3)\sbra*{\rbra*{\EE \Gradbatch{1}{1}^4+\Omega^2}^{3/4}+1}+\Omega L(1+L^4)\sbra*{\Omega\frac{\timescale h^2}{b}+\frac{\timescale h}{\invtemp}}.\label{def:e1}
\]
\end{lemma}
The proofs are deferred to, respectively, \cref{app:newpmt,app:newpmt1} in the supplementary material.

\subsection{Bounding the Error Terms from Stein's Method} \label{sec:error-term-bounds}

We can now turn to bounding the error terms $\error{exch},\error{cov},\error{rem}$ arising from our application of Stein's method. 
Recall the exchangeable pair $\rbra*{\newy,\exy}$ we constructed in \cref{sec:comparison_exchangeable} and take $\Lambda = \timescale/2$. 
Since all the three errors depend on $\newy-\exy$, we give an explicit formula (see \cref{app:proof_lemma7} in the supplementary material for a proof). 

\begin{lemma}\label{lm:difY} Recall the notation of \cref{sec:comparison_exchangeable} and \cref{eq:Q_def}.
For $t\in [0,1]$, it holds that
\[\label{difY}
&\newyt-\exyt = w\sum_{m=K+1}^{\alpha}\cbra*{Q(K+1,m)(\newpmt_{K+1}-\exparm_{K+1})}\mathbb{I}_{[\frac{m}{\alpha
},\frac{m+1}{\alpha})}(t),
\]
\text{where}\\
\[\label{difT}
\qquad\newpmt_{K+1}-\exparm_{K+1} 
    & = \frac{h}{b}\rbra*{\rbra*{-\sum_{i=1}^{b}\Heisbatch{K}{i} +\sum_{i=1}^{b}\swapHeisbatch{K}{i} }\newpmt_K + \sum_{i=1}^{b}\Gradbatch{K}{i} -\sum_{i=1}^{b} \swapGradbatch{K}{i}}\\
    &\hspace{3cm}+\sqrt{\frac{2h}{\invtemp}}\rbra*{\noise_{K}-\xi'_{K}}.
\]
\end{lemma}

We first bound the remainder term 
\[\error{rem}=
\abs{\EE\cbra*{\FD f(\newy)[\newy]-2\EE^{\newy}\cbra*{\FD f(\newy)[(\newy-\exy)\Lambda]}}}.
\]
\begin{lemma}\label{BDrem}
The following bound on $\error{rem}$ holds:
\[
\error{rem}\lesssim \frac{D_1 \timescale \spacescale\stepsize L \|g\|_M}{b^{1/2}}\sqrt{\rbra*{1+\timescale^2 \stepsize^2 L^2}\rbra*{\frac{h^2 \Omega}{b} + \frac{h}{\invtemp}}},
\]
where
\[
D_1^2 &= 1 +w^4\frac{\timescale^2 \stepsize^4}{\batchsize^2}\sbra*{\frac{\timescale\stepsize^2L^2}{b}E_1+\timescale^2\stepsize^4L^4E_1+\frac{\EE \Gradbatch{1}{1}^4}{b}+\Omega^2+\Omega}\\
&\phantom{=~}+w^4\frac{\timescale^2\stepsize^2}{\invtemp^2}\sbra*{1+\frac{\timescale\stepsize^2L^2}{\batchsize}+\timescale^2\stepsize^4L^4} 
+w^4\rbra*{\frac{1}{\batchsize}+\timescale \stepsize^2L^2}\sqrt{\Omega}(1+L^3) \frac{\timescale^3h^{11/2}L^2}{\invtemp^{3/2}}\\
&\phantom{=~} + \frac{1}{\FI^2 }\rbra*{\frac{w^2 h}{b} \Omega  + \frac{ w^2 }{\invtemp}}^2\log^2(1+\timescale\stepsize \FI )
\]
and $E_1$ is given in \cref{def:e1}.
\end{lemma}
\begin{proof}
Using \cref{lm:difY}, by exchanging the order of summation,
\[
	&\hspace{-1em} \EE\cbra*{ 2\EE^{\newy}\cbra*{\FD f(\newy)[(\newy-\exy)\Lambda]}}\\
		&=2\EE\cbra*{\frac{\timescale}{2} \FD f(\newy)\sbra*{\frac{1}{\alpha}\sum_{k=0}^{\alpha-1}\frac{\spacescale\stepsize}{b} \sum_{m=k+1}^{\timescale}\left\{ Q(k+1,m)\rbra*{b\FI -\sum_{i=1}^{b}\Heisbatch{k}{i}}\newpmt_k\right\}\ind{\cointer{\frac{m}{\timescale},\frac{m+1}{\timescale}}}(t)}}\\
   &\quad + 2\EE\cbra*{\frac{\timescale}{2} \FD f(\newy)\sbra*{\frac{1}{\alpha}\sum_{k=0}^{\alpha-1}\frac{\spacescale\stepsize}{b} \sum_{m=k+1}^{\timescale}\left\{ Q(k+1,m)\sum_{i=1}^{b}\Gradbatch{k}{i}\right\}\ind{\cointer{\frac{m}{\timescale},\frac{m+1}{\timescale}}}(t)}}\\
   &\quad + 2\EE\cbra*{\frac{\timescale}{2} \FD f(\newy)\sbra*{\frac{1}{\alpha}\sum_{k=0}^{\alpha-1}\spacescale\sqrt{\frac{2\stepsize}{\invtemp}} \sum_{m=k+1}^{\timescale}\left\{ Q(k+1,m)\noise_{k}\right\}\ind{\cointer{\frac{m}{\timescale},\frac{m+1}{\timescale}}}(t)}}\\
   & = \underbrace{\EE\cbra*{\FD f(\newy)\sbra*{\sum_{m=1}^{\timescale}\spacescale\cbra*{\left(\sum_{k=1}^{m-1} Q(k+1,m)\rbra*{\frac{\stepsize}{b} \sum_{i=1}^{b}\Gradbatch{k}{i}+\sqrt{\frac{2\stepsize}{\invtemp}}\noise_{k}} \right)}\ind{\cointer{\frac{m}{\timescale},\frac{m+1}{\timescale}}}(t)}}}_{\EE \FD f(\newy)[\newy]}\\
   &\quad  + \EE\cbra*{\FD f(\newy)\sbra*{\frac{\spacescale\stepsize}{b}\sum_{m=1}^{\timescale}\cbra*{\sum_{k=0}^{m-1}Q(k+1,m)\rbra*{b\FI  -\sum_{i=1}^{b}\Heisbatch{k}{i}}\newpmt_k} \ind{\cointer{\frac{m}{\timescale},\frac{m+1}{\timescale}}}(t)}}.
\]
Thus using the fact that $\FD f(\newy)[f] \le \|\FD f(\newy)\||\sup f|$, the Cauchy-Schwarz inequality, \cref{Mart,BDforM} (the latter of which is deferred to the supplementary material due to its technicality) and \cref{ThmM},
\[ \label{error1H}
\error{rem}
&= \left|\EE\cbra*{\FD f(\newy)\sbra*{\frac{\spacescale\stepsize}{b}\sum_{m=1}^{\timescale}\cbra*{\sum_{k=0}^{m-1}Q(k+1,m)\rbra*{b\FI  -\sum_{i=1}^{b}\Heisbatch{k}{i}}\newpmt_k} \ind{\cointer{\frac{m}{\timescale},\frac{m+1}{\timescale}}}(t)}}\right|\\
&\le \frac{\spacescale\stepsize}{b}\left|\EE\cbra*{\|\FD f(\newy)\|\sup_{m\in\{1,\cdots,\timescale\}}\sum_{k=0}^{m-1}Q(k+1,m)\rbra*{b\FI  -\sum_{i=1}^{b}\Heisbatch{k}{i}}\newpmt_k}\right|\\
&\le \frac{\spacescale\stepsize}{b}\sqrt{\EE\|\FD f(\newy)\|^2 }\sqrt{\EE\sup_{m\in\{1,\cdots,\timescale\}}\rbra*{\sum_{k=0}^{m-1}Q(k+1,m)\rbra*{b\FI  -\sum_{i=1}^{b}\Heisbatch{k}{i}}\newpmt_k}^2}\\
&\stackrel{(a)}\le \frac{\spacescale\stepsize}{b}\sqrt{\EE\|\FD f(\newy)\|^2}\sqrt{bL^2\sum_{j=0}^{\timescale}\EE\newpmt_j^2+\timescale \stepsize^2 b L^4  \sum_{j=0}^{\timescale}\sum_{k=0}^{j-1}  \EE\newpmt_k^2}\\
&\stackrel{(b)}\le \frac{\timescale \spacescale\stepsize L \|g\|_M}{b^{1/2}}\sqrt{\rbra*{1+\timescale^2 \stepsize^2 L^2}\rbra*{\frac{h^2 \Omega}{b} + \frac{2h}{\invtemp}}}\sqrt{\EE\rbra*{1+\frac{2}{3}\|\newy\|^2 +\frac{4}{3}\EE\|\newz\|^2}^2}\\
&\le \frac{D_1 \timescale \spacescale\stepsize L \|g\|_M}{b^{1/2}}\sqrt{\rbra*{1+\timescale^2 \stepsize^2 L^2}\rbra*{\frac{h^2 \Omega}{b} + \frac{2h}{\invtemp}}},
\]
where $(a)$ follows from \cref{Mart} in the supplementary material, $(b)$ follows from \cref{ThmM,BDforM},
and $D_1$ is a bound for $\sqrt{\EE\rbra*{1+\frac{2}{3}\|\newy\|^2 +\frac{4}{3}\EE\|\newz\|^2}^2}$ derived in \cref{D1D2} in the supplementary material. 
\end{proof}

Next, we bound the exchange error term 
\[
\error{exch}=\frac{\|g\|_M}{6}\Lambda \EE\|\newy-\exy\|^3.
\]
\begin{lemma}\label{BDexch}
The following bound on $\error{exch}$ holds:
\[
    \error{exch} &\lesssim  \|g\|_M\spacescale^3  \Bigg\{L^3 \frac{\timescale^{5/2}h^6}{b^{3/2}} \Bigg[\EE \Gradbatch{1}{1}^4+\Omega^2+\sqrt{\Omega}(1+L^3)\sbra*{\rbra*{\EE \Gradbatch{1}{1}^4+\Omega^2}^{3/4}+1}\\
    &\phantom{\lesssim  \|g\|_M\spacescale^3  \Bigg\{} +\Omega L(1+L^4)\sbra*{\Omega\frac{\timescale h^2}{b}+\frac{\timescale h}{\invtemp}}\Bigg]^{3/4}+\frac{L^3\timescale^{5/2}h^{9/2}}{\invtemp^{3/2}} \\
    &\phantom{\lesssim  \|g\|_M\spacescale^3  \Bigg\{} +\sqrt{\Omega}L^3(1+L^{9/4}) \frac{\timescale^{5/2}h^{45/8}}{\invtemp^{9/8}}+\frac{\timescale\stepsize^3}{b^{3/2}} (\EE \Gradbatch{1}{1}^4+\Omega^2)^{3/4}+\frac{\timescale h^{3/2}}{\invtemp^{3/2}}\Bigg\}.
    \]
\end{lemma}
\begin{proof}
From \cref{difY,difT}, using the fact $|Q(i,j)|\le 1$ (from \cref{asmp:stepsize} and \cref{asmp:curvature}) and \cref{BDforM},
\[
\error{exch} 
& = \frac{\|g\|_M \timescale \spacescale^3}{12}\frac{1}{\timescale}\sum_{k=0}^{\timescale-1}\EE\Bigg\{\sup_{m\in\{k+1,\cdots,\timescale\}}\Bigg|Q(k+1,m)(\newpmt_{k+1}-\exparm_{k+1})\Bigg|^3\Bigg\}\\
& \le \frac{\|g\|_M \timescale \spacescale^3}{12}\frac{1}{\timescale}\sum_{k=0}^{\timescale-1}\EE\Bigg|\newpmt_{k+1}-\exparm_{k+1}\Bigg|^3\\
&=\frac{\|g\|_M \timescale \spacescale^3}{12}\frac{1}{\timescale}\sum_{k=0}^{\timescale-1}\EE \Bigg|\frac{h}{b}\rbra*{\rbra*{-\sum_{i=1}^{b}\Heisbatch{k}{i} +\sum_{i=1}^{b}\swapHeisbatch{k}{i} }\newpmt_k + \sum_{i=1}^{b}\Gradbatch{k}{i} -\sum_{i=1}^{b} \swapGradbatch{k}{i}}\\
&\hspace{3.5cm}+\sqrt{\frac{2h}{\invtemp}}\rbra*{\noise_{k}-\xi'_{k}}\Bigg|^3\\
&\le \frac{\|g\|_M \timescale \spacescale^3}{12}\frac{1}{\timescale}\sum_{k=0}^{\timescale-1}\Bigg(\frac{72\stepsize^3}{b^3}\EE \Bigg|\sum_{i=1}^{b}\Heisbatch{k}{i}\Bigg|^3\EE|\newpmt_k|^3 + \frac{72\stepsize^3}{b^3}\EE \Bigg|\sum_{i=1}^{b}\Gradbatch{k}{i}\Bigg|^3 \\
&\hspace{3.5cm}+\frac{144\sqrt{2}\stepsize^{3/2}}{\invtemp^{3/2}}\EE|\noise_k|^3\Bigg)\\
&\le 6\|g\|_M \spacescale^3\rbra*{\stepsize^3 L^3 \sum_{k=0}^{\timescale-1} (\EE\newpmt_k^4)^{3/4}+\timescale\stepsize^3 b^{-3/2} (\EE \Gradbatch{1}{1}^4+3\Omega^2)^{3/4}+\frac{8\timescale h^{3/2}}{\invtemp^{3/2}}\sqrt{1/\pi}}\\
&\lesssim \|g\|_M\spacescale^3  \Bigg\{L^3 \frac{\timescale^{5/2}h^6}{b^{3/2}} \Bigg[\EE \Gradbatch{1}{1}^4+\Omega^2+\sqrt{\Omega}(1+L^3)\sbra*{\rbra*{\EE \Gradbatch{1}{1}^4+\Omega^2}^{3/4}+1}\\
&\phantom{\lesssim \|g\|_M\spacescale^3  \Bigg\{} +\Omega L(1+L^4)\sbra*{\Omega\frac{\timescale h^2}{b}+\frac{\timescale h}{\invtemp}}\Bigg]^{3/4}{\invtemp^{3/2}}+\sqrt{\Omega}L^3(1+L^{9/4}) \frac{\timescale^{5/2}h^{45/8}}{\invtemp^{9/8}}\\
&\phantom{\lesssim \|g\|_M\spacescale^3  \Bigg\{} +\frac{L^3\timescale^{5/2}h^{9/2}} +\frac{\timescale\stepsize^3}{b^{3/2}} (\EE \Gradbatch{1}{1}^4+\Omega^2)^{3/4}+\frac{\timescale h^{3/2}}{\invtemp^{3/2}}\Bigg\}.\]
\end{proof}

Finally we bound the covariance error term 
\[
		\error{cov}
			& =\abs{\EE\cbra*{\FD^2 f(\newy)[(\newy-\exy)\Lambda,\newy-\exy]}-\EE \FD^2 f(\newy)[\newz,\newz]},
	\]
where 
\[
\EE \FD^2 f(\newy)[\newz,\newz]=& \frac{ \frac{w^2 h}{b} \Omega + \frac{2w^2}{\invtemp}}{2\FI}\sum_{m,r=1}^{\timescale}\rbra*{e^{-\stepsize\FI  |m-r|}-e^{-\stepsize\FI  (m+r)}}\\
&\times\EE \FD^2 f(\newy)\sbra*{\mathbb{I}_{[\frac{m}{\alpha},\frac{m+1}{\alpha})}(t),\mathbb{I}_{[\frac{r}{\alpha},\frac{r+1}{\alpha})}(t)}.
\]
We have the following lemma, which we shall prove in \cref{app:proof_lemma10} of the supplementary material:
\begin{lemma}\label{lemma_error_cov}
We have the following bound for $\error{cov}$:
\[
    \error{cov}
    &\lesssim D_2\timescale\spacescale^2\stepsize^2\|g\|_M \\
    &\phantom{\lesssim} \times \Bigg\{\frac{\timescale\stepsize^2L^2}{b} \Bigg[\EE \Gradbatch{1}{1}^4+\Omega^2+\sqrt{\Omega}L^3\sbra*{\rbra*{\EE \Gradbatch{1}{1}^4+\Omega^2}^{3/4}+1}\Bigg]^{1/2} +\frac{\Sigma}{\invtemp}\\
    &\phantom{\lesssim \times \Bigg\{} +\frac{L^{9/2}\timescale^{3/2}\stepsize^3\Omega }{b^{3/2}}+\frac{L^{9/2}\timescale^{3/2}\stepsize^{5/2}\Omega^{1/2} }{b\invtemp^{1/2}}+\frac{\timescale\stepsize L^2}{\invtemp}+\frac{L^{7/2}\Omega^{1/4}\timescale \stepsize^{7/4}}{\invtemp^{3/4}}+\frac{\timescale^{1/2}\stepsize\Omega L}{\batchsize}\\
    &\phantom{\lesssim \times \Bigg\{} +\frac{\timescale^{1/2}\stepsize^{1/2}\sqrt{\Omega}L}{\batchsize^{1/2}\invtemp^{1/2}}+\frac{\spacescale\stepsize}{\batchsize^{3/2}}\rbra*{\rbra*{\EE \Gradbatch{1}{1}^4}^{3/4} + \Omega^{3/2}}+\frac{L\spacescale\stepsize^2\timescale  \Omega^{3/2}}{\batchsize^{3/2}}+\frac{L\spacescale\stepsize^{3/2}\timescale  \Omega}{\batchsize\invtemp^{1/2}} \\
    &\phantom{\lesssim \times \Bigg\{} +\frac{L\timescale^{1/2}\stepsize}{\batchsize^{3/2}}+\frac{L^2\timescale\stepsize^{2}}{\batchsize^{3/2}}+\frac{\stepsize\Omega\Sigma}{\batchsize}+\frac{\timescale\stepsize^2\Omega\Sigma^2}{\batchsize}+\frac{L\timescale^{1/2}\stepsize^{1/2}\Omega^{1/2}}{\batchsize\invtemp^{1/2}}+\frac{L\timescale^{1/2}}{\batchsize^{1/2}\invtemp}+\frac{\spacescale\stepsize^{1/2}\Omega}{\batchsize\invtemp^{1/2}}\\
    &\phantom{\lesssim \times \Bigg\{} +\frac{\spacescale\Omega^{1/2}}{\batchsize^{1/2}\invtemp}+\frac{L\spacescale\timescale\stepsize^{3/2}\Omega}{\batchsize\invtemp^{1/2}}+\frac{L\spacescale\timescale\stepsize\Omega^{1/2}}{\batchsize^{1/2}\invtemp}+\frac{\spacescale}{\stepsize^{1/2}\invtemp^{3/2}}+\frac{L\timescale^{1/2}}{\batchsize^{1/2}\invtemp}+\frac{L^2\timescale\stepsize}{\batchsize^{1/2}\invtemp}+\frac{\timescale\stepsize\Sigma^2}{\invtemp}\Bigg\},
\]
where
\[
    D_2 &= \Bigg\{1+w^2\frac{\timescale \stepsize^2}{\batchsize}\sbra*{\sqrt{\Omega}L(1+L^3)\frac{\timescale^{1/2}\stepsize^{3/4}}{\invtemp^{3/4}}+L^2\Omega\timescale\stepsize^2+\Omega} \\
    &\phantom{= \Bigg\{} +w^2\frac{\timescale\stepsize}{\invtemp}\sbra*{\frac{\timescale^{1/2}\stepsize L}{\batchsize^{1/2}}+1+L^2\timescale\stepsize^2} +w^2 \rbra*{\frac{\Omega\stepsize^2} {b}+\frac{\stepsize}{\invtemp}}\rbra*{1+\frac{\timescale^2 h^2 L^2}{(1-L\stepsize)^2}}\\
    &\phantom{= \Bigg\{} +w^2\frac{\timescale^{3/2}\stepsize^3L}{\batchsize^{3/2}}\sqrt{E_1} +\frac{1}{\FI ^2}\rbra*{\frac{w^2 h}{b} \Omega  + \frac{ w^2 }{\invtemp}}^2\log ^2(1+\timescale\stepsize \FI )\Bigg\}^{1/2}
\]
and $E_1$ is given in \cref{def:e1}.
\end{lemma}

\begin{remark}
    Among \cref{BDrem,BDexch,lemma_error_cov}, \cref{lemma_error_cov} is the most challenging to prove and most technically involved. In its proof, presented in \cref{app:proof_lemma10} of the supplementary material, we decompose $\error{cov}$ into several different terms, which we bound separately. One of those terms is given by
\[   &\error{cov.\Psi\Psi-z_G.1}:=\EE\Bigg\{\frac{1}{2\timescale}\sum_{k=0}^{\timescale -1}\frac{\timescale \spacescale^2 \stepsize^2}{b^2} \Bigg(\bigg(\sum_{i=1}^{b}\Gradbatch{k}{i}\bigg)^2 -b\Omega \Bigg)\\
&\quad  \times\FD^2 f(\newy)\Bigg[\sum_{m=k+1}^{\timescale}Q(k+1,m)\mathbb{I}_{[\frac{m}{\timescale
},\frac{m+1}{\timescale})}(t),\sum_{r=k+1}^{\timescale}Q(k+1,r)\mathbb{I}_{[\frac{r}{\timescale
},\frac{r+1}{\timescale})}(t)\Bigg]\Bigg\}.\label{eq:complicated_term} \]
Terms of a similar form arise frequently in applications of Stein's method of exchangeable pairs \citep{kasprzak2020exchangeable,dobler2021stein}. When bounding \cref{eq:complicated_term} in \cref{8.3} of the supplementary material, we introduce the modified process
\[
\newyt^{k\Psi} \defas \newyt&-\spacescale\sum_{m=k+1}^{\timescale}Q(k+1,m)\rbra*{\frac{\stepsize}{b}\sum_{i=1}^{b}\Gradbatch{k}{i}}\mathbb{I}_{[\frac{m}{\timescale
},\frac{m+1}{\timescale})}(t) \\
&\hspace{-6mm}-w\frac{-\stepsize\sum_{i=1}^{b}\Heisbatch{k}{i}}{\batchsize - \stepsize\sum_{i=1}^{b}\Heisbatch{k}{i}}\sum_{m=k+1}^{\timescale}\sum_{j=0}^{k-1}Q(j+1,m)\rbra*{\frac{h}{b}\sum_{i=1}^{b}\Gradbatch{j}{i}+\sqrt{\frac{2h}{\invtemp}}\noise_{j}}\ind{\cointer{\frac{m}{\samplesize},\frac{m+1}{\samplesize}}}(t),
\label{eq:newyt_subs}\]
with $Q$ defined in \cref{eq:Q_def}. 
Note that, for any $k=0,\dots,\alpha-1$, $\newyt^{k\Psi}$ and $\left(\Gradbatch{k}{i}\right)_{i=1}^b$ are independent (to see this, recall \cref{formulanewy,lm:pmt}). 
We then write  $\FD^2 f(\newy)$ in \cref{eq:complicated_term} as $\FD^2 f(\newy)-\FD^2 f(\newyt^{k\Psi}) + \FD^2 f(\newyt^{k\Psi})$ and control the part of \cref{eq:complicated_term} corresponding to $\FD^2 f(\newy)-\FD^2 f(\newyt^{k\Psi})$ separately from the $\FD^2 f(\newyt^{k\Psi})$ term. 
This type of decomposition is common in the Stein's method literature, as the latter part is typically easy to control using the aforementioned independence of $\newyt^{k\Psi}$ and $\left(\Gradbatch{k}{i}\right)_{i=1}^b$. The former part may be controlled using the fact that $\FD^2f$ is Lipschitz (see \cref{f_properties}). mainlpbound
Yet, obtaining a bound that vanishes in the limit can be challenging for certain target processes. 
In particular, to our best knowledge, before our work such a bound has not been obtained in the literature in any case in which the target process has non-independent increments, as explained in \cref{rmk:comparison_stein}. The reason is that previous works on functional Stein's method, including \citep{kasprzak2020brownian,kasprzak2020exchangeable,dobler2021stein} did not consider quantities of the form $\newyt^{k\Psi}$, as in $\cref{58}$ in the supplementary material, but instead worked with ones that in our context would take the form
\[\widetilde{\newyt^{k\Psi}}:={} & \newyt-w\sum_{m=k+1}^{\timescale}Q(k+1,m)\rbra*{\frac{\stepsize}{b}\sum_{i=1}^{b}\Gradbatch{k}{i}}\mathbb{I}_{[\frac{m}{\timescale
},\frac{m+1}{\timescale})}(t)\\
&-w\sum_{m=k+1}^{\timescale}\sum_{j=0}^{k-1}Q(j+1,m)\rbra*{\frac{h}{b}\sum_{i=1}^{b}\Gradbatch{j}{i}+\sqrt{\frac{2h}{\invtemp}}\noise_{j}}\ind{\cointer{\frac{m}{\samplesize},\frac{m+1}{\samplesize}}}(t).
\]
Such $\widetilde{\newyt^{k\Psi}}$ is also independent of $\left(\Gradbatch{k}{i}\right)_{i=1}^b$ but the distance between $\widetilde{\newyt^{k\Psi}}$ and $\newyt$ is difficult to control with a bound that vanishes in the limit. As a result, had we worked with $\widetilde{\newyt^{k\Psi}}$ instead of $\newyt^{k\Psi}$, we would not have been able to obtain a small enough bound on the part of \cref{eq:complicated_term} corresponding to $\FD^2 f(\newy)-\FD^2 f(\newyt^{k\Psi})$. Indeed, applying the Lipschitz property of $\FD^2f$, we would have struggled to obtain a small enough bound on $\|\newyt-\widetilde{\newyt^{k\Psi}}\|$. 
Hence, we deviated from this approach and used $\newyt^{k\Psi}$, which involves the (vanishing as $h\to 0$) multiplicative factor of $\frac{-\stepsize\sum_{i=1}^{b}\Heisbatch{k}{i}}{\batchsize - \stepsize\sum_{i=1}^{b}\Heisbatch{k}{i}}$ in the second line of \cref{eq:newyt_subs}. Our modified process $\newyt^{k\Psi}$ stays independent from $\left(\Gradbatch{k}{i}\right)_{i=1}^b$, despite involving the multiplicative factor. Yet, it is precisely because this multiplicative factor vanishes as $h\to 0$ that we were able to obtain a small (in expectation) bound on $\|\newyt-\newyt^{k\Psi}\|$ and thus a small bound on the part of \cref{eq:complicated_term} corresponding to $\FD^2 f(\newy)-\FD^2 f(\newyt^{k\Psi})$.
\end{remark}

\subsection{Final Bound}\label{sec:simplified_bound}
Combining the individual bounds, we obtain the following general (but still simplified, as compared to our most general \cref{MainThm} in the supplementary material) error bound. 
\begin{theorem}\label{thm:general_simplified}
Retain the notation of \cref{sec:approach,sec:main}. Suppose that processes $Y$ and $Z$ are univariate (real-valued). Let $\testfun\in M$ and $\bar\testfun = \testfun/\|\testfun\|_M$. Under \crefrange{asmp:curvature}{asmp:bounded-iterates}, if $Z_0 = \pmt_0 = 0$, $h\leq 1$, $\beta \geq 1$ and $\alpha h \le \overline{C} < \infty$, the following holds:
\[
    \lefteqn{|\EE\bar\testfun(Y)-\EE\bar\testfun(Z)|} \\
    &\lesssim C_{\max}\Bigg\{L^2 \rbra*{\frac{h}{b} + \frac{h^{1/2}}{b^{1/2}}\frac{1}{\beta^{1/2}}}\\
    &\phantom{\lesssim C_{\max}\Bigg\{} + w (1+K_1 C_R)\rbra*{\frac{h}{b} + \frac{1}{\beta} + \sbra*{\frac{h}{b^{1/2}} + \frac{h^{1/2}}{\beta^{1/2}}}\log^{1/2}(2\alpha)}\\
    &\phantom{\lesssim C_{\max}\Bigg\{} + w^2 L^{13/2}\rbra*{\frac{h^{3/2}}{b} + \frac{h}{\beta} + \frac{h^{7/4}}{\beta^{3/4}} + \frac{h^{1/2}}{b^{1/2}}\frac{h^{1/2}}{\beta^{1/2}} + \frac{h^{1/2}}{b^{1/2}}\frac{1}{\beta}}\\
    &\phantom{\lesssim C_{\max}\Bigg\{} + w^3 L^{27/4}(1+ C_R^3 K_3^3)\bigg(\frac{1}{\beta^3} + \frac{h^{25/8}}{\beta^{9/8}} + \frac{h^{7/4}}{\beta^{5/4}} + \frac{h^{1/2}}{b^{1/2}}\frac{h^{1/2}}{\beta} +\frac{h^{1/2}}{b^{1/2}}\frac{h^{7/4}}{\beta^{3/4}} \\
    &\hspace{5.5cm} + \frac{h}{b}\frac{h^{1/2}}{\beta^{1/2}} + \sbra*{\frac{h^{3}}{b^{3/2}} + \frac{h^{3/2}}{\beta^{3/2}}}\log^{3/2}(2\alpha) \bigg)\\
    &\phantom{\lesssim C_{\max}\Bigg\{} + w^4 L^{27/4}(1+ C_R K_3) \bigg(\frac{1}{\beta^{5/2}} + \frac{h^{15/8}}{\beta^{17/8}} + \frac{h}{b}\frac{h^{15/8}}{\beta^{9/8}} + \frac{h^{3/2}}{b^{3/2}}\frac{h^{1/2}}{\beta^{1/2}} \\
    &\hspace{5.5cm} + \frac{h}{b}\frac{h^{1/2}}{\beta}+ \frac{h}{b}\frac{h^{7/4}}{\beta^{3/4}} + \frac{h^{1/2}}{b^{1/2}}\frac{1}{\beta^{3/2}} \\
    &\hspace{5.5cm} +\rbra*{\frac{h^{5/2}}{b^{2}} + \frac{h^{1/2}}{\beta^{2}}}\log^{1/2}(2\alpha) \bigg)\\
    &\phantom{\lesssim C_{\max}\Bigg\{} + w^5 L \rbra*{\frac{h^{3}}{b^{5/2}} + \frac{h^{1/2}}{\beta^{5/2}} + \frac{h}{b}\frac{h^{1/2}}{\beta^{3/2}}+ \frac{h^{2}}{b^{2}}\frac{h^{1/2}}{\beta^{1/2}} + \frac{h^{3/2}}{b^{3/2}}\frac{h^{1/2}}{\beta} + \frac{h^{1/2}}{b^{1/2}}\frac{h^{1/2}}{\beta^{2}}} \Bigg\},
\]
where 
\[C_{\max} = \max \Bigg\{1, \Omega^3, \Sigma^3, \EE\Gradbatch{1}{1}^6, \overline{C}^{19/4} \Bigg\}.
\]
\begin{proof}
    The bound follows directly from \cref{BDforR,BDforZ,BDexch,BDrem,lemma_error_cov} by summing together the bounds appearing therein and applying the assumptions $\invtemp>1$, $L\geq 1$ and the assumption that $\timescale \stepsize$ is upper bounded by a positive constant $\overline{C}$.
\end{proof}
\begin{remark}
   In \cref{thm:general_simplified}, we choose to assume that $\timescale \stepsize$ is upper bounded by a positive constant and further assume that $\invtemp > 1$ as such assumptions significantly simplify the presentation of the bound and are common in the majority of applications of our theory.
   The former assumption is reasonable since the limiting process is non-trivial when $\timescale \stepsize$ converges to a constant. 
   Regarding the latter assumption, a typical choice is $\invtemp \propto \samplesize^\kappa$ for $\kappa \in (0,1]$, since otherwise 
   the stationary distribution does not converge (if $\kappa = 0$) or has no asymptotic effect (if $\kappa > 1)$. 
   We postpone the statement of our fully general bound (\cref{MainThm} in the supplementary) to the supplementary material due to its length. %
\end{remark}
\end{theorem}

\section*{Acknowledgments}
XW and JHH were partially supported by National Science Foundation CAREER award IIS-2340586.
MK was partially supported by the France 2030 program and by the European Union's
Horizon 2020 research and innovation program under grant
agreement 101024264.
JN was supported by an NSERC Discovery grant. 
SB was supported in part by Simons Foundation grant 635136.

\appendix 

\section{Statement of our most general functional error bound} \label{app:additional proofs}
In this appendix, notation ``$\lesssim$'' will mean ``less or equal up to a numerical constant'' -- i.e., ``$\leq C \times$'', where $C$ is a numerical constant independent of $\timescale, \stepsize, \invtemp, L, \Omega, \Sigma, C_R, \spacescale$, and the moments of $\Gradbatch{1}{1}$.

\begin{theorem}\label{MainThm}
Let $\testfun\in M$ and $\bar\testfun:=\testfun/\|\testfun\|_M$. Under \crefrange{asmp:curvature}{asmp:bounded-iterates}, if $Z_0 = \pmt_0 = 0$ and $h\leq 1$, $L\geq 1$, the following holds:
\[|\EE\testfun(Y)-\EE\testfun(Z)|\lesssim  \overline{\epsilon_R} + \overline{\epsilon_Z} + \overline{\error{exch}} + \overline{\error{cov}} + \overline{\error{rem}},\]

where

\[
\overline{\epsilon_R} &= w\timescale K_1C_R\rbra*{\frac{h^2\Omega}{b}+\frac{h}{\invtemp}} \\
&\quad+ w^3 \timescale^3 C_R^3 K_3^3 \rbra*{\frac{h^6}{b^3}\rbra*{\Omega^3+(\EE \Gradbatch{1}{1}^4)^{3/2}+\EE \Gradbatch{1}{1}^4\Omega+\EE\Gradbatch{1}{1}^6} +\frac{h^3}{\invtemp^3}}\\
&\quad +w^3\Bigg\{\frac{\timescale^2 \stepsize^4}{\batchsize^2}\sbra*{\frac{\timescale\stepsize^2L^2}{b}E_1+\timescale^2\stepsize^4L^4E_1+\frac{\EE \Gradbatch{1}{1}^4}{b}+\Omega^2+\Omega}\\
&\quad+\frac{\timescale^2\stepsize^2}{\invtemp^2}\sbra*{1+\frac{\timescale\stepsize^2L^2}{\batchsize}+\timescale^2\stepsize^4L^4}+\rbra*{\frac{1}{\batchsize}+\timescale \stepsize^2L^2}\sqrt{\Omega}(1+L^3) \frac{\timescale^3h^{11/2}L^2}{\invtemp^{3/2}}\Bigg\}^{3/4} \\
&\qquad  \times w\timescale\stepsize C_R K_3 \sbra*{\frac{h}{b}\rbra*{\Omega^3+(\EE \Gradbatch{1}{1}^4)^{3/2}+\EE \Gradbatch{1}{1}^4\Omega+\EE\Gradbatch{1}{1}^6}^{1/3} +\frac{1}{\invtemp}},\\
\overline{\epsilon_Z} &= \frac{\spacescale\timescale^{1/2}\stepsize^2\Omega^{1/2}\Sigma}{\batchsize^{1/2}}+\frac{\spacescale\timescale^{1/2}\stepsize^{3/2}\Sigma}{\invtemp^{1/2}}+\frac{\spacescale\stepsize\Omega^{1/2}\sqrt{\log(2\timescale)}}{\batchsize^{1/2}}+\frac{\spacescale\stepsize^{1/2}\sqrt{\log(2\timescale)}}{\invtemp^{1/2}}\\
&+\frac{\spacescale\timescale^{3/2}\stepsize^6\Omega^{3/2}\Sigma^3}{\batchsize^{3/2}}+\frac{\spacescale^3\timescale^{3/2}\stepsize^{9/2}\Sigma^3}{\invtemp^{3/2}}+\frac{\spacescale^3\stepsize^3\Omega^{3/2}\log^{3/2}(2\timescale)}{\batchsize^{3/2}}+\frac{\spacescale^3\stepsize^{3/2}\log^{3/2}(2\timescale)}{\invtemp^{3/2}}\\
&+\frac{\spacescale^4\timescale^{2}\stepsize^{5}\Omega^2\Sigma}{\batchsize^2}+\frac{\spacescale^4\timescale^{3/2}\stepsize^{4}\Omega^2\sqrt{\log(2\timescale)}}{\batchsize^2}+\frac{\spacescale^4\timescale^{2}\stepsize^{3}\Sigma}{\invtemp^2}+\frac{\spacescale^4\timescale^{3/2}\stepsize^{2}\sqrt{\log(2\timescale)}}{\invtemp^2};\\
\overline{\error{rem}} &= \frac{D_1\timescale \spacescale\stepsize L }{b^{1       /2}}\sqrt{\rbra*{1+\timescale^2 \stepsize^2 L^2}\rbra*{\frac{h^2 \Omega}{b} + \frac{h}{\invtemp}}};\\
\overline{\error{exch}} &= \spacescale^3  \Bigg\{L^3 \frac{\timescale^{5/2}h^6}{b^{3/2}} E_1^{3/4}+\frac{L^3\timescale^{5/2}h^{9/2}}{\invtemp^{3/2}}+\sqrt{\Omega}L^3(1+L^{9/4}) \frac{\timescale^{5/2}h^{45/8}}{\invtemp^{9/8}}\\
&\qquad+\frac{\timescale\stepsize^3}{b^{3/2}} (\EE \Gradbatch{1}{1}^4+\Omega^2)^{3/4}+\frac{\timescale h^{3/2}}{\invtemp^{3/2}}\Bigg\};\\
\overline{\error{cov}} &= D_2\timescale\spacescale^2\stepsize^2\Bigg\{\frac{\timescale\stepsize^2L^2}{b} \Bigg[\EE \Gradbatch{1}{1}^4+\Omega^2+\sqrt{\Omega}L^3\sbra*{\rbra*{\EE \Gradbatch{1}{1}^4+\Omega^2}^{3/4}+1}\Bigg]^{1/2}\\
&+\frac{L^{9/2}\timescale^{3/2}\stepsize^3\Omega }{b^{3/2}}+\frac{L^{9/2}\timescale^{3/2}\stepsize^{5/2}\Omega^{1/2} }{b\invtemp^{1/2}}+\frac{\timescale\stepsize L^2}{\invtemp}+\frac{L^{7/2}\Omega^{1/4}\timescale \stepsize^{7/4}}{\invtemp^{3/4}}+\frac{\timescale^{1/2}\stepsize\Omega L}{\batchsize}\\
&+\frac{\timescale^{1/2}\stepsize^{1/2}\sqrt{\Omega}L}{\batchsize^{1/2}\invtemp^{1/2}}+\frac{\spacescale\stepsize}{\batchsize^{3/2}}\rbra*{\rbra*{\EE \Gradbatch{1}{1}^4}^{3/4} + \Omega^{3/2}}+\frac{L\spacescale\stepsize^2\timescale  \Omega^{3/2}}{\batchsize^{3/2}}+\frac{L\spacescale\stepsize^{3/2}\timescale  \Omega}{\batchsize\invtemp^{1/2}} \\
&+\frac{L\timescale^{1/2}\stepsize}{\batchsize^{3/2}}+\frac{L^2\timescale\stepsize^{2}}{\batchsize^{3/2}}+\frac{\stepsize\Omega\Sigma}{\batchsize}+\frac{\timescale\stepsize^2\Omega\Sigma^2}{\batchsize}+\frac{L\timescale^{1/2}\stepsize^{1/2}\Omega^{1/2}}{\batchsize\invtemp^{1/2}}+\frac{L\timescale^{1/2}}{\batchsize^{1/2}\invtemp}+\frac{\spacescale\stepsize^{1/2}\Omega}{\batchsize\invtemp^{1/2}}\\
&+\frac{\spacescale\Omega^{1/2}}{\batchsize^{1/2}\invtemp}+\frac{L\spacescale\timescale\stepsize^{3/2}\Omega}{\batchsize\invtemp^{1/2}}+\frac{L\spacescale\timescale\stepsize\Omega^{1/2}}{\batchsize^{1/2}\invtemp}+\frac{\spacescale}{\stepsize^{1/2}\invtemp^{3/2}}+\frac{L\timescale^{1/2}}{\batchsize^{1/2}\invtemp}+\frac{L^2\timescale\stepsize}{\batchsize^{1/2}\invtemp}+\frac{\timescale\stepsize\Sigma^2}{\invtemp}+\frac{\Sigma}{\invtemp}\Bigg\}
\]

and
\[
D_1^2 &= 1 +w^4\frac{\timescale^2 \stepsize^4}{\batchsize^2}\sbra*{\frac{\timescale\stepsize^2L^2}{b}E_1+\timescale^2\stepsize^4L^4E_1+\frac{\EE \Gradbatch{1}{1}^4}{b}+\Omega^2+\Omega}\\
&+w^4\frac{\timescale^2\stepsize^2}{\invtemp^2}\sbra*{1+\frac{\timescale\stepsize^2L^2}{\batchsize}+\timescale^2\stepsize^4L^4}+w^4\rbra*{\frac{1}{\batchsize}+\timescale \stepsize^2L^2}\sqrt{\Omega}(1+L^3) \frac{\timescale^3h^{11/2}L^2}{\invtemp^{3/2}}\\
&+ \frac{1}{\FI^2 }\rbra*{\frac{w^2 h}{b} \Omega  + \frac{ w^2 }{\invtemp}}^2\log^2(1+\timescale\stepsize \FI );\\
D_2^2 &= 1+w^2\frac{\timescale \stepsize^2}{\batchsize}\sbra*{\sqrt{\Omega}L(1+L^3)\frac{\timescale^{1/2}\stepsize^{3/4}}{\invtemp^{3/4}}+L^2\Omega\timescale\stepsize^2+\Omega}+w^2\frac{\timescale\stepsize}{\invtemp}\sbra*{\frac{\timescale^{1/2}\stepsize L}{\batchsize^{1/2}}+1+L^2\timescale\stepsize^2}\\
&\quad+w^2\frac{\timescale^{3/2}\stepsize^3L}{\batchsize^{3/2}}\sqrt{E_1}+w^2 \rbra*{\frac{\Omega\stepsize^2} {b}+\frac{\stepsize}{\invtemp}}\rbra*{1+\frac{\timescale^2 h^2 L^2}{(1-L\stepsize)^2}}\\
&\quad  +\frac{1}{\FI ^2}\rbra*{\frac{w^2 h}{b} \Omega  + \frac{ w^2 }{\invtemp}}^2\log ^2(1+\timescale\stepsize \FI );\\
E_1 &= \EE \Gradbatch{1}{1}^4+\Omega^2+\sqrt{\Omega}(1+L^3)\sbra*{\rbra*{\EE \Gradbatch{1}{1}^4+\Omega^2}^{3/4}+1}+\Omega L(1+L^4)\sbra*{\Omega\frac{\timescale h^2}{b}+\frac{\timescale h}{\invtemp}}.
\]

\end{theorem}
\begin{proof}
     The bound follows directly from \cref{BDforR,BDforZ,BDexch,BDrem,lemma_error_cov} by summing together the bounds appearing therein.
\end{proof}

\section{Proof of Theorem~\ref{THM:MAIN-THEOREM-SIMPLE}}
\label{sec:proof-of-main-theorem-simple}

\subsection{Proof of {\cref{thm:main-theorem-simple}} (Numerical setting)}

By plugging in the parameters into the error bound from \cref{MainThm}, we obtain
\[
\overline{\epsilon_R} &\lesssim K_1C_R\rbra*{\rbra*{\frac{\stepsize}{\batchsize}}^{1/2}\Omega+\frac{1}{\invtemp^{1/2}}} \\
&\quad+ C_R^3 K_3^3 \rbra*{\frac{h^{3/2}}{b^{3/2}}\rbra*{\Omega^3+(\EE \Gradbatch{1}{1}^4)^{3/2}+\EE \Gradbatch{1}{1}^4\Omega+\EE\Gradbatch{1}{1}^6} +\frac{1}{\invtemp^{3/2}}}\\
&\quad +\Bigg\{\frac{\stepsize^{1/2}}{\batchsize^{1/2}}\sbra*{L^4E_1+\frac{\EE \Gradbatch{1}{1}^4}{b}+\Omega^2+\Omega}^{3/4}+\frac{L^{3}}{\invtemp^{1/2}}+\rbra*{\frac{1}{\batchsize^{3/4}}+ \stepsize^{3/4} L^{3/2}}\Omega^{3/8}L^{15/4} \frac{h^{15/8}}{\invtemp^{1/8}}\Bigg\} \\
&\qquad  \times C_R K_3 \sbra*{\rbra*{\Omega^3+(\EE \Gradbatch{1}{1}^4)^{3/2}+\EE \Gradbatch{1}{1}^4\Omega+\EE\Gradbatch{1}{1}^6}^{1/3} +1},\\
\overline{\epsilon_Z} &\lesssim \stepsize\Omega^{1/2}\Sigma+\stepsize\Sigma+\stepsize^{1/2}\Omega^{1/2}\sqrt{\log(\stepsize^{-1})}+\stepsize^{1/2}\sqrt{\log(\stepsize^{-1})}+\frac{\stepsize^4\Omega^{3/2}\Sigma^3}{\batchsize}+\stepsize^{3}\Sigma^3\\
&\quad+\stepsize^{3/2}\Omega^{3/2}\log^{3/2}(\stepsize^{-1})+\stepsize^{3/2}\log^{3/2}(\stepsize^{-1})+\stepsize\Omega^2\Sigma+\stepsize^{1/2}\Omega^2\sqrt{\log(\stepsize^{-1})}+\stepsize\Sigma\\
&\quad +\stepsize^{1/2}\sqrt{\log(\stepsize^{-1})};\\
&\overline{\error{rem}} \lesssim D_1  L \sqrt{\rbra*{1+\timescale^2 \stepsize^2 L^2}\rbra*{\frac{h\Omega}{b} + \frac{1}{\invtemp}}};\\
&\overline{\error{exch}} \lesssim L^3 \stepsize^{2}E_1^{3/4}+L^3h^{2}+\sqrt{\Omega}L^3(1+L^{9/4}) h^{11/4}\batchsize^{3/8}+\stepsize^{1/2} (\EE \Gradbatch{1}{1}^4+\Omega^2)^{3/4}+h^{1/2};\\
&\overline{\error{cov}} \lesssim D_2\Bigg\{\stepsize L^2 \Bigg[\EE \Gradbatch{1}{1}^4+\Omega^2+\sqrt{\Omega}L^3\sbra*{\rbra*{\EE \Gradbatch{1}{1}^4+\Omega^2}^{3/4}+1}\Bigg]^{1/2}+\frac{L^{9/2}\stepsize^{3/2}\Omega }{b^{1/2}}\\
&\quad+\frac{L^{9/2}h\Omega^{1/2} }{\invtemp^{1/2}}+hL^2+L^{7/2}\Omega^{1/4} \stepsize^{3/2}\batchsize^{1/4}+\stepsize^{1/2}\Omega L+\stepsize^{1/2}\sqrt{\Omega}L\\
&\quad+\stepsize^{1/2}\rbra*{\rbra*{\EE \Gradbatch{1}{1}^4}^{3/4} + \Omega^{3/2}}+L\stepsize^{1/2}  \Omega^{3/2}+L\stepsize^{1/2}  \Omega+\frac{L\stepsize^{1/2}}{\batchsize^{1/2}}+\frac{L^2\stepsize}{\batchsize^{1/2}} +\stepsize\Omega\Sigma\\
&\quad+\stepsize\Omega\Sigma^2+\frac{L\Omega^{1/2}}{\invtemp^{1/2}}+\frac{L\stepsize^{1/2}}{\batchsize^{1/2}}+\stepsize^{1/2}\Omega\\
&\quad+\Omega^{1/2}\stepsize^{1/2}+L\stepsize^{1/2}\Omega+\frac{L\spacescale^{1/2}\Omega^{1/2}}{\batchsize^{1/2}}+\stepsize^{1/2}+\frac{L\stepsize^{1/2}}{\batchsize^{1/2}}+\frac{L^2\stepsize}{\batchsize^{1/2}}+\Sigma^2\stepsize+\Sigma\stepsize\Bigg\}
,\]
and
\[
D_1^2 &\lesssim 1 +\frac{\stepsize L^2}{b}E_1+\stepsize^2L^4E_1+\frac{\EE \Gradbatch{1}{1}^4}{b}+\Omega^2+\Omega\\
&+1+\frac{\stepsize L^2}{\batchsize}+\stepsize^2L^4+\rbra*{\frac{1}{\batchsize}+ \stepsize L^2}\sqrt{\Omega}(1+L^3) h^{2}b^{1/2}L^2\\
&+ \frac{1}{\FI^2 }\rbra*{\Omega  + 1}^2\log^2(1+ \FI )\in O(L^9);\\
D_2^2 &\lesssim 1+\sbra*{\sqrt{\Omega}L(1+L^3)\frac{\stepsize^{1/4}}{\invtemp^{3/4}}+L^2\Omega\stepsize+\Omega}+\sbra*{\frac{\stepsize^{1/2} L}{\batchsize^{1/2}}+1+L^2\stepsize}\\
&\quad+\frac{\stepsize^{1/2}L}{\batchsize^{1/2}}\sqrt{E_1}+\Omega\stepsize+\stepsize\rbra*{1+\frac{L^2}{(1-L\stepsize)^2}}  +\frac{1}{\FI ^2}\log ^2(1+ \FI )\in O(L^4);\\
&E_1\in O(L^5).
\]
The bound in \cref{thm:main-theorem-simple}, in the numerical setting, now follows by \cref{MainThm} and a straightforward simplification of the bounds presented above.
\qed

\subsection{Proof of \cref{thm:main-theorem-simple} (Statistical setting)}

We apply the assumptions of the statistical setting part of \cref{thm:main-theorem-simple}, together with the fact that $\batchsize\leq\samplesize$. By plugging in the parameters into the error bound from \cref{MainThm}, we obtain a bound expressed in terms of $m,n,b$:

\[
\overline{\epsilon_R} &\lesssim \frac{m}{ \samplesize^{1/2}} K_1C_R\rbra*{\Omega +1} \\
&\quad+ \frac{m^3}{\samplesize^{3/2}} C_R^3 K_3^3 \sbra*{\rbra*{\Omega^3+(\EE \Gradbatch{1}{1}^4)^{3/2}+\EE \Gradbatch{1}{1}^4\Omega+\EE\Gradbatch{1}{1}^6} +1}\\
&\quad +\frac{m^{5/2}\batchsize^{1/2} C_R K_3 L^{21/4}}{\samplesize^{1/2}}\sbra*{E_1+\EE \Gradbatch{1}{1}^4+\Omega^2+\Omega+1+\sqrt{\Omega}}^{3/4} \\
&\qquad  \times  \sbra*{\rbra*{\Omega^3+(\EE \Gradbatch{1}{1}^4)^{3/2}+\EE \Gradbatch{1}{1}^4\Omega+\EE\Gradbatch{1}{1}^6}^{1/3} +1},\\
\overline{\epsilon_Z} &\lesssim \frac{m^{1/2}\batchsize\Omega^{1/2}\Sigma}{\samplesize}+\frac{m^{1/2}\batchsize\Sigma}{\samplesize}+\frac{\batchsize^{1/2}\Omega^{1/2}\sqrt{\log(m\samplesize/\batchsize)}}{\samplesize^{1/2}}+\frac{\batchsize^{1/2}\sqrt{\log(m\samplesize/\batchsize)}}{\samplesize}\\
&+\frac{m^{3/2}\batchsize^3\Omega^{3/2}\Sigma^3}{\samplesize^4}+\frac{\batchsize^3m^{3/2}\Sigma^3}{\samplesize^3}+\frac{\batchsize^{3/2}\Omega^{3/2}\log^{3/2}(m\samplesize/\batchsize)}{\samplesize^{3/2}}+\frac{\batchsize^{3/2}\log^{3/2}(m\samplesize/\batchsize)}{\samplesize^{3/2}}\\
&+\frac{\batchsize m^2\Omega^2\Sigma}{\samplesize}+\frac{m^{3/2}\batchsize^{1/2}\Omega^2\sqrt{\log(m\samplesize/\batchsize)}}{\samplesize^{1/2}}+\frac{m^2\batchsize\Sigma}{\samplesize}+\frac{m^{3/2}\batchsize^{1/2}\sqrt{\log(m\samplesize/\batchsize)}}{\samplesize^{1/2}};\\
\overline{\error{rem}} &\lesssim \frac{D_1m^2  L^2 }{\samplesize^{1/2}}\sqrt{ \Omega +1};\\
\overline{\error{exch}} &\lesssim \frac{\batchsize^{1/2}m^{5/2}L^{21/4}}{\samplesize^{1/2}}  \sbra*{E_1^{3/4}+\sqrt{\Omega} +(\EE \Gradbatch{1}{1}^4+\Omega^2)^{3/4}+1};\\
\overline{\error{cov}} &\lesssim D_2m\batchsize\Bigg\{\frac{mL^2}{\samplesize} \Bigg[\EE \Gradbatch{1}{1}^4+\Omega^2+\sqrt{\Omega}L^3\sbra*{\rbra*{\EE \Gradbatch{1}{1}^4+\Omega^2}^{3/4}+1}\Bigg]^{1/2}\\
&+\frac{L^{9/2}m^{3/2}\Omega }{\samplesize^{3/2}}+\frac{L^{9/2}m^{3/2}\Omega^{1/2} }{\samplesize^{3/2}}+\frac{m L^2}{\samplesize}+\frac{L^{7/2}\Omega^{1/4}m\batchsize^{3/4}}{\samplesize^{3/2}}+\frac{m^{1/2}\Omega L}{\batchsize^{1/2}\samplesize^{1/2}}\\
&+\frac{\sqrt{\Omega}L}{\batchsize^{1/2}\samplesize^{1/2}}+\frac{1}{\batchsize^{1/2}\samplesize^{1/2}}\rbra*{\rbra*{\EE \Gradbatch{1}{1}^4}^{3/4} + \Omega^{3/2}}+\frac{Lm \Omega^{3/2}}{\samplesize^{1/2}}+\frac{Lm  \Omega}{\batchsize^{1/2}\samplesize^{1/2}} \\
&+\frac{Lm^{1/2}}{\batchsize\samplesize^{1/2}}+\frac{L^2m}{\batchsize^{1/2}\samplesize}+\frac{\Omega\Sigma}{\samplesize}+\frac{m\Omega\Sigma^2}{\samplesize}+\frac{Lm^{1/2}\Omega^{1/2}}{\batchsize\samplesize^{1/2}}+\frac{Lm^{1/2}}{\batchsize\samplesize^{1/2}}+\frac{\Omega}{\batchsize^{1/2}\samplesize^{1/2}}\\
&+\frac{\Omega^{1/2}}{\batchsize^{1/2}\samplesize^{1/2}}+\frac{Lm\Omega}{\batchsize^{1/2}\samplesize^{1/2}}+\frac{Lm\Omega^{1/2}}{\batchsize^{1/2}\samplesize^{1/2}}+\frac{1}{\batchsize^{1/2}\samplesize^{1/2}}+\frac{Lm^{1/2}}{\batchsize\samplesize^{1/2}}+\frac{L^2m}{\batchsize^{1/2}\samplesize}+\frac{m\Sigma^2}{\samplesize}+\frac{\Sigma }{\samplesize}\Bigg\}
\]

and

\[
D_1^2 &\lesssim 1 +m^2\sbra*{\frac{m}{\samplesize}E_1+\frac{m^2\batchsize^2}{\samplesize^2}L^4E_1+\frac{\EE \Gradbatch{1}{1}^4}{b}+\Omega^2+\Omega}+m^2\sbra*{1+\frac{m}{\samplesize}+\frac{m^2\batchsize^2}{\samplesize^2}L^4}\\
&+\rbra*{\frac{1}{\batchsize}+\frac{\batchsize m}{\samplesize}L^2}\sqrt{\Omega}(1+L^3) \frac{m^3\batchsize^{5/2}L^2}{\samplesize^2}+ \frac{1}{\FI^2 }\rbra*{ \Omega  + 1}^2\log^2(1+m \FI );\\
D_2^2 &\lesssim 1+m^2\sbra*{\sqrt{\Omega}L^4 +L^2\Omega  +\Omega}+m^2L^2\\
&\quad+m^{3/2}\sqrt{E_1}+ \rbra*{\Omega  +1}\rbra*{1+m^2 L^2} +\frac{1}{\FI ^2}\rbra*{ \Omega  + 1}^2\log ^2(1+m \FI );\\
E_1 &\lesssim \EE \Gradbatch{1}{1}^4+\Omega^2+\sqrt{\Omega}L^3\sbra*{\rbra*{\EE \Gradbatch{1}{1}^4+\Omega^2}^{3/4}+1}+\Omega L^5\rbra*{\Omega +1}.
\]

The bound now follows from a further straightforward simplification of the bounds above.\qed

\section{Proof of Corollary \ref{cor_levy_wasserstein}}
\label{sec:proof_levy_wass}
\subsection{Auxiliary lemma}
First, we present a lemma, which will be used in the main part of the proof. \Cref{BDforSMfull} is proved in \cref{sec:lemma_levy_proof}.
\begin{lemma}\label{BDforSMfull}
Fix $r\geq 1$ and  suppose that \cref{asmp:curvature,asmp:stepsize} hold and that \cref{asmp:bounded-iterates} holds for $p=2r$. 
Suppose moreover that $\EE\left[\left|\Gradbatch{1}{1}\right|^{2r}\right]<\infty$ and $h<\min(1,\overline{\stepsize})$ , $b\geq 1$. Then
   \[
\EE\max_{k\in\{1,\dots,\timescale\}}|\pmt_k|^{2r} \lesssim &\timescale^{r/2}\Bigg\{\rbra*{1+\frac{L^rh^{r/2}}{b^{r/2}}}\frac{h^{3r/2}}{b^{r}} + \rbra*{1+\frac{L^r}{\invtemp^{r/2}}}\frac{h^{r}}{b^{r/2}\invtemp^{r/2}} + \frac{h^{r/2}}{\invtemp^{r}}   \\
&+ \rbra*{1+\frac{C_R^{2r}h^{2r}+h^rL^r}{b^r}}\frac{\timescale^{r/2}h^{2r}}{b^r}+ \rbra*{1+\frac{C_R^{2r}h^r+h^rL^r}{\invtemp^r}}\frac{\alpha^{r/2}h^{r}}{\invtemp^r}\\
&+\frac{\alpha^{r/2}h^{2r}}{\invtemp^{2r}}+\frac{\alpha^{r/2}h^{r}}{\invtemp^{3r/2}}\Bigg\}.
\]
\end{lemma}

\subsection{Main body of the proof of Corollary \ref{cor_levy_wasserstein}}
For any $w\in\skspace$ with $\|w\|<\infty$ and $\epsilon>0$, let the $\epsilon$-regularized version of $w$ be defined as follows:
\[
w_{\epsilon}(s)\defas \EE[w(s+\epsilon U)],
\]
where $U\sim\text{Uniform}(-1,1)$ and where we set $w(t)=w(1)$ for $t>1$ and $w(t)=0$ for $t<0$.

Furthermore, let $c_0(v)=1+v+\sqrt{2}v^2+\sqrt{\frac{50}{\pi}}v^3$ and, for any positive $\delta$ and $\epsilon$, let
\[
C_0(\epsilon,\delta)\defas c_0\left(\frac{\sqrt{1+\epsilon^3}}{\epsilon\delta}\right).
\]
As in \citet[eq. (1.12)]{barbour2024stein}, note that, for $0<\epsilon\leq\frac{1}{2}$ and $0<\delta\leq 2$, 
\[\label{c0bound}
C_0(\epsilon,\delta)\leq 10.5 \left(\frac{1}{\epsilon\delta}\right)^3.
\]
Finally, recalling the notation of \Cref{thm:main-theorem-simple}, let
\[\label{kappa1}
\kappa_1 = C_{\mathrm{num}}\cbra*{\stepsize^{1/2}\log^{1/2}(\stepsize^{-1}) + \invtemp^{-1/2}}
\]

\subsubsection*{Bounding the L\'evy--Prokhorov distance}
Now, by \citet[Corollary 1.3]{barbour2024stein}, for any positive $\delta,\epsilon,\theta$ and $\gamma$,
\[ \label{d_lp_bound1}
\begin{split}
  \lefteqn{d_\mathrm{LP}\left(\law{Y},\law{Z}\right)} \\
&\leq 2(\theta+\gamma) \vee \left\{  C_0(\epsilon,\delta)\kappa_1 +\Pr(\|Y_{\epsilon}-Y\|\geq \theta)+\Pr(\|Z_{\epsilon}-Z\|\geq \theta)+6e^{-\frac{\gamma^2}{8\delta^2}}\right\}.  
\end{split}
\]

First, note that, by \cref{c0bound,kappa1}, 
\[\label{c_0_bound}
C_0(\epsilon,\delta)\kappa_1\leq \frac{\tilde{C}_1(C_RL^7+C_R^3L^7+C_R)}{(\epsilon\delta)^3} \left[\invtemp^{-1/2}+\sqrt{\abs{\stepsize \log(\stepsize^{-1})}}\right]
\]
for a constant $\tilde{C}_1$, which does not depend on $\epsilon,\delta,\theta,\gamma,w,\stepsize,\invtemp, b, C_R, L$.

Second, let us concentrate on the term $\Pr(\|Z_{\epsilon}-Z\|\geq \theta)$. In order to bound this term, we note that
$Z_t\defas w\sqrt{\frac{\stepsize^2\timescale}{b}\Omega+\frac{2\stepsize\timescale}{\invtemp}}\int_0^te^{-\stepsize\FI \timescale(t-s)}\d W_s$, $t\in[0,1]$, and, therefore, for $t\in[0,1]$,
\[
Z_t 
&\stackrel{{D}}=w\sqrt{\frac{\stepsize^2\timescale}{2\stepsize\FI \timescale b}\Omega+\frac{\stepsize\timescale}{\stepsize\FI \timescale\invtemp}}e^{-\stepsize\FI \timescale t}W\left(e^{2\stepsize\FI \timescale t}\right) \\
&= w\sqrt{\frac{\stepsize}{2\FI b}\Omega+\frac{1}{\FI \invtemp}} W\left(1-e^{-2\stepsize\FI \timescale t}\right).
\]

It follows that, for $s<t$,
\[
\lefteqn{\EE |Z_t-Z_s|^2} \\
&= w^2 \left(\frac{\stepsize}{2\FI b}\Omega+\frac{1}{\FI \invtemp}\right)\EE \left[\left|W\left(1-e^{-2\stepsize\FI \timescale t}\right)-W\left(1-e^{-2\stepsize\FI \timescale s}\right)\right|^2\right]\\
&=w^2 \left(\frac{\stepsize}{2\FI b}\Omega+\frac{1}{\FI \invtemp}\right)\left[1-e^{-2\stepsize\FI \timescale t} - \left(1 - e^{-2\stepsize\FI \timescale s}\right)\right]\\
&=w^2 \left(\frac{\stepsize}{2\FI b}\Omega+\frac{1}{\FI \invtemp}\right)\left[e^{-2\stepsize\FI \timescale s} - e^{-2\stepsize\FI \timescale t}\right]\\
&=w^2 \left(\frac{\stepsize}{2\FI b}\Omega+\frac{1}{\FI \invtemp}\right)e^{-2\stepsize\FI \timescale s}\left[1 - e^{-2\stepsize\FI \timescale (t-s)}\right]\\
&\leq w^2\timescale \left(\frac{\stepsize^2}{b}\Omega+\frac{2\stepsize}{\invtemp}\right)(t-s). 
\]
Using \cite[Remark 1.7]{barbour2024stein} and our assumptions on $w,\stepsize,b,\invtemp$, we conclude that, for any $r\geq 1$ and any $\epsilon\in\left(\timescale^{-1},1\right)$,
\[\label{z_eps}
\Pr(\|Z_{\epsilon}-Z\|\geq \theta)\leq \tilde{C}_2\epsilon^{r-1}\theta^{-2r},
\]
for a constant $\tilde{C}_2$ independent of $\epsilon,\theta,w,\stepsize,b,\invtemp, L, C_R$.

Thirdly, let us upper-bound the term $\Pr(\|Y_{\epsilon}-Y\|\geq \theta)$. Let us fix $r>1$. Note that
\[
&\EE\left|Y_t-Y_s\right|^{2r}\\
&=w^{2r}\EE\left|\sum_{l=\lfloor \timescale s\rfloor+1}^{\lfloor \timescale t\rfloor}\stepsize \,\gradstochloss{l}(\newpmt_l) + \sqrt{\frac{2\stepsize}{\invtemp}}\noise_{l}\right|^{2r}\\
&= w^{2r}\EE\left|\sum_{l=\lfloor \timescale s\rfloor+1}^{\lfloor \timescale t\rfloor}\frac{\stepsize}{b}\sum_{i=1}^{b} \left(\psi_{I(l, i)} - \sigma_{I(l, i)}\pmt_l + \grad\obsloss{I(l, i)}(\pmt_l) - \widehat\grad\obsloss{I(l, i)}(\pmt_l)\right) + \sqrt{\frac{2\stepsize}{\invtemp}}\noise_{l}\right|^{2r}\\
&\leq 4^{2r-1}w^{2r}\Bigg[\left(\frac{\stepsize}{b}\right)^{2r}\EE\left|\sum_{l=\lfloor \timescale s\rfloor+1}^{\lfloor \timescale t\rfloor}\sum_{i=1}^{b} \left(-\sigma_{{I(l,i)}}\theta_l\right) \right|^{2r}+\left(\frac{\stepsize}{b}\right)^{2r}\EE\left|\sum_{l=\lfloor \timescale s\rfloor+1}^{\lfloor \timescale t\rfloor}\sum_{i=1}^{b}\psi_{I(l,i)}\right|^{2r} \\
&\hspace{1cm} + \left(\frac{\stepsize}{b}\right)^{2r}\EE\left|\sum_{l=\lfloor \timescale s\rfloor+1}^{\lfloor \timescale t\rfloor}\sum_{i=1}^{b}\grad\obsloss{I(l, i)}(\pmt_l) - \widehat\grad\obsloss{I(l, i)}(\pmt_l)\right|^{2r}
+\left(\frac{2\stepsize}{\invtemp}\right)^r\EE\left|\sum_{l=\lfloor \timescale s\rfloor+1}^{\lfloor \timescale t\rfloor} \noise_{l}\right|^{2r}\Bigg].
\]
We shall look at the four summands separately, in turn. Using \Cref{BDforSMfull} and the assumption that $\timescale \stepsize\in O(1)$ it follows that:
\[
&\EE\max_{k\in\{1,\dots,\timescale\}}|\pmt_k|^{2r} \leq C_5\Bigg\{\left(1+\frac{L^r\stepsize^{r/2}}{b^{r/2}}+\frac{C_R^{2r}\stepsize^{2r}}{b^r}\right)\frac{\stepsize^{r}}{b^{r}} + \left(1+\frac{L^r}{\invtemp^{r/2}}\right)\frac{\stepsize^{r/2}}{b^{r/2}\invtemp^{r/2}}  \\
&\hspace{3cm}+ \left(1+\frac{C_R^{2r}\stepsize^{2r}+\stepsize^rL^r}{b^{r}}\right)\frac{1}{\invtemp^r}+\frac{\stepsize^{r}}{\invtemp^{2r}}+\frac{1}{\invtemp^{3r/2}}\Bigg\}\\
\intertext{and}
&\EE\max_{k\in\{1,\dots,\timescale\}}|\pmt_k|^{4r} \leq C_5\Bigg\{\left(1+\frac{L^{2r}\stepsize^{r}}{b^{r}}+\frac{C_R^{4r}\stepsize^{4r}}{b^{2r}}\right)\frac{\stepsize^{2r}}{b^{2r}} + \left(1+\frac{L^{2r}}{\invtemp^{r}}\right)\frac{\stepsize^{r}}{b^{r}\invtemp^{r}}  \\
&\hspace{3cm}+ \left(1+\frac{C_R^{4r}\stepsize^{4r}+\stepsize^{2r}L^{2r}}{b^{2r}}\right)\frac{1}{\invtemp^{2r}}+\frac{\stepsize^{2r}}{\invtemp^{4r}}+\frac{1}{\invtemp^{3r}}\Bigg\}.
\]
Therefore, there exists uniform constants $\tilde{C}_r,\tilde{C}_r'$, not depending on $\timescale,\stepsize,b,\invtemp$, such that
\[
&\EE\left|\sum_{l=\lfloor \timescale s\rfloor+1}^{\lfloor \timescale t\rfloor} \noise_{l}\right|^{2r}=(2r-1)!! \left(\lfloor \timescale t\rfloor - \lfloor \timescale s\rfloor\right)^r\le \tilde{C}_r'\timescale^r (t-s)^r;\\
&\EE\left|\sum_{l=\lfloor \timescale s\rfloor+1}^{\lfloor \timescale t\rfloor}\sum_{i=1}^{b}\grad\obsloss{I(l, i)}(\pmt_l) - \widehat\grad\obsloss{I(l, i)}(\pmt_l)\right|^{2r}\leq C_R^{2r}b^{2r} \left(\lfloor \timescale t\rfloor - \lfloor \timescale s\rfloor\right)^{2r} \EE\left[\sup_{1\leq l\leq\timescale}\left|\theta_l\right|^{4r}\right]\\
&\hspace{3cm}\leq \tilde{C}_rC_R^{2r}b^{2r}\timescale^{2r}\left(t-s\right)^{2r}\Bigg\{\left(1+\frac{L^{2r}\stepsize^{r}}{b^{r}}+\frac{C_R^{4r}\stepsize^{4r}}{b^{2r}}\right)\frac{\stepsize^{2r}}{b^{2r}}   \\
&\hspace{3cm}+ \left(1+\frac{L^{2r}}{\invtemp^{r}}\right)\frac{\stepsize^{r}}{b^{r}\invtemp^{r}}+ \left(1+\frac{C_R^{4r}\stepsize^{4r}+\stepsize^{2r}L^{2r}}{b^{2r}}\right)\frac{1}{\invtemp^{2r}}+\frac{\stepsize^{2r}}{\invtemp^{4r}}+\frac{1}{\invtemp^{3r}}\Bigg\};\\
&\hspace{3cm}\leq \tilde{C}_r'C_R^{2r}\left(t-s\right)^{2r}\Bigg\{\left(1+\frac{L^{2r}\stepsize^{r}}{b^{r}}+\frac{C_R^{4r}\stepsize^{4r}}{b^{2r}}\right) + \left(1+\frac{L^{2r}}{\invtemp^r}\right)\frac{\timescale^{r}b^r}{\invtemp^{r}} \\
&\hspace{3cm}+ \left(1+\frac{C_R^{4r}\stepsize^{4r}+\stepsize^{2r}L^{2r}}{b^{2r}}\right)\frac{\timescale^{2r}b^{2r}}{\invtemp^{2r}}  +\frac{b^{2r}}{\invtemp^{4r}}+\frac{\timescale^{2r}b^{2r}}{\invtemp^{3r}}\Bigg\};\\
&\EE\left|\sum_{l=\lfloor \timescale s\rfloor+1}^{\lfloor \timescale t\rfloor}\sum_{i=1}^{b} \left(-\sigma_{I(l,i)}\theta_l\right) \right|^{2r}\leq L^{2r}b^{2r} \left(\lfloor \timescale t\rfloor - \lfloor \timescale s\rfloor\right)^{2r} \EE\left[\sup_{1\leq l\leq\timescale}\left|\theta_l\right|^{2r}\right]\\
&\hspace{2cm}\leq \tilde{C}_rL^{2r}b^{2r}\timescale^{2r}\left(t-s\right)^{2r}\Bigg\{\left(1+\frac{L^r\stepsize^{r/2}}{b^{r/2}}+\frac{C_R^{2r}\stepsize^{2r}}{b^r}\right)\frac{\stepsize^{r}}{b^{r}}\\
&\hspace{2cm}+ \left(1+\frac{L^r}{\beta^{r/2}}\right)\frac{\stepsize^{r/2}}{b^{r/2}\invtemp^{r/2}} + \left(1+\frac{L^r\stepsize^{r/2}}{b^{r/2}}+\frac{C_R^{2r}\stepsize^{2r}}{b^r}\right)\frac{1}{\invtemp^{r}}   +\frac{\stepsize^{r}}{\invtemp^{2r}}+\frac{1}{\invtemp^{3r/2}}\Bigg\} \\
&\hspace{2cm}\leq \tilde{C}_r'L^{2r}\left(t-s\right)^{2r}\Bigg\{ \left(1+\frac{L^r\stepsize^{r/2}}{b^{r/2}}+\frac{C_R^{2r}\stepsize^{2r}}{b^r}\right)\timescale^{r}b^{r} + \left(1+\frac{L^r}{\beta^{r/2}}\right)\frac{\timescale^{3r/2}b^{3r/2}}{\invtemp^{r/2}}\\
&\hspace{2cm}+ \left(1+\frac{L^r\stepsize^{r/2}}{b^{r/2}}+\frac{C_R^{2r}\stepsize^{2r}}{b^r}\right)\frac{\timescale^{2r}b^{2r}}{\invtemp^{r}}   +\frac{\timescale^{r}b^{2r}}{\invtemp^{2r}}+\frac{\timescale^{2r}b^{2r}}{\invtemp^{3r/2}}\Bigg\} ;\\
&\EE\left|\sum_{l=\lfloor \timescale s\rfloor+1}^{\lfloor \timescale t\rfloor}\sum_{i=1}^{b}\psi_{I(l,i)}\right|^{2r}\leq \tilde{C}_r' \timescale^r\left(t-s\right)^{r}\EE\left|\psi_{I(1,1)}\right|^{2r}.
\]
Therefore, we obtain that, for some constants $K_r$ and $\hat{C}_r$, independent of $w,\stepsize,b,\invtemp$, we have that 
\[
&\EE\left|Y_t-Y_s\right|^{2r} \\
\le& K_rw^{2r}(t-s)^r\Bigg\{\left(\frac{\stepsize}{b}\right)^{2r}\Bigg[C_R^{2r}\left(1+\frac{L^{2r}\stepsize^{r}}{b^{r}}+\frac{C_R^{4r}\stepsize^{4r}}{b^{2r}}\right) + C_R^{2r}\left(1+\frac{L^{2r}}{\invtemp^r}\right)\frac{\timescale^{r}b^r}{\invtemp^{r}} \\
&+ C_R^{2r}\left(1+\frac{C_R^{4r}\stepsize^{4r}+\stepsize^{2r}L^{2r}}{b^{2r}}\right)\frac{\timescale^{2r}b^{2r}}{\invtemp^{2r}}  +\frac{b^{2r}}{\invtemp^{4r}}+\frac{\timescale^{2r}b^{2r}}{\invtemp^{3r}} \\
&+L^{2r}\left(1+\frac{L^r\stepsize^{r/2}}{b^{r/2}}+\frac{C_R^{2r}\stepsize^{2r}}{b^r}\right)\timescale^{r}b^{r}+ \left(L^{2r}+\frac{L^{3r}}{\invtemp^{r/2}}\right)\frac{\timescale^{3r/2}b^{3r/2}}{\invtemp^{r/2}}  \\
&+ L^{2r}\left(1+\frac{L^r\stepsize^{r/2}}{b^{r/2}}+\frac{C_R^{2r}\stepsize^{2r}}{b^r}\right)\frac{\timescale^{2r}b^{2r}}{\invtemp^{r}}   +\frac{\timescale^{r}b^{2r}}{\invtemp^{2r}}+\frac{\timescale^{2r}b^{2r}}{\invtemp^{3r/2}}+\timescale^r \Bigg]+\timescale^r\left(\frac{\stepsize}{\invtemp}\right)^r\Bigg\}\\
\leq& \hat{C}_rL^{2r}(t-s)^r.\label{y_exp}
\]
Moreover, it follows from \cref{y_exp} that for any $\eta>0$,
\[\label{y_prob}
\lefteqn{\max_{1\leq k\leq \timescale} \timescale\, \Pr \left[\sup_{(k-1)/\timescale\leq s\leq k/\timescale}\left|Y_s-Y_{(k-1)/\timescale}\right|>\eta\right]} \\
&\qquad = \max_{1\leq k\leq \timescale} \timescale\,\Pr \left[\left|Y_{k/\timescale}-Y_{(k-1)/\timescale}\right|>\eta\right]\\
&\qquad \leq \hat{C}_rL^{2r}\timescale^{1-r}\eta^{-2r}.
\]
It now follows from \citet[Lemma 1.4 and its proof on p. 24-25]{barbour2024stein} and \cref{y_exp,y_prob} that for any $\epsilon\in (\timescale^{-1},1)$, and some constant $\hat{C}_r'$, not depending on $\theta,\timescale,b,\invtemp,\stepsize$,
\[\label{y_eps}
	\mathbb{P}\left[\|Y_{\epsilon}-Y\|>\theta\right]\leq& 2\hat{C}_r\timescale^{1-r} \left(\frac{\theta(1-2^{-(r-1)/(4r)})}{26}\right)^{-2r}+\hat{C}_r'L^{2r}\frac{\epsilon^{r-1}}{\theta^{2r}}.
\]
Finally, it follows from \cref{d_lp_bound1,c_0_bound,z_eps,y_eps} that, for any $\delta,\theta,\gamma>0$, $\timescale^{-1}<\epsilon<1$, and a constant $\hat{C}_r''$ independent of $w,\stepsize,\invtemp,b,\delta,\theta,\gamma,\epsilon$,
\[
&d_\mathrm{LP}\left(\law{Y},\law{Z}\right)\\
\leq &\hat{C}_r''\Bigg\{\theta+\gamma+ \frac{C_RL^7+C_R^3}{(\epsilon\delta)^{3}}\rbra*{\stepsize^{1/2}\log^{1/2}(1/\stepsize) + \invtemp^{-1/2}} + \stepsize^{r-1} \theta^{-2r}+ L^{2r}\epsilon^{r-1}\theta^{-2r} \Bigg\}\\
&+6 e^{-\gamma^2/(8\delta^2)}.
\]
Following the logic of \citep[Examples 1.8 and 1.10]{barbour2024stein}, and assuming that $\stepsize$ is small enough, we choose:
\[
\gamma=2 \stepsize^{\frac{r-1}{20r-2}}\sqrt{10\log\left(\stepsize^{-1}\right)};\quad \epsilon = \stepsize^{\frac{2r+1}{20r-2}};\quad \delta = \stepsize^{\frac{r-1}{20r-2}};\quad \theta=\stepsize^{\frac{r-1}{20r-2}}
\]
and obtain that, for a constant $\hat{C}_r'''$, independent of $\timescale,\stepsize,b,\invtemp,w$,
\[
d_\mathrm{LP}\left(\law{Y},\law{Z}\right)\leq\hat{C}_r'''\stepsize^{\frac{1}{20}-\frac{9}{200r-20}}\left\{\rbra*{C_RL^7+C_R^3}\rbra*{\sqrt{\log(\stepsize^{-1})} + \stepsize^{-\frac{1}{2}}\invtemp^{-\frac{1}{2}}}+L^{2r}\right\}.
\]

\subsubsection*{Bounding the bounded Wasserstein distance}
Now, by \citet[Corollary 1.3]{barbour2024stein},
\[\label{d_bw_bound1}
d_\mathrm{BW}\left(\law{Y},\law{Z}\right)\leq \EE\|X_{\epsilon}-X\|+\EE\|Z_{\epsilon}-Z\|+2\delta \EE\left\|B_{[0,1]}\right\|+2\delta + 12\epsilon^{-2}\delta^{-2}\kappa_1
\]
for any positive $\delta,\epsilon$. By integration of \cref{z_eps,y_eps}, we obtain that, for any $r>1$ and for some constant $\bar{C}_r$, independent of $\epsilon,\delta,\stepsize,w,\invtemp,b$,
\[
\EE \|Z_{\epsilon}-Z\|\leq \bar{C}_r\epsilon^{\frac{r-1}{2r}},\quad \EE \|Y_{\epsilon}-Y\|\leq \bar{C}_rL^{2r}\epsilon^{\frac{r-1}{2r}} + \bar{C}_r\stepsize^{r-1}\epsilon^{(r-1)\left(\frac{1}{2r}-1\right)}.
\] 
Choosing $\epsilon=\stepsize^{\frac{r}{7r-3}}$ and $\delta=\epsilon^{\frac{r-1}{2r}}$, we obtain that, for a uniform constant $\bar{C}_r'$ (independent of $\stepsize$, $b$, and $\beta$),
\[
d_\mathrm{BW}\left(\law{Y},\law{Z}\right)\leq \bar{C}_r'\stepsize^{\frac{1}{14}-\frac{2}{49r-21}}\left\{\rbra*{C_RL^7+C_R^3}\rbra*{\sqrt{\log(\stepsize^{-1})} + \stepsize^{-\frac{1}{2}}\invtemp^{-\frac{1}{2}}}+L^{2r}\right\}.
\]
\qed
\subsection{Proof of \cref{BDforSMfull}}\label{sec:lemma_levy_proof}
For all $i,j$, let $\rmdbatch{i}{j}(\pmt)=\grad\obsloss{I(i,j)}(\pmt) - \widehat\grad\obsloss{I(i,j)}(\pmt)$
Note that, for all $k\in\{1,\dots,\timescale\},$
\begin{align}
    \pmt_k^2 =& \pmt_{k-1}^2+2\pmt_{k-1}\bigg(-\frac{h}{b}\sum_{i=1}^{b}\Heisbatch{k-1}{i}\pmt_{k-1}+\frac{h}{b}\sum_{i=1}^{b}\Gradbatch{k-1}{i}+ \frac{h}{b}\sum_{i=1}^b\rmdbatch{k-1}{i}(\pmt_{k-1})\\
    &\hspace{10cm}+\sqrt{\frac{2h}{\invtemp}}\noise_{k-1}\bigg)\\
    &+\rbra*{-\frac{h}{b}\sum_{i=1}^{b}\Heisbatch{k-1}{i}\pmt_{k-1}+\frac{h}{b}\sum_{i=1}^{b}\Gradbatch{k-1}{i}+ \frac{h}{b}\sum_{i=1}^b\rmdbatch{k-1}{i}(\pmt_{k-1})+\sqrt{\frac{2h}{\invtemp}}\noise_{k-1}}^2\\
    \le&\pmt_{k-1}^2+2\pmt_{k-1}\Bigg(\rbra*{h\FI -\frac{h}{b}\sum_{i=1}^{b}\Heisbatch{k-1}{i}}\pmt_{k-1}+\frac{h}{b}\sum_{i=1}^{b}\Gradbatch{k-1}{i}\\
    &\hspace{7cm}+\frac{h}{b}\sum_{i=1}^b\rmdbatch{k-1}{i}(\pmt_{k-1})+\sqrt{\frac{2h}{\invtemp}}\noise_{k-1}\Bigg)\\
    &+\rbra*{-\frac{h}{b}\sum_{i=1}^{b}\Heisbatch{k-1}{i}\pmt_{k-1}+\frac{h}{b}\sum_{i=1}^{b}\Gradbatch{k-1}{i}+ \frac{h}{b}\sum_{i=1}^b\rmdbatch{k-1}{i}(\pmt_{k-1})+\sqrt{\frac{2h}{\invtemp}}\noise_{k-1}}^2\\
    \le&2\sum_{j=0}^{k-1}\pmt_j\rbra*{\rbra*{h\FI -\frac{h}{b}\sum_{i=1}^{b}\Heisbatch{j}{i}}\pmt_{j}+\frac{h}{b}\sum_{i=1}^{b}\Gradbatch{j}{i}+ \frac{h}{b}\sum_{i=1}^b\rmdbatch{j}{i}(\pmt_{j})+\sqrt{\frac{2h}{\invtemp}}\noise_{j}}\\
    &+\sum_{j=0}^{k-1}\rbra*{-\frac{h}{b}\sum_{i=1}^{b}\Heisbatch{j}{i}\pmt_{j}+\frac{h}{b}\sum_{i=1}^{b}\Gradbatch{j}{i}+ \frac{h}{b}\sum_{i=1}^b\rmdbatch{j}{i}(\pmt_{j})+\sqrt{\frac{2h}{\invtemp}}\noise_{j}}^2.  
\end{align}
It follows that
\[
\pmt_k^{2r}\le &3^{r-1} \left|2\sum_{j=0}^{k-1}\pmt_j\rbra*{\rbra*{h\FI -\frac{h}{b}\sum_{i=1}^{b}\Heisbatch{j}{i}}\pmt_{j}+\frac{h}{b}\sum_{i=1}^{b}\Gradbatch{j}{i}+\sqrt{\frac{2h}{\invtemp}}\noise_{j}}\right|^r\\
&+3^{r-1}\left|2\sum_{j=0}^{k-1}\pmt_j\frac{h}{b}\sum_{i=1}^b\rmdbatch{j}{i}(\pmt_{j})\right|^r \\
&+3^{r-1}\left\{\sum_{j=0}^{k-1}\rbra*{-\frac{h}{b}\sum_{i=1}^{b}\Heisbatch{j}{i}\pmt_{j}+\frac{h}{b}\sum_{i=1}^{b}\Gradbatch{j}{i}+ \frac{h}{b}\sum_{i=1}^b\rmdbatch{j}{i}(\pmt_{j})+\sqrt{\frac{2h}{\invtemp}}\noise_{j}}^2\right\}^r.  
\]

Now, note that
\[
\sum_{j=0}^{k-1}\pmt_j\rbra*{\rbra*{h\FI -\frac{h}{b}\sum_{i=1}^{b}\Heisbatch{j}{i}}\pmt_{j}+\frac{h}{b}\sum_{i=1}^{b}\Gradbatch{j}{i}+\sqrt{\frac{2h}{\invtemp}}\noise_{j}}
\]
is a martingale and so, by Burkholder--Davis--Gundy, there exists positive constants $C_1$ and $C_2$, such that
\[
\qquad&\EE\max_{k\in\{0,\dots,\timescale\}}\left|\sum_{j=0}^{k-1}\pmt_j\rbra*{\rbra*{h\FI -\frac{h}{b}\sum_{i=1}^{b}\Heisbatch{j}{i}}\pmt_{j}+\frac{h}{b}\sum_{i=1}^{b}\Gradbatch{j}{i}+\sqrt{\frac{2h}{\invtemp}}\noise_{j}}\right|^r\\
\leq &C_1\EE\left\{\left[\sum_{j=0}^{\timescale-1}\pmt_j^2\rbra*{\rbra*{h\FI -\frac{h}{b}\sum_{i=1}^{b}\Heisbatch{j}{i}}\pmt_{j}+\frac{h}{b}\sum_{i=1}^{b}\Gradbatch{j}{i}+\sqrt{\frac{2h}{\invtemp}}\noise_{j}}^2\right]^{r/2}\right\}\\
\leq &C_1\left\{\EE\left[\sum_{j=0}^{\timescale-1}\pmt_j^2\rbra*{\rbra*{h\FI -\frac{h}{b}\sum_{i=1}^{b}\Heisbatch{j}{i}}\pmt_{j}+\frac{h}{b}\sum_{i=1}^{b}\Gradbatch{j}{i}+\sqrt{\frac{2h}{\invtemp}}\noise_{j}}^2\right]^{r}\right\}^{1/2}\\
\leq &3^{r/2}C_1\left\{\EE\left[\sum_{j=0}^{\timescale-1}|\pmt_j|^2\rbra*{\left|\rbra*{h\FI -\frac{h}{b}\sum_{i=1}^{b}\Heisbatch{j}{i}}\pmt_{j}\right|^2+\left|\frac{h}{b}\sum_{i=1}^{b}\Gradbatch{j}{i}\right|^2+\left|\sqrt{\frac{2h}{\invtemp}}\noise_{j}\right|^2}\right]^r\right\}^{1/2}\\
\le & 3^{3r/2-1}C_1\Bigg\{\timescale^{r-1}\EE\Bigg[\sum_{j=0}^{\timescale-1}|\pmt_j|^{2r}\Bigg(\left|\rbra*{h\FI -\frac{h}{b}\sum_{i=1}^{b}\Heisbatch{j}{i}}\pmt_{j}\right|^{2r}+\left|\frac{h}{b}\sum_{i=1}^{b}\Gradbatch{j}{i}\right|^{2r}\\
&\hspace{2cm}+\left|\sqrt{\frac{2h}{\invtemp}}\noise_{j}\right|^{2r}\Bigg)\Bigg]\Bigg\}^{1/2}\\
\le&C_2\left\{\timescale^{r-1}\sum_{j=0}^{\timescale-1}\rbra*{\frac{h^{2r}}{b^r}L^{2r}\EE|\pmt_j|^{4r} + \rbra*{\frac{h^{2r}}{b^r}\max\left\{\EE\left[\left|\Gradbatch{j}{i}\right|^{2r}\right],\Omega\right\}+\rbra*{\frac{2h}{\invtemp}}^{r}}\EE|\pmt_j|^{2r}}\right\}^{1/2},\label{max_theta1}
\]
where the last inequality follows from upper bounds on absolute moments of sums of i.i.d. mean-zero random variables (see e.g. \citep[Theorem 3]{rosenthal1970}).
Moreover, 
\[\label{max_theta2}
\EE\max_{k\in\{0,\dots,\timescale\}}\left|\sum_{j=0}^{k-1}\frac{h}{b}\pmt_j\sum_{i=1}^b\rmdbatch{j}{i}(\pmt_{j})\right|^r\leq C_R^r\timescale^{r-1}h^r\sum_{j=0}^{\timescale-1}\EE|\pmt_j|^{3r},
\]
and, for a positive constant $C_3$,
\[\label{max_theta3}
\qquad\quad &\EE\max_{k\in\{0,\dots,\timescale\}}\left[\sum_{j=0}^{k-1}\rbra*{-\frac{h}{b}\sum_{i=1}^{b}\Heisbatch{j}{i}\pmt_{j}+\frac{h}{b}\sum_{i=1}^{b}\Gradbatch{j}{i}+ \frac{h}{b}\sum_{i=1}^b\rmdbatch{j}{i}(\pmt_{j})+\sqrt{\frac{2h}{\invtemp}}\noise_{j}}^2\right]^r\\
\leq &4^r\timescale^{r-1}\EE\sum_{j=0}^{\timescale-1}\Bigg[\rbra*{-\frac{h}{b}\sum_{i=1}^{b}\Heisbatch{j}{i}\pmt_{j}}^2+\rbra*{\frac{h}{b}\sum_{i=1}^{b}\Gradbatch{j}{i}}^2+ \rbra*{\frac{h}{b}\sum_{i=1}^b\rmdbatch{j}{i}(\pmt_{j})}^2\\
&\hspace{10cm}+\rbra*{\sqrt{\frac{2h}{\invtemp}}\noise_{j}}^2\Bigg]^r\\
\le & C_3\timescale^{r-1} \sum_{j=0}^{\timescale-1}\left[h^{2r}L^{2r}\EE|\pmt_j|^{2r} + \frac{h^{2r}}{b^r}\EE\left[\left|\Gradbatch{1}{1}\right|^{2r}\right] + C_R^{2r}h^{2r}\EE|\pmt_j|^{4r} + \frac{2^r h^r}{\invtemp^r}\right].
\]
Our assumptions imply that  for a uniform constant $C_4$,
\[\label{max_theta4}
\EE|\pmt_j|^{4r}\le C_4\left(\left(\frac{h}{b}\right)^{2r} + \left(\frac{1}{\invtemp}\right)^{2r} \right).
\]
It follows from \cref{max_theta1,max_theta2,max_theta3,max_theta4} that, for a uniform constant $C_5$,
\[
\EE\max_{k\in\{1,\dots,\timescale\}}|\pmt_k|^{2r} \leq &C_5\timescale^{r/2}\Bigg\{\rbra*{1+\frac{L^rh^{r/2}}{b^{r/2}}}\frac{h^{3r/2}}{b^{r}} + \rbra*{1+\frac{L^r}{\invtemp^{r/2}}}\frac{h^{r}}{b^{r/2}\invtemp^{r/2}} + \frac{h^{r/2}}{\invtemp^{r}}   \\
&+ \rbra*{1+\frac{C_R^{2r}h^{2r}+h^rL^r}{b^r}}\frac{\timescale^{r/2}h^{2r}}{b^r}+ \rbra*{1+\frac{C_R^{2r}h^r+h^rL^r}{\invtemp^r}}\frac{\alpha^{r/2}h^{r}}{\invtemp^r}\\
&+\frac{\alpha^{r/2}h^{2r}}{\invtemp^{2r}}+\frac{\alpha^{r/2}h^{r}}{\invtemp^{3r/2}}\Bigg\}.
\]
\qed
\section{Proofs of bound on empirical information matrix convergence rates}
\label{proof_of_bdonfi}
       \begin{proof}
        \[
            \abs{\FI^{(\samplesize)}- \FI^{(\infty)}} 
                & \leq 
                    \abs{\frac{1}{\samplesize} \sum_{i=1}^\samplesize \hess\ell_i(\MLEparm{\samplesize}) - \frac{1}{\samplesize} \sum_{i=1}^\samplesize \hess\ell_i(\trueparm)}
                    + \abs{\frac{1}{\samplesize} \sum_{i=1}^\samplesize \hess\ell_i(\trueparm) - \FI^{(\infty)}}
        \]
        Using \cref{lem:as-maximal-ineq} we have that 
        \[
            \abs{\frac{1}{\samplesize} \sum_{i=1}^\samplesize \hess\ell_i(\MLEparm{\samplesize}) - \frac{1}{\samplesize} \sum_{i=1}^\samplesize \hess\ell_i(\trueparm)}
                \leq C_{\theta} \samplesize^{\frac{1}{c}-b}.
        \]
        all but finitely often with probability 1. 
        
        Let $v_j = \EE[\abs{\hess\ell_1(\trueparm)-\FI^{(\infty)}}^4]$ for $j\in\{2,4\}$. 
        One can verify that
        \[\EE\sbra{\rbra{\frac{1}{\samplesize} \sum_{i=1}^\samplesize \hess\ell_i(\trueparm) - \FI^{(\infty)}}^4} \leq \frac{16 v_4}{\samplesize^3} + \frac{3 v_2^2}{\samplesize^2}.\]
        Thus, by Markov's inequality
        \[
            \Pr\rbra{\abs{\frac{1}{\samplesize} \sum_{i=1}^\samplesize \hess\ell_i(\trueparm) - \FI^{(\infty)}}>\frac{1}{k_\samplesize}}
                & \leq k_\samplesize^4 \rbra{\frac{16 v_4}{\samplesize^3} + \frac{3 v_2^2}{\samplesize^2}}
        \]
        Thus, taking $k_\samplesize = \frac{\samplesize^{1/4}}{(\log(\samplesize))^2}$, this is summable, so 
        $\abs{\frac{1}{\samplesize} \sum_{i=1}^\samplesize \hess\ell_i(\trueparm) - \FI^{(\infty)}}\leq \frac{\log(n)^2}{n^{1/4}}$ all but finitely often with probability 1.
    \end{proof}

    \begin{proof}
        \[
            \abs{\foFI^{(\samplesize)} - \foFI^{(\infty)}} 
                & \leq 
                    \abs{\frac{1}{\samplesize} \sum_{i=1}^\samplesize \grad\ell_i(\MLEparm{\samplesize})^{\otimes 2} - \frac{1}{\samplesize} \sum_{i=1}^\samplesize \grad\ell_i(\trueparm)^{\otimes 2}}
                    + \abs{\frac{1}{\samplesize} \sum_{i=1}^\samplesize \grad\ell_i(\trueparm)^{\otimes 2} - \foFI^{(\infty)}}
        \]
        
        \[
            & \abs{\frac{1}{\samplesize} \sum_{i=1}^\samplesize \grad\ell_i(\MLEparm{\samplesize})^{\otimes 2} - \frac{1}{\samplesize} \sum_{i=1}^\samplesize \grad\ell_i(\trueparm)^{\otimes 2}} \\
                &\quad  \leq \frac{1}{\samplesize} \sum_{i=1}^\samplesize \abs{\grad\ell_i(\MLEparm{\samplesize})^{\otimes 2} - \grad\ell_i(\trueparm)^{\otimes 2}}
        \]
        Now, 
        \[
            \grad\ell_i(\MLEparm{\samplesize})
                & = \grad\ell_i(\trueparm) + \int_\trueparm^\MLEparm{\samplesize} \hess\ell_i(\theta) \dee \theta,
        \]
        so
        \[
        & \grad\ell_i(\MLEparm{\samplesize})^{\otimes 2} - \grad\ell_i(\trueparm)^{\otimes 2}\\
            & \quad = 2 \grad\ell_i(\trueparm) \int_\trueparm^\MLEparm{\samplesize} \hess\ell_i(\theta) \dee \theta + \rbra{\int_\trueparm^\MLEparm{\samplesize} \hess\ell_i(\theta) \dee \theta}^2.
        \]
        Letting $H_i = \norm{\hess\ell_i(\cdot)}_{L^\infty}^d$,
        \[
        & \abs{\grad\ell_i(\MLEparm{\samplesize})^{\otimes 2} - \grad\ell_i(\trueparm)^{\otimes 2}}\\
            & \quad = 2 \grad\ell_i(\trueparm) H_i \abs{\MLEparm{\samplesize} - \trueparm} + H_i^2 \abs{\MLEparm{\samplesize}-\trueparm}^2
        \]
        Using ``average less than max'', and \cref{lem:as-maximal-ineq}, we have that 
        \[
            \abs{\foFI^{(\samplesize)} - \foFI^{(\infty)}} 
                & \leq \max_{i\in\range{\samplesize}} 2 \grad\ell_i(\trueparm) H_i \abs{\MLEparm{\samplesize} - \trueparm} + H_i^2 \abs{\MLEparm{\samplesize}-\trueparm}^2 \\
                & \leq 2 C_{\theta} n^{\frac{1}{8} +\frac{1}{d} -b} + C_{\theta}^2 n^{\frac{2}{d}-2b}                
        \]
        all but finitely often with probability 1. 
        
        Using the same argument as in the proof of \cref{prop:sigma-bound}  
        $\abs{\frac{1}{\samplesize} \sum_{i=1}^\samplesize \grad\ell_i(\trueparm)^{\otimes 2} - \FI^{(\infty)}}\leq \frac{\log(n)^2}{n^{1/4}}$ all but finitely often with probability 1.
    \end{proof}

\section{Proofs of the results from Section \ref{sec:boundforouprocess}}
\label{sec:proof_maxineqou}

\subsection{Proof of \cref{theoremmaxineqouprocesses}}\label{sec:maxou_proof}
Note that the infinitesimal generator of the Ornstein--Uhlenbeck process defined in \cref{ourOUprocessfamily} is given by 
\[
\mathscr{A}_X = -\driftcoeff x \frac{\d}{\d x} + \frac{1}{2}e^{-2\driftcoeff t} \noisefun(t)^2 \frac{\d^2}{\d x^2}.
\]
For any $x \in \R$, let
\[
F_t(x) = e^{\RiccatiSolution(t) x^2}-1,
\]
where $\RiccatiSolution(t) : [0,1] \to \R$ is a positive, differentiable function. Then, we can write
\[\label{eqwithgenerator}
\partial_t F_t + \mathscr{A}_X F_t &= \RiccatiSolution'(t)x^2 e^{\RiccatiSolution(t) x^2} -
                                     2\driftcoeff x^2 \RiccatiSolution(t) e^{\RiccatiSolution(t) x^2} \\
  &\quad +  \frac{1}{2}e^{-2\driftcoeff t}
    \noisefun(t)^2 e^{\RiccatiSolution(t) x^2}\left( 2\RiccatiSolution(t) + 4 \RiccatiSolution(t)^2 x^2
    \right) \\
  &= x^2 e^{\RiccatiSolution(t) x^2} \left[ \RiccatiSolution'(t) - 2\driftcoeff \RiccatiSolution(t) + 2
    e^{-2\driftcoeff t} \noisefun(t)^2\RiccatiSolution(t)^2 \right]  \\
  &\quad + e^{-2\driftcoeff t} \noisefun(t)^2 \RiccatiSolution(t)e^{\RiccatiSolution(t) x^2}.
\]
Note that equating the quantity in the square bracket to zero yields
the Riccati equation (or more specifically the Bernoulli equation)
\[\label{riccatieq}
\RiccatiSolution'(t) - 2\driftcoeff \RiccatiSolution(t) + 2 e^{-2\driftcoeff t} \noisefun(t)^2\RiccatiSolution(t)^2 = 0,
\]
where the unknown function is $\RiccatiSolution(t)$. As $0$ is a particular
solution, this Riccati equation can be solved explicitly by 
\[\label{expressionbetat}
\RiccatiSolution(t) = \frac{e^{2\driftcoeff t}}{\gamma+\int_0^t \noisefun(s)^2 \d s},
\]
where $\gamma >0$ is an integration constant free to be chosen. From now on, let $\RiccatiSolution(t)$ be the above solution to the Riccati equation \cref{riccatieq}, so that 
\[
F_t(x) = e^{\RiccatiSolution(t) x^2}-1 = \on{exp} \left[\frac{e^{2\driftcoeff t}x^2}{\gamma+\int_0^t \noisefun(s)^2 \d s} \right] -1.
\]
As $x \mapsto F_t(x)$ is an even function for all $t \in [0,1]$, we
have $F_t(X_t) = F_t(\abs{X_t})$, and by It\^o's formula, it holds that 
\[\label{itoformulaonFt}
F_t(\abs{X_t}) &= \int_0^t \left( \partial_t F_s + \mathscr{A}_X F_s
  \right)(X_s)\d s + \int_0^t \partial_x F_s(X_s)e^{-\driftcoeff s}\noisefun(s)\d W_s\\
  &= \int_0^t \frac{\noisefun(s)^2}{\gamma+\int_0^s \noisefun(u)^2
    \d u}e^{\RiccatiSolution(t)X_s^2}\d s + \int_0^t \partial_x F_s(X_s)e^{-\driftcoeff s}\noisefun(s)\d W_s,
\]
where the last equality comes from \cref{eqwithgenerator} and the
fact that $\RiccatiSolution(t)$ solves the Riccati equation
\cref{riccatieq}. Let 
\[
I_t = \int_0^t \frac{\noisefun(s)^2}{\gamma+\int_0^s \noisefun(u)^2
    \d u}e^{\RiccatiSolution(s)X_s^2}\d s.
\]
Taking expectation in \cref{itoformulaonFt} and using the optional
sampling theorem yields 
\[\label{propissatisfied}
\EE \left( F_{\tau}(\abs{X_{\tau}}) \right) = \EE \left( I_{\tau} \right)
\]
for all bounded stopping times $\tau$ of $X$. Now, let $H_t(x) =
F_t^{-1}(x)$. It is easily seen that 
\[
H_t(x) = e^{-\driftcoeff t} \sqrt{\gamma+\int_0^t \noisefun(s)^2\d s}\sqrt{\log \left( 1+x \right)}.
\]
 Denoting $H(x) = \sqrt{\log \left( 1+x \right)}$ so that
 \[
 H_t(x) =
 e^{-\driftcoeff t} \sqrt{\gamma+\int_0^t \noisefun(s)^2\d s}H(x),
 \]
 we can write,
 for any $t \in [0,1]$, 
\[\label{firstboundonsup}
\EE \left( \sup_{0\leq s \leq t} \abs{X_s} \right) &=
    \EE \left( \sup_{0\leq s \leq t} H_s(F_s(\abs{X_s}))\right)\\
  &\leq \sup_{0\leq s \leq t} \left\{ e^{-\driftcoeff s} \sqrt{\gamma+\int_0^s \noisefun(u)^2\d u} \right\}\EE \left( \sup_{0\leq s \leq t} H(F_s(\abs{X_s})) \right).
\]
In order to continue, we need the following domination principle
proved in \cite[Proposition 4.7]{revuzyorbook} and
\cite[Proposition 2.1]{graversen_maximal_2000}, which we restate here for convenience.
\begin{proposition}
  \label{dominationprinciple}
Let $\left( \Omega,\mathcal{F}, (\mathcal{F}_{t})_{t \geq 0},
  \mathbb{P} \right)$ be a filtered probability space, let $\left\{ X_t : t \geq 0 \right\}$ be an $(\mathcal{F}_{t})$-adapted non-negative
right-continuous process, let $\left\{ A_t : t \geq 0 \right\}$
be an $(\mathcal{F}_{t})$-adapted increasing continuous process satisfying $A_0 = 0$, and
let $H : \R_+ \to \R_+$ be an increasing continuous function
satisfying $H(0) = 0$. Assume that 
\[
\EE \left( X_{\tau} \right) \leq \EE \left( A_{\tau} \right)
\]
for all bounded $(\mathcal{F}_{t})$-stopping times $\tau$. Then, 
\[
\EE \left(\sup_{0 \leq s \leq \tau} H(X_s) \right) \leq \EE \left( \tilde{H}(A_{\tau}) \right)
\]
for all $(\mathcal{F}_{t})$-stopping times $\tau$, where 
\[
\tilde{H}(x) = x \int_x^{\infty}\frac{1}{s}dH(s) + 2H(x)
\]
for all $x \geq 0$.
\end{proposition}
\noindent Note that \cref{propissatisfied} shows that  \cref{dominationprinciple} is satisfied by
taking $X_t = F_t(\abs{X_t})$ and $A_t = I_t$, so that we can write 
\[
\EE \left( \sup_{0\leq s \leq t} H(F_s(\abs{X_s})) \right) \leq \EE \left( \tilde{H}(I_t)\right),
\]
with $H(x) = \sqrt{\log(1+x)}$. As $\tilde{H}(x) \leq 3 H(x)$ (see
\cite[Equation 2.22]{graversen_maximal_2000}), we get
\[
\EE \left( \sup_{0\leq s \leq t} H(F_s(\abs{X_s})) \right) \leq 3\EE \left( \sqrt{\log(1+I_t)}\right),
\]
and consequently, using \cref{firstboundonsup}, we can write
\[\label{prefinalbound}
\EE \left( \sup_{0\leq s \leq t} \abs{X_s} \right) \leq 3\sup_{0\leq s \leq t} \left\{ e^{-\driftcoeff s} \sqrt{\gamma+\int_0^s \noisefun(u)^2\d u} \right\}\EE \left( \sqrt{\log(1+I_t)}\right).
\]
Now, note that, by It\^o's formula and \cref{riccatieq}, it holds that
\[
\EE  \left(e^{\RiccatiSolution(t)X_t^2}\right) = 1 + \int_0^t \frac{\noisefun(s)^2}{\gamma+\int_0^s\noisefun(u)^2\d u}\EE \left(e^{\RiccatiSolution(s)X_s^2}  \right)\d s,
\]
so that $\EE  \left(e^{\RiccatiSolution(t)X_t^2}\right)$ satisfies a first
order linear ODE and is hence given by 
\[\label{gronwall}
\EE  \left(e^{\RiccatiSolution(t)X_t^2}\right) = \on{exp} \left[\int_0^t
  \frac{\noisefun(s)^2}{\gamma+\int_0^s\noisefun(u)^2\d u}\d s  \right] = 1 +
  \frac{\int_0^t\noisefun(s)^2 \d s}{\gamma},
\]
where the last equality comes from the fact that 
\[
\noisefun(s)^2 =
\frac{d}{\d s}\left[ \gamma+\int_0^s\noisefun(u)^2\d u \right],
\]
so that the
integral can be explicitly computed.
Combining Jensen's inequality and the equality obtained in
\cref{gronwall} allows us to write
\[
\EE \left( \sqrt{\log(1+I_t)}\right) &\leq
  \sqrt{\log(1+\EE \left(I_t\right))}\\
  &= \sqrt{\log \left(1+\int_0^t \frac{\noisefun(s)^2}{\gamma+\int_0^s \noisefun(u)^2
    \d u}\EE \left(e^{\RiccatiSolution(s)X_s^2}\right)\d s\right)}\\
  &= \sqrt{\log \left(1+\int_0^t \frac{\noisefun(s)^2}{\gamma+\int_0^s \noisefun(u)^2
    \d u}\left( 1 +
    \frac{\int_0^s\noisefun(u)^2 \d u}{\gamma} \right)\d s\right)}\\
  &= \sqrt{\log \left(1+ \frac{1}{\gamma}\int_0^t\noisefun(s)^2\d s + \log
    \left( 1+ \frac{1}{\gamma}\int_0^t\noisefun(s)^2\d s \right)\right)},
\]
and finally 
\[
\EE \left( \sup_{0\leq s \leq t} \abs{X_s} \right) &\leq
  3\sup_{0\leq s \leq t} \left\{ e^{-\driftcoeff s} \sqrt{\gamma+\int_0^s
  \noisefun(u)^2\d u} \right\}\\
  &\quad \times \sqrt{\log \left(1+ \frac{1}{\gamma}\int_0^t\noisefun(s)^2\d s + \log
    \left( 1+ \frac{1}{\gamma}\int_0^t\noisefun(s)^2\d s \right)\right)}.
\]
This $L^1$ maximal inequality allows us to apply \cite[Theorem 2.3]{chen_moderate_2021} with 
\[
g(t) &= \sup_{0\leq s \leq t} \left\{ e^{-\driftcoeff s} \sqrt{\gamma+\int_0^s
  \noisefun(u)^2\d u} \right\}\\
  &\qquad\qquad \times \sqrt{\log \left(1+ \frac{1}{\gamma}\int_0^t\noisefun(s)^2\d s + \log
    \left( 1+ \frac{1}{\gamma}\int_0^t\noisefun(s)^2\d s \right)\right)}
\]
and $F(x) = x^p$ for any $p >0$, which yields the desired $L^p$ maximal inequality 
\[
\EE \left( \sup_{0\leq s \leq t} \abs{X_s}^p \right) &\leq
  C_p\sup_{0\leq s \leq t} \left\{  e^{-p\driftcoeff s} \left(\gamma+\int_0^s
  \noisefun(u)^2\d u\right)^{\frac{p}{2}} \right\} \\
  &\quad \times \log^{\frac{p}{2}} \left(1+ \frac{1}{\gamma}\int_0^t\noisefun(s)^2\d s + \log
    \left( 1+ \frac{1}{\gamma}\int_0^t\noisefun(s)^2\d s \right)\right),
\]
where $C_p$ is a constant depending solely on $p$ coming from the $L^1$ maximal inequality in \cite[Theorem 2.3]{chen_moderate_2021}.

\begin{remark}
Note that \cite[Theorem 2.3]{chen_moderate_2021} used in the last step of the above proof is given in the
context of time-homogeneous diffusions whereas we consider a
time-inhomogeneous one. However, a closer inspection of the proof of \cite[Theorem 2.3]{chen_moderate_2021} reveals that it only relies on the strong
Markov property of the process (which our Ornstein--Uhlenbeck \cref{ourOUprocessfamily} process has) and not on the time-homogeneity, which is why \cite[Theorem 2.3]{chen_moderate_2021} also applies in our time-inhomogeneous context. 
\end{remark}

\subsection{Proof of \cref{applicationtooursituation}}\label{sec:aptoz_proof}
Let $a$ and $q(t)$ be as in \cref{theoremmaxineqouprocesses} and
assume there exists an antiderivative $G(t)$ of $\noisefun(t)^2$ with
$G(0)>0$. Then, choosing
the constant $\gamma$ appearing in \cref{eq:mainlpbound} to be $\gamma = G(0)$, we have, for any $t \in [0,1]$,
\[
\gamma + \int_0^t \noisefun(s)^2 \d s = G(t)\quad\mbox{and}\quad 1 + \frac{1}{\gamma}\int_0^t \noisefun(s)^2 \d s = \frac{G(t)}{G(0)}.
\]
Hence, the bound \cref{eq:mainlpbound} of \cref{theoremmaxineqouprocesses} can be rewritten to read 
\[\label{mainboundforapplications123}
  \EE \left( \sup_{0\leq s \leq t} \abs{X_s}^p \right) \leq
  C_p\sup_{0\leq s \leq t} \left\{ \left[e^{-2a s} G(s) \right]^{\frac{p}{2}}  \right\}\log^{\frac{p}{2}} \left(\frac{G(t)}{G(0)} + \log
    \left( \frac{G(t)}{G(0)} \right)\right).
\]
We start with the proof of \eqref{applicationbound1}. Note that $X_t = e^{-Bt}\int_0^t e^{Bs}\left(
  \sqrt{A}-\sqrt{\tilde{A}} \right)\d W_s$ satisfies
\[
  \begin{cases}
    \displaystyle \d X_t = -B X_t \d t + \left(\sqrt{A} - \sqrt{\tilde{A}}\right)\d W_t\\
    \displaystyle X_0 = 0
  \end{cases},
\]
which is the Ornstein-Uhlenbeck process appearing in
\cref{ourOUprocessfamily} with $a=B$ and $\noisefun(t) =
e^{Bt}\left(\sqrt{A} - \sqrt{\tilde{A}}\right)$. Let $G(t)$ denote the antiderivative of $\noisefun(t)^2$ given by 
\[
G(t) = e^{2Bt}\frac{\left(\sqrt{A} - \sqrt{\tilde{A}}\right)^2}{2B},
\]
so that $G(0) = \left(\sqrt{A} - \sqrt{\tilde{A}}\right)^2/2B >0$. We
can hence apply \cref{theoremmaxineqouprocesses} in the form of
\cref{mainboundforapplications123}. As $G(1)/G(0) = e^{2B}$, we
obtain 
\[
\EE \left( \sup_{0\leq t \leq 1} \left\vert e^{-Bt}\int_0^t e^{Bs}\left(
      \sqrt{A}-\sqrt{\tilde{A}} \right)\d W_s\right\rvert ^p \right)
&\leq  \frac{C_p}{2^{p/2}B^{p/2}}\log^{\frac{p}{2}} \left(e^{2B} +
  2B\right).
\]
Using the fact that $\log(1+x) \leq x$ for any $x >-1$ together with
$e^{-2B} \leq 1$, we have
\[
\log \left(e^{2B} +
  2B\right) = 2B + \log \left( 1 + 2B e^{-2B} \right)  \leq 4B,
\]
and hence
\[
\EE \left( \sup_{0\leq t \leq 1} \left\vert e^{-Bt}\int_0^t e^{Bs}\left(
      \sqrt{A}-\sqrt{\tilde{A}} \right)\d W_s\right\rvert ^p \right) \leq
2^{p/2}C_p\abs{\sqrt{A}-\sqrt{\tilde{A}}}^p.
\]
As the mean value theorem ensures that there exists $A^{*}$ in between
$A$ and $\tilde{A}$ such that
\[
\sqrt{A}-\sqrt{\tilde{A}} = \frac{A -
  \tilde{A}}{2 \sqrt{A^{*}}},
\]
we can further write, using the fact
that $A^{*} \geq \min \{A,\tilde{A}\}$, 
\[
\EE \left( \sup_{0\leq t \leq 1} \left\vert e^{-Bt}\int_0^t e^{Bs}\left(
      \sqrt{A}-\sqrt{\tilde{A}} \right)\d W_s\right\rvert ^p \right) \leq
\frac{C_p}{2^{p/2} \min \{A,\tilde{A}\}^{p/2}}\abs{A-\tilde{A}}^p.
\]
We now turn to the proof of \cref{applicationbound2}. In this case, $Y_t = e^{-Bt}\int_0^t
    \sqrt{\tilde{A}}\left(e^{Bs} - e^{\tilde{B}s}
     \right)\d W_s$ satisfies
\[
  \begin{cases}
    \displaystyle \d Y_t = -B Y_t \d t + e^{-Bt}\sqrt{\tilde{A}}\left(e^{Bt} - e^{\tilde{B}t}
     \right)\d W_t\\
    \displaystyle Y_0 = 0
  \end{cases},
\]
which is the Ornstein-Uhlenbeck process appearing in
\cref{ourOUprocessfamily} with $a=B$ and $\noisefun(t) =
\sqrt{\tilde{A}}\left(e^{Bt} - e^{\tilde{B}t}
     \right)$. Let $G(t)$ denote the antiderivative of $\noisefun(t)^2$ given by 
\[
G(t) &= \frac{\tilde{A}}{2B}e^{2Bt} -
\frac{2\tilde{A}}{B+\tilde{B}}e^{(B + \tilde{B})t} +
\frac{\tilde{A}}{2\tilde{B}}e^{2\tilde{B}t}\\
&= \tilde{A} \left(
  \frac{\tilde{B}(B+\tilde{B})e^{2Bt} - 4B\tilde{B}e^{(B+\tilde{B})t}
    + B(B+\tilde{B})e^{2\tilde{B}t}}{2B\tilde{B}(B+\tilde{B})} \right),
\]
so that
\[\label{G0secondpart}
G(0) = \tilde{A} \left(
  \frac{\tilde{B}(B+\tilde{B})- 4B\tilde{B}
    + B(B+\tilde{B})}{2B\tilde{B}(B+\tilde{B})} \right) =
\frac{\tilde{A}(\tilde{B}-B)^2}{2B\tilde{B}(B+\tilde{B})} >0.
\]
We can hence apply \cref{theoremmaxineqouprocesses} in the form of
\cref{mainboundforapplications123}. Observe that
\[
G(t) &= \frac{\tilde{A}e^{2Bt}}{2B\tilde{B}(B+\tilde{B})}\left( (\tilde{B} -
B)^2   - 4B\tilde{B}\left(  e^{(\tilde{B}-B)t} - 1 \right)
+ B(B+\tilde{B})\left(e^{2(\tilde{B}-B)t} -1 \right)\right).
\]
Applying the mean value theorem twice yields
\[
G(t) &= \frac{\tilde{A}e^{2Bt}}{2B\tilde{B}(B+\tilde{B})}\left( (\tilde{B} -
B)^2  +t(\tilde{B}-B) \left(2B(B+\tilde{B})e^{2(\tilde{B}-B)t_1}  - 4B\tilde{B} e^{(\tilde{B}-B)t_2} 
\right)\right),
\]
where $t_1,t_2 \in (0,t)$. We continue by writing
\[
  G(t) &=\frac{\tilde{A}e^{2Bt}}{2B\tilde{B}(B+\tilde{B})} \\
  &\left( (\tilde{B} -
B)^2  +t(\tilde{B}-B) \left(2B(B-\tilde{B})e^{2(\tilde{B}-B)t_1}  +
  4B\tilde{B}\left(   e^{2(\tilde{B}-B)t_1} - e^{2(\tilde{B}-B)\frac{t_2}{2}} \right)
\right)\right).
\]
Applying the mean value theorem once more yields
\[
  G(t) &= \frac{\tilde{A}e^{2Bt}(\tilde{B} -
B)^2}{2B\tilde{B}(B+\tilde{B})} \left( 1 +t
\left(8B\tilde{B}e^{2(\tilde{B}-B)t_3} \left( t_1 - \frac{t_2}{2}
  \right) - 2Be^{2(\tilde{B}-B)t_1}
\right)\right),
\]
where $t_3$ is in between $t_1$ and $t_2/2$. As $t \leq 1$, we can
further write
\[\label{Gtsecondpart}
  G(t) &\leq  \frac{\tilde{A}e^{2Bt}(\tilde{B} -
B)^2}{2B\tilde{B}(B+\tilde{B})} \left( 1 +
\left(8B\tilde{B}e^{2\abs{\tilde{B}-B}}  + 2Be^{2\abs{\tilde{B}-B}}
\right)\right)\\
&\leq \frac{\tilde{A}e^{2Bt}(\tilde{B} -
B)^2}{2B\tilde{B}(B+\tilde{B})} \left( 1 + 2B(1+4\tilde{B})e^{2\abs{\tilde{B}-B}} \right).
\]
Using \cref{G0secondpart} together with $t\leq 1$ immediatly gives us
\[\label{GtG0secondpart}
\frac{G(1)}{G(0)} \leq e^{2B} \left( 1 + 2B(1+4\tilde{B})e^{2\abs{\tilde{B}-B}} \right).
\]
Using \cref{Gtsecondpart} and \cref{GtG0secondpart} in the bound
\cref{mainboundforapplications123} of \cref{theoremmaxineqouprocesses} finally yields
\[
&\EE \left( \sup_{0\leq t \leq 1} \left\vert e^{-Bt}\int_0^t
    \sqrt{\tilde{A}}\left(e^{Bs} - e^{\tilde{B}s}
     \right)\d W_s\right\rvert ^p \right)\\
&\qquad\qquad\qquad\qquad  \leq  C_p \left(
  \frac{\tilde{A}(1+2B(1+4\tilde{B}))(3+4\tilde{B})e^{4 \abs{\tilde{B}-B}}}{\tilde{B}(B+\tilde{B})} \right)^{p/2}\abs{\tilde{B}-B}^p.
\]
It remains to prove \cref{applicationbound3}. Here, $V_t = e^{-\tilde{B}t}\int_0^t
    \sqrt{\tilde{A}} e^{\tilde{B}s}\d W_s$ satisfies
\[
  \begin{cases}
    \displaystyle \d V_t = -\tilde{B} V_t \d t + \sqrt{\tilde{A}}\d W_t\\
    \displaystyle V_0 = 0
  \end{cases},
\]
which is the Ornstein-Uhlenbeck process appearing in
\cref{ourOUprocessfamily} with $a=\tilde{B}$ and $\noisefun(t) =
\sqrt{\tilde{A}}e^{\tilde{B}t}$. Let $G(t)$ denote the antiderivative of $\noisefun(t)^2$ given by 
\[
G(t) = \frac{\tilde{A}}{2\tilde{B}}e^{2\tilde{B}t}
\]
so that $G(0) = \tilde{A}/2\tilde{B}>0$. We can hence apply \cref{theoremmaxineqouprocesses} in the form of
\cref{mainboundforapplications123}. As $G(1)/G(0) = e^{2\tilde{B}}$,
we immediately get 
\[
\EE \left( \sup_{0\leq t \leq 1} \left\vert e^{-\tilde{B}t}\int_0^t
    \sqrt{\tilde{A}} e^{\tilde{B}s}\d W_s\right\rvert ^p \right) \leq
2^{p/2}C_p\tilde{A}^{p/2},
\]
which concludes the proof. \qed

\section{Proofs of the results from Section \ref{sec:main-theorem-proof}}

\subsection{Proof of \cref{lm:pmt}}\label{app:proof_lemma2}

When $k=1$, $\newpmt_1 = \frac{\stepsize}{b}\sum_{i=1}^{b}\Gradbatch{0}{i}+\sqrt{\frac{2h}{\invtemp}}\noise_{0}$. Now assume for $k=m-1$, the lemma is correct. Then
\[\label{Expforpmt}
\newpmt_{m} 
&= \newpmt_{m-1} + \stepsize\,\gradstochlinloss{m-1}(\newpmt_{m-1}) + \sqrt{\frac{2\stepsize}{\invtemp}}\noise_{m-1}\\
&= \newpmt_{m-1}+\frac{h}{b}\rbra*{-\sum_{i=1}^{b}\Heisbatch{m-1}{i}\newpmt_{m-1}+\sum_{i=1}^{b}\Gradbatch{m-1}{i} }+\sqrt{\frac{2h}{\invtemp}}\noise_{m-1}\\
&=\rbra*{1-\frac{h}{b}\sum_{i=1}^{b}\Heisbatch{m-1}{i}}\newpmt_{m-1}+\frac{h}{b}\sum_{i=1}^{b}\Gradbatch{m-1}{i} +\sqrt{\frac{2h}{\invtemp}}\noise_{m-1}\\
& = \rbra*{1-\frac{h}{b}\sum_{i=1}^{b}\Heisbatch{m-1}{i}}\rbra*{\sum_{j=0}^{m-2} Q(j+1,m-1)\rbra*{\frac{h}{b}\sum_{i=1}^{b}\Gradbatch{j}{i}+\sqrt{\frac{2h}{\invtemp}}\noise_{j}}}\\
&\quad+\frac{h}{b}\sum_{i=1}^{b}\Gradbatch{m-1}{i} +\sqrt{\frac{2h}{\invtemp}}\noise_{m-1}\\
&= \rbra*{\sum_{j=0}^{m-2} Q(j+1,m)\rbra*{\frac{h}{b}\sum_{i=1}^{b}\Gradbatch{j}{i}+\sqrt{\frac{2h}{\invtemp}}\noise_{j}}}\\
&\quad+Q(m,m)\rbra*{\frac{h}{b}\sum_{i=1}^{b}\Gradbatch{m-1}{i} +\sqrt{\frac{2h}{\invtemp}}\noise_{m-1}}\\
&= \sum_{j=0}^{m-1} Q(j+1,m)\rbra*{\frac{h}{b}\sum_{i=1}^{b}\Gradbatch{j}{i}+\sqrt{\frac{2h}{\invtemp}}\noise_{j}}.
\]\qed

\subsection{Proof of \cref{BDforM}}\label{app:newpmt}
Using the expression from \cref{lm:pmt}:
\[
\newpmt_k= \sum_{j=0}^{k-1} Q(j+1,k)\rbra*{\frac{h}{b}\sum_{i=1}^{b}\Gradbatch{j}{i}+\sqrt{\frac{2h}{\invtemp}}\noise_{j}};
\]
combined with the fact that $Q(i,j)\le 1$ and independence between terms, we obtain

\[\EE \newpmt_k^2 &= \EE\sum_{j=0}^{k-1}Q(j+1,k)^2 \rbra*{\frac{h}{b}\sum_{i=1}^{b}\Gradbatch{j}{i}+\sqrt{\frac{2h}{\invtemp}}\noise_{j}}^2\\
&=\sum_{j=0}^{k-1}(1-h\FI )^{2(k-j-1)}\rbra*{\frac{h^2 \Omega}{b} + \frac{2h}{\invtemp}}\\
&\le \timescale \rbra*{\frac{h^2 \Omega}{b} + \frac{2h}{\invtemp}}.\]
Moreover,

\[&\EE \newpmt_k^4\\
&\lesssim \EE\sbra*{\sum_{j=0}^{k-1}Q(j+1,k)^4 \rbra*{\frac{h}{b}\sum_{i=1}^{b}\Gradbatch{j}{i}+\sqrt{\frac{2h}{\invtemp}}\noise_{j}}^4}\\
&\quad+ \EE\Bigg[\sum_{0<r<s<k+1}Q(r+1,k)^3\rbra*{\frac{h}{b}\sum_{i=1}^{b}\Gradbatch{r}{i}+\sqrt{\frac{2h}{\invtemp}}\noise_{r}}^3\\
&\qquad\times Q(s+1,k)\rbra*{\frac{h}{b}\sum_{i=1}^{b}\Gradbatch{s}{i}+\sqrt{\frac{2h}{\invtemp}}\noise_{s}}\Bigg]\\
&\quad + \EE\Bigg[\sum_{0<r<s<k+1}Q(r+1,k)^2\rbra*{\frac{h}{b}\sum_{i=1}^{b}\Gradbatch{r}{i}+\sqrt{\frac{2h}{\invtemp}}\noise_{r}}^2\\
&\qquad \times Q(s+1,k)^2\rbra*{\frac{h}{b}\sum_{i=1}^{b}\Gradbatch{s}{i}+\sqrt{\frac{2h}{\invtemp}}\noise_{s}}^2\Bigg]\\
&\quad+ \EE\Bigg[\sum_{0<r<s<t<k+1}Q(r+1,k)^2\rbra*{\frac{h}{b}\sum_{i=1}^{b}\Gradbatch{r}{i}+\sqrt{\frac{2h}{\invtemp}}\noise_{r}}^2 \\
&\qquad \times Q(s+1,k)\rbra*{\frac{h}{b}\sum_{i=1}^{b}\Gradbatch{s}{i}+\sqrt{\frac{2h}{\invtemp}}\noise_{s}} Q(t+1,k)\rbra*{\frac{h}{b}\sum_{i=1}^{b}\Gradbatch{t}{i}+\sqrt{\frac{2h}{\invtemp}}\noise_{t}}\Bigg]\\
&\lesssim \timescale \EE \sbra*{\rbra*{\frac{h}{b}\sum_{i=1}^{b}\Gradbatch{1}{i}+\sqrt{\frac{2h}{\invtemp}}\noise_{1}}^4}+ 4\timescale^2 \EE\sbra*{\Bigg|\frac{h}{b}\sum_{i=1}^{b}\Gradbatch{1}{i}+\sqrt{\frac{2h}{\invtemp}}\noise_{1}\Bigg|^3} \\
&\quad \times \EE\Bigg\{-3 \rbra*{\frac{h}{b}\sum_{i=1}^{b}\Heisbatch{1}{i}}\rbra*{\frac{h}{b}\sum_{i=1}^{b}\Gradbatch{1}{i}}+3\rbra*{\frac{h}{b}\sum_{i=1}^{b}\Heisbatch{1}{i}}^2\rbra*{\frac{h}{b}\sum_{i=1}^{b}\Gradbatch{1}{i}}\\
&\quad -\rbra*{\frac{h}{b}\sum_{i=1}^{b}\Heisbatch{1}{i}}^3\rbra*{\frac{h}{b}\sum_{i=1}^{b}\Gradbatch{1}{i}}\Bigg\}\\
&\quad+3\timescale^2 \cbra*{\EE\sbra*{\rbra*{\frac{h}{b}\sum_{i=1}^{b}\Gradbatch{1}{i}+\sqrt{\frac{2h}{\invtemp}}\noise_{1}}^2}}^2\\
&\quad+ 6\timescale^3 \EE\cbra*{\rbra*{\frac{h}{b}\sum_{i=1}^{b}\Gradbatch{1}{i}+\sqrt{\frac{2h}{\invtemp}}\noise_{1}}^2}\\
&\qquad \times\EE\cbra*{ -2\rbra*{\frac{h}{b}\sum_{i=1}^{b}\Heisbatch{1}{i}}\rbra*{\frac{h}{b}\sum_{i=1}^{b}\Gradbatch{1}{i}}+\rbra*{\frac{h}{b}\sum_{i=1}^{b}\Heisbatch{1}{i}}^2\rbra*{\frac{h}{b}\sum_{i=1}^{b}\Gradbatch{1}{i}}}\\
&\qquad \times\EE\Bigg\{-3 \rbra*{\frac{h}{b}\sum_{i=1}^{b}\Heisbatch{1}{i}}\rbra*{\frac{h}{b}\sum_{i=1}^{b}\Gradbatch{1}{i}}+3\rbra*{\frac{h}{b}\sum_{i=1}^{b}\Heisbatch{1}{i}}^2\rbra*{\frac{h}{b}\sum_{i=1}^{b}\Gradbatch{1}{i}}\\
&\qquad -\rbra*{\frac{h}{b}\sum_{i=1}^{b}\Heisbatch{j}{i}}^3\rbra*{\frac{h}{b}\sum_{i=1}^{b}\Gradbatch{1}{i}}\Bigg\}\\
&\lesssim \timescale\EE\rbra*{\rbra*{\frac{h}{b}\sum_{i=1}^{b}\Gradbatch{1}{i}}^4+\rbra*{\sqrt{\frac{2h}{\invtemp}}\noise_{1}}^4}+4\timescale^2\EE\rbra*{\abs*{\frac{h}{b}\sum_{i=1}^{b}\Gradbatch{1}{i}}^3+\abs*{\sqrt{\frac{2h}{\invtemp}}\noise_{1}}^3}\\
&\times\EE\Bigg\{-\frac{3\stepsize^2}{b}\EE\sbra*{\sigma_1\Gradbatch{1}{1}}+\frac{3\stepsize^3}{b^2}\rbra*{\EE\sbra*{\sigma_1^2\Gradbatch{1}{1}}+2(b-1)\EE\sbra*{\sigma_{I(1,1)}\Gradbatch{1}{1}}\EE\sigma_{I(1,1)}}\\
&\quad-\frac{\stepsize^4}{b^3}\bigg(\EE\sbra*{\sigma_{I(1,1)}^3\Gradbatch{1}{1}}+3(b-1)\EE\sbra*{\sigma_1^2\Gradbatch{1}{1}}\EE\sigma_1+3(b-1)\EE\sbra*{\sigma_1\Gradbatch{1}{1}}\EE\sigma_{I(1,1)}^2\\
&\quad+6(b-1)(b-2)\EE\sbra*{\sigma_{I(1,1)}\Gradbatch{1}{1}}\rbra*{\EE\sigma_{I(1,1)}}^2\bigg)\Bigg\}+3\timescale^2 \rbra*{\frac{h^2 \Omega}{b} + \frac{2h}{\invtemp}}^2\\
&\quad+6\timescale^3\rbra*{\frac{h^2 \Omega}{b} + \frac{3h}{\invtemp}}\times\EE\Bigg\{-\frac{2\stepsize^2}{b}\EE\sbra*{\sigma_1\Gradbatch{1}{1}}+\frac{3\stepsize^3}{b^2}\rbra*{\EE\sbra*{\sigma_1^2\Gradbatch{1}{1}}+(b-1)\EE\sbra*{\sigma_1\Gradbatch{1}{1}}\EE\sigma_1}
\\
&\quad-\frac{\stepsize^4}{b^3}\bigg(\EE\sbra*{\sigma_1^3\Gradbatch{1}{1}}+(b-1)\EE\sbra*{\sigma_1^2\Gradbatch{1}{1}}\EE\sigma_1+(b-1)\EE\sbra*{\sigma_1\Gradbatch{1}{1}}\EE\sigma_1^2\\
&\quad+(b-1)(b-2)\EE\sbra*{\sigma_1\Gradbatch{1}{1}}\left(\EE\sigma_{I(1,1)}\right)^2\bigg)\Bigg\}\\
&\quad\times\EE\cbra*{-\frac{2\stepsize^2}{b}\EE\sigma_1\Gradbatch{1}{1}+\frac{\stepsize^3}{b^2}\rbra*{\EE\sbra*{\sigma_1^2\Gradbatch{1}{1}}+(b-1)\EE\sbra*{\sigma_1\Gradbatch{1}{1}}\EE\sigma_1}}.
\]

Using the fact that $\sigma_{i}\le L$, $\EE|\Gradbatch{1}{1}|\le \sqrt{\Omega}$, we have

\[\label{EQ4}
\EE \newpmt_k^4&\lesssim \timescale\rbra*{\frac{h^4}{b^3}(\EE \Gradbatch{1}{1}^4+b\Omega^2)+\frac{\stepsize^2}{\invtemp^2}}\\
&\quad+\timescale^2 \rbra*{\rbra*{\frac{h^4}{b^3}(\EE \Gradbatch{1}{1}^4+b\Omega^2)}^{3/4}+\rbra*{\frac{\stepsize^2}{\invtemp^2}}^{3/4}}\\
&\quad\times\sqrt{\Omega}\rbra*{\frac{h^2L}{b}+\frac{h^3L^2b}{b^2}+\frac{h^4L^3b^2}{b^3}}+\timescale^2 \rbra*{\frac{h^4 \Omega^2}{b^2}+\frac{h^2}{\invtemp^2}}\\
&\quad+\timescale^3\rbra*{\frac{h^2 \Omega}{b} + \frac{2h}{\invtemp}}
\sqrt{\Omega}\rbra*{\frac{h^2L}{b}+\frac{h^3L^2b}{b^2}+\frac{h^4L^3b^2}{b^3}}\\
&\quad \times\sqrt{\Omega}\rbra*{\frac{2h^2 L}{b}+\frac{h^3 L^2b}{b^2}}\\
&\lesssim \timescale \rbra*{\frac{h^4}{b^2}(\EE \Gradbatch{1}{1}^4+\Omega^2)+\frac{h^2}{
\invtemp^2}}\\
&+\sqrt{\Omega}(1+L^3)\timescale^2 \rbra*{\rbra*{\EE \Gradbatch{1}{1}^4+3\Omega^2}^{3/4} \frac{h^3}{b^{9/4}}+ \frac{h^{3/2}}{\invtemp^{3/2}} }\frac{h^2}{b}\\
&+\timescale^2 \rbra*{\frac{h^4 \Omega^2}{b^2}+\frac{h^2}{\invtemp^2}}+\timescale^3 \Omega L(1+L^4)\rbra*{\frac{h^2 \Omega}{b} + \frac{2h}{\invtemp}}
\frac{h^2 }{b}\times\frac{h^2}{b}\\
&\lesssim  \frac{\timescale^2h^4}{b^2} \Bigg\{\EE \Gradbatch{1}{1}^4+\Omega^2+\sqrt{\Omega}(1+L^3)\sbra*{\rbra*{\EE \Gradbatch{1}{1}^4+\Omega^2}^{3/4}+1}\\
&+\Omega L(1+L^4)\sbra*{\Omega\frac{\timescale h^2}{b}+\frac{\timescale h}{\invtemp}}\Bigg\}+\frac{\timescale^2h^2}{\invtemp^2}+\sqrt{\Omega}(1+L^3) \frac{\timescale^2h^{7/2}}{\invtemp^{3/2}}.
\]\qed

\subsection{Proof of \cref{BDforSM}}\label{app:newpmt1}

Since $\FI  \ge 0$, 
\[
\newpmt_k^2 &= \newpmt_{k-1}^2 + 2\newpmt_{k-1}\rbra*{-\frac{h}{b}\sum_{i=1}^{b}\Heisbatch{k-1}{i}\newpmt_{k-1}+\frac{h}{b}\sum_{i=1}^{b}\Gradbatch{k-1}{i} +\sqrt{\frac{2h}{\invtemp}}\noise_{k-1}}\\
&\quad+\rbra*{-\frac{h}{b}\sum_{i=1}^{b}\Heisbatch{k-1}{i}\newpmt_{k-1}+\frac{h}{b}\sum_{i=1}^{b}\Gradbatch{k-1}{i} +\sqrt{\frac{2h}{\invtemp}}\noise_{k-1}}^2\\
&\le \newpmt_{k-1}^2 + 2\newpmt_{k-1}\rbra*{\rbra*{h\FI -\frac{h}{b}\sum_{i=1}^{b}\Heisbatch{k-1}{i}}\newpmt_{k-1}+\frac{h}{b}\sum_{i=1}^{b}\Gradbatch{k-1}{i} +\sqrt{\frac{2h}{\invtemp}}\noise_{k-1}}\\
&\quad+\rbra*{-\frac{h}{b}\sum_{i=1}^{b}\Heisbatch{k-1}{i}\newpmt_{k-1}+\frac{h}{b}\sum_{i=1}^{b}\Gradbatch{k-1}{i} +\sqrt{\frac{2h}{\invtemp}}\noise_{k-1}}^2\\
&\le 2\sum_{j=0}^{k-1}\newpmt_{j}\rbra*{\rbra*{h\FI -\frac{h}{b}\sum_{i=1}^{b}\Heisbatch{k-1}{i}}\newpmt_{j}+\frac{h}{b}\sum_{i=1}^{b}\Gradbatch{j}{i} +\sqrt{\frac{2h}{\invtemp}}\noise_{j}}\\
&\quad+\sum_{j=0}^{k-1}\rbra*{-\frac{h}{b}\sum_{i=1}^{b}\Heisbatch{j}{i}\newpmt_{j}+\frac{h}{b}\sum_{i=1}^{b}\Gradbatch{j}{i} +\sqrt{\frac{2h}{\invtemp}}\noise_{j}}^2\label{eq:eta_k_2}
\]

Thus we have 
\[\EE \max_{k\in\{0,\cdots,\timescale\}} \newpmt_k^2 \le &2\EE \max_{k\in\{0,\cdots,\timescale\}} \sum_{j=0}^{k-1}\newpmt_{j}\rbra*{\rbra*{h\FI -\frac{h}{b}\sum_{i=1}^{b}\Heisbatch{j}{i}}\newpmt_{j}+\frac{h}{b}\sum_{i=1}^{b}\Gradbatch{j}{i} +\sqrt{\frac{2h}{\invtemp}}\noise_{j}}\\
&\quad +\EE \max_{k\in\{0,\cdots,\timescale\}} \sum_{j=0}^{k-1}\rbra*{-\frac{h}{b}\sum_{i=1}^{b}\Heisbatch{j}{i}\newpmt_{j}+\frac{h}{b}\sum_{i=1}^{b}\Gradbatch{j}{i} +\sqrt{\frac{2h}{\invtemp}}\noise_{j}}^2\]

Note that $\cbra*{\newpmt_{j}\rbra*{\rbra*{h\FI -\frac{h}{b}\sum_{i=1}^{b}\Heisbatch{j}{i}}\newpmt_{j}+\frac{h}{b}\sum_{i=1}^{b}\Gradbatch{j}{i} +\sqrt{\frac{2h}{\invtemp}}\noise_{j}}}_{j=0}^{\timescale}$ is a martingale difference sequence since $\rbra*{h\FI -\frac{h}{b}\sum_{i=1}^{b}\Heisbatch{j}{i}}$, $\frac{h}{b}\sum_{i=1}^{b}\Gradbatch{j}{i}$ and $\sqrt{\frac{2h}{\invtemp}}\noise_{j}$ are all mean zero and independent with $\cbra*{\newpmt_i,I_{(i,\cdot)},\noise_{i}}_{i \le j-1}$. Thus \[\sum_{j=0}^{k-1}\newpmt_{j}\rbra*{\rbra*{h\FI -\frac{h}{b}\sum_{i=1}^{b}\Heisbatch{j}{i}}\newpmt_{j}+\frac{h}{b}\sum_{i=1}^{b}\Gradbatch{j}{i} +\sqrt{\frac{2h}{\invtemp}}\noise_{j}}\] is a martingale.

By Burkholder--Davis--Gundy's inequality, 
\[
&\EE \max_{k\in\{0,\cdots,\timescale\}} \sum_{j=0}^{k-1}\newpmt_{j}\rbra*{\rbra*{h\FI -\frac{h}{b}\sum_{i=1}^{b}\Heisbatch{j}{i}}\newpmt_{j}+\frac{h}{b}\sum_{i=1}^{b}\Gradbatch{j}{i} +\sqrt{\frac{2h}{\invtemp}}\noise_{j}}\\
&\lesssim \EE\cbra*{\sbra*{\sum_{j=0}^{\timescale-1}\newpmt_{j}^2\rbra*{\rbra*{h\FI -\frac{h}{b}\sum_{i=1}^{b}\Heisbatch{j}{i}}\newpmt_{j}+\frac{h}{b}\sum_{i=1}^{b}\Gradbatch{j}{i} +\sqrt{\frac{2h}{\invtemp}}\noise_{j}}^2}^{1/2}}\\
&\lesssim \cbra*{\EE\sum_{j=0}^{\timescale-1}\newpmt_{j}^2\rbra*{\rbra*{h\FI -\frac{h}{b}\sum_{i=1}^{b}\Heisbatch{j}{i}}\newpmt_{j}+\frac{h}{b}\sum_{i=1}^{b}\Gradbatch{j}{i} +\sqrt{\frac{2h}{\invtemp}}\noise_{j}}^2}^{1/2}\\
&\lesssim  \cbra*{\EE\sum_{j=0}^{\timescale-1}\newpmt_{j}^2\sbra*{\rbra*{h\FI -\frac{h}{b}\sum_{i=1}^{b}\Heisbatch{j}{i}}^2\newpmt_{j}^2+\rbra*{\frac{h}{b}\sum_{i=1}^{b}\Gradbatch{j}{i}}^2 +\rbra*{\sqrt{\frac{2h}{\invtemp}}\noise_{j}}^2}}^{1/2}\\
&\lesssim \rbra*{\frac{h^2L^2}{b}\sum_{j=0}^{\timescale-1}\EE\newpmt_{j}^4+\rbra*{\frac{h^2\Omega}{b}+\frac{2h}{\invtemp}}\sum_{j=0}^{\timescale-1}\EE\newpmt_j^2}^{1/2}.
\]

Moreover,
\[
&\EE \max_{k\in\{0,\cdots,\timescale\}} \sum_{j=0}^{k-1}\rbra*{-\frac{h}{b}\sum_{i=1}^{b}\Heisbatch{j}{i}\newpmt_{j}+\frac{h}{b}\sum_{i=1}^{b}\Gradbatch{j}{i} +\sqrt{\frac{2h}{\invtemp}}\noise_{j}}^2\\
&\le \EE \sum_{j=0}^{\timescale-1
}\rbra*{-\frac{h}{b}\sum_{i=1}^{b}\Heisbatch{j}{i}\newpmt_{j}+\frac{h}{b}\sum_{i=1}^{b}\Gradbatch{j}{i} +\sqrt{\frac{2h}{\invtemp}}\noise_{j}}^2\\
&\lesssim \EE\sum_{j=0}^{\timescale-1}\cbra*{\rbra*{-\frac{h}{b}\sum_{i=1}^{b}\Heisbatch{j}{i}}^2\newpmt_j^2+\rbra*{\frac{h}{b}\sum_{i=1}^{b}\Gradbatch{j}{i}}^2+\rbra*{\sqrt{\frac{2h}{\invtemp}}\noise_{j}}^2}\\
&\lesssim h^2L^2\sum_{j=0}^{\timescale-1}\EE\newpmt_j^2 + \frac{\timescale h^2\Omega}{b}+\frac{\timescale h}{\invtemp}.\label{eq:max_eta_k_2}
\]

Thus 
\[\label{BDmt2}
\EE \max_{k\in\{0,\cdots,\timescale\}} \newpmt_k^2 \lesssim & \rbra*{\frac{h^2L^2}{b}\sum_{j=0}^{\timescale-1}\EE\newpmt_{j}^4+\rbra*{\frac{h^2\Omega}{b}+\frac{2h}{\invtemp}}\sum_{j=0}^{\timescale-1}\EE\newpmt_j^2}^{1/2} \\
&\quad +h^2L^2\sum_{j=0}^{\timescale-1}\EE\newpmt_j^2 + \frac{\timescale h^2\Omega}{b}+\frac{\timescale h}{\invtemp}\\
&\lesssim\frac{\timescale^{3/2}h^3L}{b^{3/2}} \Bigg\{\EE \Gradbatch{1}{1}^4+\Omega^2+\sqrt{\Omega}(1+L^3)\sbra*{\rbra*{\EE \Gradbatch{1}{1}^4+\Omega^2}^{3/4}+1}\\
&+\Omega L(1+L^4)\sbra*{\Omega\frac{\timescale h^2}{b}+\frac{\timescale h}{\invtemp}}\Bigg\}^{1/2}+\frac{\timescale^{3/2}h^2L}{\invtemp \batchsize^{1/2}}+\sqrt{\Omega}L(1+L^3) \frac{\timescale^{3/2}h^{11/4}}{\invtemp^{3/4}\batchsize}\\
&\quad +\timescale\rbra*{\frac{h^2\Omega}{b}+\frac{2h}{\invtemp}} +L^2 h^2 \timescale^2 \rbra*{\frac{h^2\Omega}{b}+\frac{h}{\invtemp}}.
\]
Furthermore, it follows from \cref{eq:eta_k_2} that
\[
\newpmt_k^4 \le  &\Bigg\{2\sum_{j=0}^{k-1}\newpmt_{j}\rbra*{\rbra*{h\FI -\frac{h}{b}\sum_{i=1}^{b}\Heisbatch{k-1}{i}}\newpmt_{j}+\frac{h}{b}\sum_{i=1}^{b}\Gradbatch{j}{i} +\sqrt{\frac{2h}{\invtemp}}\noise_{j}}\\
&\quad+\sum_{j=0}^{k-1}\rbra*{-\frac{h}{b}\sum_{i=1}^{b}\Heisbatch{j}{i}\newpmt_{j}+\frac{h}{b}\sum_{i=1}^{b}\Gradbatch{j}{i} +\sqrt{\frac{2h}{\invtemp}}\noise_{j}}^2\Bigg\}^2\\
&\lesssim \cbra*{\sum_{j=0}^{k-1}\newpmt_{j}\rbra*{\rbra*{h\FI -\frac{h}{b}\sum_{i=1}^{b}\Heisbatch{k-1}{i}}\newpmt_{j}+\frac{h}{b}\sum_{i=1}^{b}\Gradbatch{j}{i} +\sqrt{\frac{2h}{\invtemp}}\noise_{j}}}^2\\
&\quad +\cbra*{\sum_{j=0}^{k-1}\rbra*{-\frac{h}{b}\sum_{i=1}^{b}\Heisbatch{j}{i}\newpmt_{j}+\frac{h}{b}\sum_{i=1}^{b}\Gradbatch{j}{i} +\sqrt{\frac{2h}{\invtemp}}\noise_{j}}^2}^2.
\]

With a computation similar to the one that led to \cref{eq:max_eta_k_2}, we have

\[\label{BDmt4}
&\EE \max_{k\in\{0,\cdots,\timescale\}} \newpmt_k^4\\
\lesssim &\frac{h^2L^2}{b}\sum_{j=0}^{\timescale-1}\EE\newpmt_{j}^4+\rbra*{\frac{h^2\Omega}{b}+\frac{2h}{\invtemp}}\sum_{j=0}^{\timescale-1}\EE\newpmt_j^2 \\
&\quad+ \timescale\cbra*{h^4L^4\sum_{j=0}^{\timescale-1}\EE\newpmt_j^4 +\frac{\timescale h^4}{b^3}\rbra*{\EE \Gradbatch{1}{1}^4+3b\Omega^2}+\frac{\timescale h^2}{\invtemp^2}}\\
&\lesssim \rbra*{\frac{1}{\batchsize}+\timescale \stepsize^2L^2}\frac{\timescale^3h^6L^2}{b^2} \Bigg\{\EE \Gradbatch{1}{1}^4+\Omega^2+\sqrt{\Omega}(1+L^3)\sbra*{\rbra*{\EE \Gradbatch{1}{1}^4+\Omega^2}^{3/4}+1}\\
&+\Omega L(1+L^4)\sbra*{\Omega\frac{\timescale h^2}{b}+\frac{\timescale h}{\invtemp}}\Bigg\}+\rbra*{\frac{1}{\batchsize}+\timescale \stepsize^2L^2}\cbra*{\frac{\timescale^3h^4L^2}{\invtemp^2}+\sqrt{\Omega}(1+L^3) \frac{\timescale^3h^{11/2}L^2}{\invtemp^{3/2}}}\\
&+\frac{\timescale^2 h^4}{b^3}\rbra*{\EE \Gradbatch{1}{1}^4+3b\Omega^2}+\frac{\timescale^2 h^2}{\invtemp^2}+\timescale^2 \rbra*{\frac{h^2 \Omega}{b} + \frac{2h}{\invtemp}}^2\\
\lesssim &\rbra*{\frac{1}{\batchsize}+\timescale \stepsize^2L^2}\frac{\timescale^3h^6L^2}{b^2} \Bigg\{\EE \Gradbatch{1}{1}^4+\Omega^2+\sqrt{\Omega}(1+L^3)\sbra*{\rbra*{\EE \Gradbatch{1}{1}^4+\Omega^2}^{3/4}+1}\\
&+\Omega L(1+L^4)\sbra*{\Omega\frac{\timescale h^2}{b}+\frac{\timescale h}{\invtemp}}\Bigg\}+\rbra*{\frac{1}{\batchsize}+\timescale \stepsize^2L^2}\cbra*{\frac{\timescale^3h^4L^2}{\invtemp^2}+\sqrt{\Omega}(1+L^3) \frac{\timescale^3h^{11/2}L^2}{\invtemp^{3/2}}}\\
&+\frac{\timescale^2 h^4}{b^3}\rbra*{\EE \Gradbatch{1}{1}^4+b\Omega^2+b\Omega}+\frac{\timescale^2 h^2}{\invtemp^2}.&
\]\qed

\subsection{Proof of \cref{BDforR}}
\label{app:diffY}
Note that, by an argument similar to \cref{Expforpmt}, for $k\geq 1$,
\[\pmt_{\iter}= Q(0,k)\pmt_0 +\sum_{j=0}^{k-1} Q(j+1,k)\cbra*{\frac{h}{b}\sum_{i=1}^{b}\Gradbatch{j}{i}+\sqrt{\frac{2h}{\invtemp}}\noise_{j}+\stepsize\sbra*{\gradstochloss{j}(\pmt_j)-\gradstochlinloss{j}(\pmt_j)}},\]
where notation $Q(j,k)$ comes from \cref{eq:Q_def},
and, thus
\[Y_t -\newy_t = \spacescale \rbra*{\pmt_{\floor{\timescale t}}-\newpmt_{\floor{\timescale t}}} =
\begin{cases}
\spacescale\stepsize\sum_{j=0}^{\floor{\timescale t}-1} Q(j+1,\floor{\timescale t})\rbra*{\gradstochloss{j}(\pmt_j)-\gradstochlinloss{j}(\pmt_j)},&\text{if }t\geq \frac{1}{\timescale}\\
0,&\text{otherwise.}
\end{cases}
\]
First we use the mean value theorem and H\"older's inequality to bound $\abs{\EE g(\newy) -\EE g(Y)}$. With similar arguments as in \cite[page 21-22]{dobler2021stein}, we have
\[\abs{\EE g(Y)-\EE g(\newy)}\le \|g\|_{M} \cbra*{\EE\|Y-\newy\|+2\EE\|Y-\newy\|^3+2(\EE\|\newy\|^3)(\EE\|Y-\newy\|^3)^{1/3}}. \label{eq:proof_lemma5_1}
\]
Now, note that, by \cref{asmp:stepsize,asmp:hosmooth}, $Q(i,j)\le 1$ and $\abs{\gradstochloss{j}(\pmt_j)-\gradstochlinloss{j}(\pmt_j)} \le C_R \pmt_j^2$ and so 
\[\EE\|Y-\newy\|\le  wh C_R \sum_{j=0}^{\timescale-1}\EE\pmt_j^2,\qquad\EE\|Y-\newy\|^3\le\EE\cbra*{\rbra*{wh C_R\sum_{j=0}^{\timescale-1}\pmt_j^2}^3}.
\]
It follows from \cref{asmp:bounded-iterates} that 
\[\EE \pmt_j^2 &\lesssim K_1\rbra*{\frac{h\Omega}{b}+\frac{1}{\invtemp}};\\
\EE \pmt_j^6 &\lesssim K_3^3 h^3\sbra*{\EE\rbra*{\frac{1}{b}\sum_{i=1}^{b}\Gradbatch{j}{i}}^6+\frac{1}{h^3\invtemp^3}}\\
&\lesssim K_3^3 h^3 \sbra*{\frac{\Omega^3}{b^3} +\frac{(\EE \Gradbatch{1}{1}^4)^{3/2}+\Omega\EE \Gradbatch{1}{1}^4}{b^4}+\frac{\EE\Gradbatch{1}{1}^6}{b^5} +\frac{1}{h^3\invtemp^3}}\\
&\lesssim K_3^3\sbra*{\frac{h^3}{b^3}\rbra*{\Omega^3+(\EE \Gradbatch{1}{1}^4)^{3/2}+\EE \Gradbatch{1}{1}^4\Omega+\EE\Gradbatch{1}{1}^6} +\frac{1}{\invtemp^3}}.\]
Thus
\[&\EE\|Y-\newy\|\lesssim  w\timescale K_1C_R\rbra*{\frac{h^2\Omega}{b}+\frac{h}{\invtemp}};\\
&\EE\|Y-\newy\|^3 \lesssim  w^3 \timescale^3 C_R^3 K_3^3 \sbra*{\frac{h^6}{b^3}\rbra*{\Omega^3+(\EE \Gradbatch{1}{1}^4)^{3/2}+\EE \Gradbatch{1}{1}^4\Omega+\EE\Gradbatch{1}{1}^6} +\frac{ h^3}{\invtemp^3}}.\label{eq:proof_lemma5_2}\]
Applying \cref{BDforSM} and noting that $\EE\|\newy\|^4 = \EE\|w\newpmt_{\floor{\timescale t}}\|^4$,
we have \[\EE \|\newy\|^3 &\le \rbra*{\EE\|\newy\|^4}^{3/4}\\&
\lesssim w^3\Bigg\{\frac{\timescale^2 \stepsize^4}{\batchsize^2}\sbra*{\frac{\timescale\stepsize^2L^2}{b}E_1+\timescale^2\stepsize^4L^4E_1+\frac{\EE \Gradbatch{1}{1}^4}{b}+\Omega^2+\Omega}\\
&+\frac{\timescale^2\stepsize^2}{\invtemp^2}\sbra*{1+\frac{\timescale\stepsize^2L^2}{\batchsize}+\timescale^2\stepsize^4L^4}+\rbra*{\frac{1}{\batchsize}+\timescale \stepsize^2L^2}\sqrt{\Omega}(1+L^3) \frac{\timescale^3h^{11/2}L^2}{\invtemp^{3/2}}\Bigg\}^{3/4}, \label{eq:proof_lemma5_3}
\]
where
\[
E_1=\EE \Gradbatch{1}{1}^4+\Omega^2+\sqrt{\Omega}(1+L^3)\sbra*{\rbra*{\EE \Gradbatch{1}{1}^4+\Omega^2}^{3/4}+1}+\Omega L(1+L^4)\sbra*{\Omega\frac{\timescale h^2}{b}+\frac{\timescale h}{\invtemp}}.
\]
Thus, it follows from \cref{eq:proof_lemma5_1,eq:proof_lemma5_2,eq:proof_lemma5_3} that

\[
&\abs{\EE g(\newy) -\EE g(Y)}\\ &\lesssim \|g\|_M \Bigg\{w\timescale K_1C_R\rbra*{\frac{h^2\Omega}{b}+\frac{h}{\invtemp}}\\
&\quad+ w^3 \timescale^3 C_R^3 K_3^3 \rbra*{\frac{h^6}{b^3}\rbra*{\Omega^3+(\EE \Gradbatch{1}{1}^4)^{3/2}+\EE \Gradbatch{1}{1}^4\Omega+\EE\Gradbatch{1}{1}^6} +\frac{h^3}{\invtemp^3}}\\
&\quad +w^3\Bigg\{\frac{\timescale^2 \stepsize^4}{\batchsize^2}\sbra*{\frac{\timescale\stepsize^2L^2}{b}E_1+\timescale^2\stepsize^4L^4E_1+\frac{\EE \Gradbatch{1}{1}^4}{b}+\Omega^2+\Omega}\\
&\quad+\frac{\timescale^2\stepsize^2}{\invtemp^2}\sbra*{1+\frac{\timescale\stepsize^2L^2}{\batchsize}+\timescale^2\stepsize^4L^4}+\rbra*{\frac{1}{\batchsize}+\timescale \stepsize^2L^2}\sqrt{\Omega}(1+L^3) \frac{\timescale^3h^{11/2}L^2}{\invtemp^{3/2}}\Bigg\}^{3/4} \\
&\qquad  \times w\timescale C_R K_3 \sbra*{\frac{32h^6}{b^3}\rbra*{\Omega^3+(\EE \Gradbatch{1}{1}^4)^{3/2}+\EE \Gradbatch{1}{1}^4\Omega+\EE\Gradbatch{1}{1}^6} +\frac{h^3}{\invtemp^3}}^{1/3}\Bigg\}.
\]
\qed

\subsection{Proof of \cref{BDforZ}}
\label{app:diffZ}
As in \cite[pages 21-22]{dobler2021stein}, we apply the mean value theorem to obtain
\[\label{eq:bound_z_decomp}
|\EE g(Z)-\EE g(\newz)| \le \|g\|_{M} \left\{\EE  \|Z-\newz\|+2\EE \|Z-\newz\|^3+2(\EE \|\newz\|^3)(\EE \|Z-\newz\|^3)^{1/3}\right\}.
\]
Thus we need to bound $\EE  \|Z-\newz\|$,  $\EE \|Z-\newz\|^3$ and $\EE \|\newz\|^3.$ 
Throughout the proof, we will use the notation:
\[
B := \timescale \stepsize \FI; \qquad A:= \frac{w^2 \alpha h^2}{b} \Omega  + \frac{ w^2 \alpha h}{\invtemp}.
\]

\subsubsection{Bounding $\EE\|\newz\|^3$}
From \cite[Theorem 2.2]{jia2020moderate} we know that for some constant $C_z>0$, 
\[
\EE\|\newz\|^3 \le  \frac{C_zA^{3/2}}{B^{3/2}}\log^{3/2}(1+B)=& \frac{C_z}{(\timescale\stepsize \FI )^{3/2}}\rbra*{\frac{w^2 \alpha h^2}{b} \Omega  + \frac{ w^2 \alpha h}{\invtemp}}^{3/2}\log^{3/2}(1+\timescale\stepsize \FI )\\
=&\frac{C_z}{\FI ^{3/2}}\rbra*{\frac{w^2 h}{b} \Omega  + \frac{ w^2 }{\invtemp}}^{3/2}\log^{3/2}(1+\timescale\stepsize \FI ).\label{eq:bound_third_z}\]

\subsubsection{Bounding $\EE  \|Z-\newz\|$}
Note that
\[
&\EE  \|Z-\newz\|\\
= &\EE  \sup_{t\in[0,1]}\left|Z_t-\newzt\right|\\
=&\EE  \sup_{t\in[0,1]}\left|\sqrt{A}\int_{0}^{t}e^{-B(t-s)}\d W_s - \sqrt{A}\int_{0}^{\frac{\floor{\timescale t}}{\timescale }}e^{-B(\frac{\floor{\timescale t}}{\timescale }-s)}\d W_s \right|\\
\le &\EE  \sup_{t\in[0,1]}\left|\int_{0}^{\frac{\floor{\timescale t}}{\timescale }}\sqrt{A}e^{-B(t-s)}\d W_s-\sqrt{A}e^{-B(\frac{\floor{\timescale t}}{\timescale }-s)}\d W_s\right|\\
&+\EE  \sup_{t\in[0,1]}\left|\int_{\frac{\floor{\timescale t}}{\timescale }}^{t}\sqrt{A}e^{-B(t-s)}\d W_s\right|\\
= &\EE  \sup_{t\in[0,1]}\left|e^{-Bt}-e^{-B\frac{\floor{\timescale t}}{\timescale }}\right| \left|\int_{0}^{\frac{\floor{\timescale t}}{\timescale }} \sqrt{A}e^{Bs}\d W_s
\right|+\EE  \sup_{t\in[0,1]}\left|\int_{\frac{\floor{\timescale t}}{\timescale }}^{t}\sqrt{A}e^{-B(t-s)}\d W_s\right|.
\]

By the local 1-Lipschitz property of $e^{x}$, 
we obtain

\[
\EE  \sup_{t\in[0,1]}\left|e^{-Bt}-e^{-B\frac{\floor{\timescale t}}{\timescale }}\right|\left|\int_{0}^{\frac{\floor{\timescale t}}{\timescale }} \sqrt{A}e^{Bs}\d W_s
\right|\le &\frac{B}{\timescale}\EE  \sup_{t\in[0,1]}\left|\int_{0}^{\frac{\floor{\timescale t}}{\timescale }} \sqrt{A}e^{-B(\frac{\floor{\timescale t}}{\timescale}-s)}\d W_s
\right|\\
=& \frac{B}{\timescale}\EE\|\newz\|\le \frac{B}{\timescale}\rbra*{\EE\|\newz\|^3}^{1/3}.
\]
Moreover, using the fact that $e^{-B(t-s)} \le 1,$ for $s\in\rbra*{\frac{\floor{\timescale t}}{\timescale},t}$, and \citep[Lemma 3]{mod2008},

\[
&\EE\sup_{t\in[0,1]}\left|\int_{\frac{\floor{\timescale t}}{\timescale }}^{t}\sqrt{A}e^{-B(t-s)}\d W_s\right|\le\EE\sup_{t\in[0,1]}\left|\int_{\frac{\floor{\timescale t}}{\timescale }}^{t}\sqrt{A}\d W_s\right|\le \frac{30}{\sqrt{\pi \log 2}} \rbra*{\frac{A\log(2\timescale)}{\timescale}}^{1/2}.\]
So, plugging in $B = \timescale \stepsize \FI $ and $A = \frac{w^2 \alpha h^2}{b} \Omega  + \frac{ w^2 \alpha h}{\invtemp}$ and using \cref{eq:bound_third_z}, we obtain
\[
&\EE\|Z-\newz\|\le \sqrt{\frac{w^2 h}{b} \Omega  + \frac{ w^2 }{\invtemp}} \rbra*{\stepsize \sqrt{C_z \FI \log(1+\timescale\stepsize \FI )}+\frac{30}{\sqrt{\pi \log 2}}\sqrt{h \log(2\timescale)}}.\label{eq:bound_first_dif}
\]
\subsubsection{Bounding $\EE  \|Z-\newz\|^3$}
Similarly,
\[
&\EE  \|Z-\newz\|^3\\
=& \EE  \sup_{t\in[0,1]}\left|Z_t-\newzt\right|^3\\
=&\EE  \sup_{t\in[0,1]}\left|\sqrt{A}\int_{0}^{t}e^{-B(t-s)}\d W_s - \sqrt{A}\int_{0}^{\frac{\floor{\timescale t}}{\timescale }}e^{-B(\frac{\floor{\timescale t}}{\timescale }-s)}\d W_s \right|^3\\
\le& 4\EE  \sup_{t\in[0,1]}\left|\int_{0}^{\frac{\floor{\timescale t}}{\timescale }}\sqrt{A}e^{-B(t-s)}\d W_s-\sqrt{A}e^{-B(\frac{\floor{\timescale t}}{\timescale }-s)}\d W_s\right|^3\\
&+4\EE  \sup_{t\in[0,1]}\left|\int_{\frac{\floor{\timescale t}}{\timescale }}^{t}\sqrt{A}e^{-B(t-s)}\d W_s\right|^3\\
=&4\EE  \sup_{t\in[0,1]}\left|e^{-Bt}-e^{-B\frac{\floor{\timescale t}}{\timescale }}\right|^3\left|\int_{0}^{\frac{\floor{\timescale t}}{\timescale }} \sqrt{A}e^{Bs}\d W_s\right|^3 +4\EE  \sup_{t\in[0,1]}\left|\int_{\frac{\floor{\timescale t}}{\timescale }}^{t}\sqrt{A}e^{-B(t-s)}\d W_s\right|^3.
\]

By the local 1-Lipschitz property of $e^{x}$,

\[
\EE  \sup_{t\in[0,1]}\left|e^{-Bt}-e^{-B\frac{\floor{\timescale t}}{\timescale }}\right|^3 \left|\int_{0}^{\frac{\floor{\timescale t}}{\timescale }} \sqrt{A}e^{Bs}\d W_s
\right|^3\le & \frac{B^3}{\timescale^3}\EE \sup_{t\in[0,1]}\left|\int_{0}^{\frac{\floor{\timescale t}}{\timescale }} \sqrt{A}e^{-B(\frac{\floor{\timescale t}}{\timescale }-s)}\d W_s
\right|^3\\
=& \frac{B^3}{\timescale^3}\EE\|\newz\|^3.
\]
Now, using the fact that $e^{-B(t-s)} \le 1,$ for $s\in(\frac{\floor{\timescale t}}{\timescale},t)$ and \citep[Lemma 3]{mod2008}, we obtain
\[
&\EE\sup_{t\in[0,1]}\left|\int_{\frac{\floor{\timescale t}}{\timescale }}^{t}\sqrt{A}e^{-B(t-s)}\d W_s\right|^3\le\EE\sup_{t\in[0,1]}\left|\int_{\frac{\floor{\timescale t}}{\timescale }}^{t}\sqrt{A}\d W_s\right|^3\le \frac{1080}{\sqrt{\pi \log^3 2}} \rbra*{\frac{A\log(2\timescale)}{\timescale}}^{3/2}
.\]
So, plugging in $B = \timescale \stepsize \FI $ and $A = \frac{w^2 \alpha h^2}{b} \Omega  + \frac{ w^2 \alpha h}{\invtemp}$ and using \cref{eq:bound_third_z},
\[
&\EE  \|Z-\newz\|^3\\
\le &\rbra*{\frac{w^2 h}{b} \Omega  + \frac{ w^2 }{\invtemp}}^{3/2}\rbra*{4C_z\stepsize^3 \FI ^{3/2} \log^{3/2}(1+\timescale \stepsize \FI )+ \frac{1080}{\sqrt{\pi \log^3 2}}\rbra*{h \log(2\timescale)}^{3/2}}.\label{eq:bound_third_dif}
\]
\subsubsection{Conclusion}
Combining \cref{eq:bound_z_decomp,eq:bound_first_dif,eq:bound_third_z,eq:bound_third_dif} together, and using the fact that $\log(1+x) \le x$, we have 
\[
&|\EE g(Z)-\EE g(\newz)|\\
\le& \|g\|_M\Bigg\{\sqrt{\frac{w^2 h}{b} \Omega  + \frac{ w^2 }{\invtemp}} \rbra*{h^{3/2} \timescale^{1/2}\sqrt{C_z}\FI  +\frac{30}{\sqrt{\pi \log 2}}\sqrt{h \log(2\timescale)}}\\
&+ 2\rbra*{\frac{w^2 h}{b} \Omega  + \frac{ w^2 }{\invtemp}}^{3/2}\rbra*{4C_z\stepsize^{9/2}\timescale^{3/2} \FI ^{3} + \frac{1080}{\sqrt{\pi \log^3 2}}\rbra*{h \log(2\timescale)}^{3/2}}\\
&+2C_z\timescale^{3/2}h^{3/2}\rbra*{\frac{w^2 h}{b} \Omega  + \frac{ w^2 }{\invtemp}}^{2}\rbra*{4^{1/3}C_z^{1/3} \FI h^{3/2}\timescale^{1/2} +6\rbra*{\frac{25}{\pi \log^3 2}}^{1/6}\sqrt{h \log(2\timescale)} }
\Bigg\}\\
\lesssim& \|g\|_M\Bigg\{\frac{\spacescale\timescale^{1/2}\stepsize^2\Omega^{1/2}\Sigma}{\batchsize^{1/2}}+\frac{\spacescale\timescale^{1/2}\stepsize^{3/2}\Sigma}{\invtemp^{1/2}}+\frac{\spacescale\stepsize\Omega^{1/2}\sqrt{\log(2\timescale)}}{\batchsize^{1/2}}+\frac{\spacescale\stepsize^{1/2}\sqrt{\log(2\timescale)}}{\invtemp^{1/2}}\\
&+\frac{\spacescale\timescale^{3/2}\stepsize^6\Omega^{3/2}\Sigma^3}{\batchsize^{3/2}}+\frac{\spacescale^3\timescale^{3/2}\stepsize^{9/2}\Sigma^3}{\invtemp^{3/2}}+\frac{\spacescale^3\stepsize^3\Omega^{3/2}\log^{3/2}(2\timescale)}{\batchsize^{3/2}}+\frac{\spacescale^3\stepsize^{3/2}\log^{3/2}(2\timescale)}{\invtemp^{3/2}}\\
&+\frac{\spacescale^4\timescale^{2}\stepsize^{5}\Omega^2\Sigma}{\batchsize^2}+\frac{\spacescale^4\timescale^{3/2}\stepsize^{4}\Omega^2\sqrt{\log(2\timescale)}}{\batchsize^2}+\frac{\spacescale^4\timescale^{2}\stepsize^{3}\Sigma}{\invtemp^2}+\frac{\spacescale^4\timescale^{3/2}\stepsize^{2}\sqrt{\log(2\timescale)}}{\invtemp^2}\Bigg\}.\qed
\]

\subsection{Proof of \cref{lm:difY}}\label{app:proof_lemma7}
Applying the update rule for $\newpmt$ and $\exparm$ recursively, we have, for $m\geq 1$,
\[
\lefteqn{\newpmt_{K+m}-\exparm_{K+m}} \\
&=\newpmt_{K+m-1} - \rbra*{\frac{h}{b}\sum_{i=1}^{b}\Heisbatch{K+m-1}{i}}\newpmt_{K+m-1}+\frac{h}{b}\sum_{i=1}^{b}\Gradbatch{K+m-1}{i} +\sqrt{\frac{2h}{\invtemp}}\noise_{K+m-1}\\
&\phantom{=~}-\rbra*{\exparm_{K+m-1} - \rbra*{\frac{h}{b}\sum_{i=1}^{b}\Heisbatch{K+m-1}{i}}\exparm_{K+m-1}+\frac{h}{b}\sum_{i=1}^{b}\Gradbatch{K+m-1}{i} +\sqrt{\frac{2h}{\invtemp}}\noise_{K+m-1}}\\
&= \rbra*{1-\frac{h}{b}\sum_{i=1}^{b}\Heisbatch{K+m-1}{i}}\rbra*{\newpmt_{K+m-1}-\exparm_{K+m-1}}\\
&= Q(K+1,K+m)(\newpmt_{K+1}-\exparm_{K+1}).
\]
Moreover,
\[
\lefteqn{\newpmt_{K+1}-\exparm_{K+1}} \\
&= \newpmt_{K} - \rbra*{\frac{h}{b}\sum_{i=1}^{b}\Heisbatch{K}{i}}\newpmt_{K}+\frac{h}{b}\sum_{i=1}^{b}\Gradbatch{K}{i} +\sqrt{\frac{2h}{\invtemp}}\noise_{K}\\
&\quad- \rbra*{\newpmt_K - \rbra*{\frac{h}{b}\sum_{i=1}^{b}\swapHeisbatch{K}{i}}\newpmt_{K}+\frac{h}{b}\sum_{i=1}^{b}\swapGradbatch{K}{i} +\sqrt{\frac{2h}{\invtemp}}\xi'_{K}}\\
&=\frac{h}{b}\rbra*{\rbra*{-\sum_{i=1}^{b}\Heisbatch{K}{i} +\sum_{i=1}^{b}\swapHeisbatch{K}{i} }\newpmt_K + \sum_{i=1}^{b}\Gradbatch{K}{i} -\sum_{i=1}^{b} \swapGradbatch{K}{i}}+\sqrt{\frac{2h}{\invtemp}}\rbra*{\noise_{K}-\xi'_{K}}.
\]\qed

\subsection{Proof of \cref{lemma_error_cov}}\label{app:proof_lemma10}

\subsubsection{Decomposition of the error $\error{cov}$}
First, in order to prove \cref{lemma_error_cov}, we use a simpler notation for each part of $\newy-\exy$. 
From \cref{difY},
\[
\newyt-\exyt &= w\sum_{m=k+1}^{\timescale}Q(k+1,m)(\newpmt_{k+1}-\exparm_{k+1})\mathbb{I}_{[\frac{m}{\alpha
},\frac{m+1}{\alpha})}(t)\\
&=\underbrace{w\sum_{m=k+1}^{\timescale}\frac{h}{b}Q(k+1,m)\rbra*{-\sum_{i=1}^{b}\Heisbatch{k}{i} +\sum_{i=1}^{b}\swapHeisbatch{k}{i} }\newpmt_k\mathbb{I}_{[\frac{m}{\alpha
},\frac{m+1}{\alpha})}(t)}_{\Sigma}\\
&\quad+\underbrace{w\sum_{m=k+1}^{\timescale}\frac{h}{b}Q(k+1,m)\rbra*{\sum_{i=1}^{b}\Gradbatch{k}{i} -\sum_{i=1}^{b} \swapGradbatch{k}{i}}\mathbb{I}_{[\frac{m}{\alpha
},\frac{m+1}{\alpha})}(t)}_{\Psi}\\
&\quad+\underbrace{w\sum_{m=k+1}^{\timescale}\sqrt{\frac{2h}{\invtemp}}Q(k+1,m)\rbra*{\noise_{k}-\xi'_{k}}\mathbb{I}_{[\frac{m}{\alpha
},\frac{m+1}{\alpha})}(t)}_{\noise}.
\]

And

\[
&\EE D^2\steintestfun(\newy)[\newz,\newz]\\
= &\underbrace{\frac{w^2 h \Omega}{2b\FI}\sum_{m,r=1}^{\timescale}\rbra*{e^{-h \FI  |m-r|}-e^{-h \FI  (m+r)}}\EE \FD^2 f(\newy)\sbra*{\mathbb{I}_{[\frac{m}{\alpha},\frac{m+1}{\alpha})}(t),\mathbb{I}_{[\frac{r}{\alpha},\frac{r+1}{\alpha})}}(t)}_{z_G}\\
&+\underbrace{\frac{\spacescale^2} {\invtemp \FI}\sum_{m,r=1}^{\timescale}\rbra*{e^{-h \FI  |m-r|}-e^{-h \FI  (m+r)}}\EE \FD^2 f(\newy)\sbra*{\mathbb{I}_{[\frac{m}{\alpha},\frac{m+1}{\alpha})}(t),\mathbb{I}_{[\frac{r}{\alpha},\frac{r+1}{\alpha})}}(t)}_{z_\noise}.
\]

Thus the quadratic form can be decomposed in this way:
\[
&\EE \cbra*{\FD^2 f(\newy)[(\newy-\exy)\lambda,\newy-\exy]-D^2\steintestfun(\newy)[\newz,\newz]}\\
&= \error{cov.\Sigma\Sigma} +\error{cov.\Sigma\Psi}+\error{cov.\Psi\Psi-z_G}+\error{cov.\Sigma\noise}+\error{cov.\Psi\noise}+\error{cov.\noise\noise-z_\noise},\label{eq:quad_decompos}
\]
where
\[
\error{cov.\Sigma\Sigma}
:=&\Bigg|\EE\frac{1}{\timescale}\sum_{k=0}^{\timescale -1}\frac{\timescale \spacescale^2 \stepsize^2}{2b^2}\\
&\times\EE^{\newy} \FD^2 f(\newy)\Bigg[\sum_{m=k+1}^{\timescale}Q(k+1,m)\rbra*{-\sum_{i=1}^{b}\Heisbatch{k}{i} +\sum_{i=1}^{b}\swapHeisbatch{k}{i} }\newpmt_k\mathbb{I}_{[\frac{m}{\timescale
},\frac{m+1}{\timescale})}(t),\\
&\hspace{2cm}\sum_{r=k+1}^{\timescale}Q(k+1,r)\rbra*{-\sum_{i=1}^{b}\Heisbatch{k}{i} +\sum_{i=1}^{b}\swapHeisbatch{k}{i} }\newpmt_k\mathbb{I}_{[\frac{r}{\timescale
},\frac{r+1}{\timescale})}(t)\Bigg]\Bigg|;\\
\error{cov.\Sigma\Psi}
:=&\Bigg|\EE\frac{1}{\timescale}\sum_{k=0}^{\timescale -1}\frac{\timescale \spacescale^2 \stepsize^2}{b^2}\rbra*{-\sum_{i=1}^{b}\Heisbatch{k}{i}\sum_{i=1}^{b}\Gradbatch{k}{i} +b\FI (\sum_{i=1}^{b}\Gradbatch{k}{i})-b \sigma_{\Sigma\Psi} }\newpmt_k \\
& \times \FD^2 f(\newy)\Bigg[\sum_{m=k+1}^{\timescale}Q(k+1,m)\mathbb{I}_{[\frac{m}{\timescale
},\frac{m+1}{\timescale})}(t),\sum_{r=k+1}^{\timescale}Q(k+1,r)\mathbb{I}_{[\frac{r}{\timescale
},\frac{r+1}{\timescale})}(t)\Bigg]\Bigg|;\\
\error{cov.\Psi\Psi-z_G}
&:= \Bigg|\EE\Bigg\{\frac{1}{2\timescale}\sum_{k=0}^{\timescale -1}\frac{\timescale \spacescale^2 \stepsize^2}{b^2}\\
&\qquad\times\EE^{\newy} \FD^2 f(\newy)\Bigg[\sum_{m=k+1}^{\timescale}Q(k+1,m)\rbra*{\sum_{i=1}^{b}\Gradbatch{k}{i} -\sum_{i=1}^{b} \swapGradbatch{k}{i}}\mathbb{I}_{[\frac{m}{\timescale
},\frac{m+1}{\timescale})}(t),\\
&\qquad\qquad\qquad\qquad\sum_{r=k+1}^{\timescale}Q(k+1,r)\rbra*{\sum_{i=1}^{b}\Gradbatch{k}{i} -\sum_{i=1}^{b} \swapGradbatch{k}{i}}\mathbb{I}_{[\frac{r}{\timescale
},\frac{r+1}{\timescale})}(t)\Bigg]\\
&-\frac{w^2 h \Omega}{2b\FI}\sum_{m,r=1}^{\timescale}\rbra*{e^{-h \FI  |m-r|}-e^{-h \FI  (m+r)}}\EE \FD^2 f(\newy)\sbra*{\mathbb{I}_{[\frac{m}{\alpha},\frac{m+1}{\alpha})},\mathbb{I}_{[\frac{r}{\alpha},\frac{r+1}{\alpha})}}\Bigg\}\Bigg|;\\
\error{cov.\Sigma\noise}:=&\Bigg|\EE\frac{1}{\timescale}\sum_{k=0}^{\timescale -1}\frac{\sqrt{2}\timescale \spacescale^2 \stepsize^{3/2}}{b\invtemp^{1/2}}\EE^{\newy} \FD^2 f(\newy)\Bigg[\sum_{m=k+1}^{\timescale}Q(k+1,m)\rbra*{\noise_{k}-\xi'_{k}}\mathbb{I}_{[\frac{m}{\timescale
},\frac{m+1}{\timescale})}(t),\\
&\hspace{2cm}\sum_{r=k+1}^{\timescale}Q(k+1,r)\rbra*{-\sum_{i=1}^{b}\Heisbatch{k}{i} +\sum_{i=1}^{b}\swapHeisbatch{k}{i} }\newpmt_k\mathbb{I}_{[\frac{r}{\timescale
},\frac{r+1}{\timescale})}(t)\Bigg]\Bigg|;\\
\error{cov.\Psi\noise}:=&\frac{1}{\timescale}\sum_{k=0}^{\timescale -1}\frac{\sqrt{2}\timescale \spacescale^2 \stepsize^{3/2}}{b\invtemp^{1/2}}\EE^{\newy} \FD^2 f(\newy)\Bigg[\sum_{m=k+1}^{\timescale}Q(k+1,m)\rbra*{\noise_{k}-\xi'_{k}}\mathbb{I}_{[\frac{m}{\timescale
},\frac{m+1}{\timescale})}(t),\\
&\hspace{3cm}\sum_{r=k+1}^{\timescale}Q(k+1,r)\rbra*{\sum_{i=1}^{b}\Gradbatch{k}{i} -\sum_{i=1}^{b} \swapGradbatch{k}{i}}\mathbb{I}_{[\frac{r}{\timescale
},\frac{r+1}{\timescale})}(t)\Bigg];\\
\error{cov.\noise\noise-z_\noise}:=&\Bigg|\EE\Bigg\{\frac{1}{2\timescale}\sum_{k=0}^{\timescale -1}\frac{2\timescale \spacescale^2 \stepsize}{\invtemp}\EE^{\newy} \FD^2 f(\newy)\Bigg[\sum_{m=k+1}^{\timescale}Q(k+1,m)\rbra*{\noise_{k}-\xi'_{k}}\mathbb{I}_{[\frac{m}{\timescale
},\frac{m+1}{\timescale})}(t),\\
&\hspace{6cm}\sum_{r=k+1}^{\timescale}Q(k+1,r)\rbra*{\noise_{k}-\xi'_{k}}\mathbb{I}_{[\frac{r}{\timescale
},\frac{r+1}{\timescale})}(t)\Bigg]\\
&\hphantom{~=\Bigg|\EE\Bigg\{}-\frac{\spacescale^2} {\invtemp \FI}\sum_{m,r=1}^{\timescale}\rbra*{e^{-h \FI  |m-r|}-e^{-h \FI  (m+r)}}\EE \FD^2 f(\newy)\sbra*{\mathbb{I}_{[\frac{m}{\alpha},\frac{m+1}{\alpha})},\mathbb{I}_{[\frac{r}{\alpha},\frac{r+1}{\alpha})}}\Bigg\}\Bigg|.
\]

These summands are bounded in the following subsections. In bounding each of these summands, among other tools, we will repeatedly apply \cref{f_properties}.

\subsubsection{Bounding $\error{cov.\Sigma\Sigma}$}\label{app:bounding_err_cov_sig_sig}
Since $\sigma_i \le L$, note that
\[
&\error{cov.\Sigma\Sigma}\\
=&\Bigg|\EE\frac{1}{\timescale}\sum_{k=0}^{\timescale -1}\frac{\timescale \spacescale^2 \stepsize^2}{2b^2}\EE^{\newy} \rbra*{\sum_{i=1}^{b}\Heisbatch{k}{i} -\sum_{i=1}^{b}\swapHeisbatch{k}{i} }^2\newpmt_k^2 \\
&\qquad \times \FD^2 f(\newy)\Bigg[\sum_{m=k+1}^{\timescale}Q(k+1,m)\mathbb{I}_{[\frac{m}{\timescale
},\frac{m+1}{\timescale})}(t),\sum_{r=k+1}^{\timescale}Q(k+1,r)\mathbb{I}_{[\frac{r}{\timescale
},\frac{r+1}{\timescale})}(t)\Bigg]\Bigg|\\
&\le \frac{1}{\timescale}\sum_{k=0}^{\timescale -1}\Bigg|\EE\frac{\timescale \spacescale^2 \stepsize^2}{2b^2} 4b^2L^2\newpmt_k^2 \\
&\qquad\times\FD^2 f(\newy)\Bigg[\sum_{m=k+1}^{\timescale}Q(k+1,m)\mathbb{I}_{[\frac{m}{\timescale
},\frac{m+1}{\timescale})}(t),\sum_{r=k+1}^{\timescale}Q(k+1,r)\mathbb{I}_{[\frac{r}{\timescale
},\frac{r+1}{\timescale})}(t)\Bigg]
\Bigg|.\]

Therefore, since $Q(i,j)\le 1,\forall i,j$, using that $\FD^2 f(\newy)[f,g] \le \|\FD^2 f(\newy)\|\sup|fg|$ and Cauchy-Schwartz and \cref{BDforM},
\[\label{eq:cov1_bound}
\error{cov.\Sigma\Sigma}
\le& 2\spacescale^2 \stepsize^2 L^2 \sum_{k=0}^{\timescale -1}\EE\sbra*{\newpmt_k^2 \|\FD^2 f(\newy)\|}\\
\le& 2\spacescale^2 \stepsize^2 L^2 \sqrt{\EE\|\FD^2 f(\newy)\|^2}\sum_{k=0}^{\timescale -1}\sqrt{\EE\newpmt_k^4}\\
\le& 2\spacescale^2 \stepsize^2 L^2 \|g\|_M \sum_{k=0}^{\timescale -1}\sqrt{\EE\newpmt_k^4}\sqrt{\EE\rbra*{1+\frac{1}{3}\|\newy\| +\frac{1}{3}\EE\|\newz\|^2}^2}\\
\le& 2D_2\spacescale^2 \timescale^2\stepsize^3L^2 \|g\|_M \Bigg\{\frac{h}{b} \Bigg[\EE \Gradbatch{1}{1}^4+\Omega^2+\sqrt{\Omega}(1+L^3)\sbra*{\rbra*{\EE \Gradbatch{1}{1}^4+\Omega^2}^{3/4}+1}\\
&+\Omega L(1+L^4)\sbra*{\Omega\frac{\timescale h^2}{b}+\frac{\timescale h}{\invtemp}}\Bigg]^{1/2}+\frac{1}{\invtemp}+\Omega^{1/4}\sqrt{1+L^3}\frac{h^{3/4}}{\invtemp^{3/4}}\Bigg\}.
\]
where $D_2$ is a bound for $\sqrt{\EE\rbra*{1+\frac{1}{3}\|\newy\| +\frac{1}{3}\EE\|\newz\|^2}^2}$ which will be given in \cref{D1D2}. 

\subsubsection{Bounding $\error{cov.\Sigma\Psi}$} \label{app:bounding_err_cov_sig_psi}
First, note that
\[
&\frac{1}{\timescale}\sum_{k=0}^{\timescale -1}\frac{\timescale \spacescale^2 \stepsize^2}{b^2}\EE^{\newy} \FD^2 f(\newy)\Bigg[\sum_{m=k+1}^{\timescale}Q(k+1,m)\rbra*{\sum_{i=1}^{b}\Gradbatch{k}{i} -\sum_{i=1}^{b} \swapGradbatch{k}{i}}\mathbb{I}_{[\frac{m}{\timescale
},\frac{m+1}{\timescale})}(t),\\
&\hphantom{~\frac{1}{\timescale}\sum_{k=0}^{\timescale -1}\frac{\timescale \spacescale^2 \stepsize^2}{b^2}\EE^{\newy} \FD^2 f(\newy)\Bigg[}\sum_{r=k+1}^{\timescale}Q(k+1,r)\rbra*{-\sum_{i=1}^{b}\Heisbatch{k}{i} +\sum_{i=1}^{b}\swapHeisbatch{k}{i} }\newpmt_k\mathbb{I}_{[\frac{r}{\timescale
},\frac{r+1}{\timescale})}(t)\Bigg]\\
&=\frac{1}{\timescale}\sum_{k=0}^{\timescale -1}\frac{\timescale \spacescale^2 \stepsize^2}{b^2}\EE^{\newy} \rbra*{\sum_{i=1}^{b}\Gradbatch{k}{i} -\sum_{i=1}^{b} \swapGradbatch{k}{i}}\rbra*{-\sum_{i=1}^{b}\Heisbatch{k}{i} +\sum_{i=1}^{b}\swapHeisbatch{k}{i} }\newpmt_k\\
&\qquad \times \FD^2 f(\newy)\Bigg[\sum_{m=k+1}^{\timescale}Q(k+1,m)\mathbb{I}_{[\frac{m}{\timescale
},\frac{m+1}{\timescale})}(t),\qquad\sum_{r=k+1}^{\timescale}Q(k+1,r)\mathbb{I}_{[\frac{r}{\timescale
},\frac{r+1}{\timescale})}(t)\Bigg]\\
&=\frac{1}{\timescale}\sum_{k=0}^{\timescale -1}\frac{\timescale \spacescale^2 \stepsize^2}{b^2} \rbra*{-\sum_{i=1}^{b}\Heisbatch{k}{i}\sum_{i=1}^{b}\Gradbatch{k}{i} +b\FI \rbra*{\sum_{i=1}^{b}\Gradbatch{k}{i}}-b \sigma_{\Sigma\Psi} }\newpmt_k \\
&\qquad\times  \FD^2 f(\newy)\Bigg[\sum_{m=k+1}^{\timescale}Q(k+1,m)\mathbb{I}_{[\frac{m}{\timescale
},\frac{m+1}{\timescale})}(t),
\sum_{r=k+1}^{\timescale}Q(k+1,r)\mathbb{I}_{[\frac{r}{\timescale
},\frac{r+1}{\timescale})}(t)\Bigg]
,\]
where $\sigma_{\Sigma\Psi} \defas \EE \psi_1\sigma_1$

Since $Q(i,j)\le 1,\forall i,j$, using that $\FD^2 f(\newy)[f,g] \le \|\FD^2 f(\newy)\|\sup|fg|$ and Cauchy-Schwartz and \cref{BDforM}, the expectation of this term 
\[\label{eq:cov2_bound}
&\hspace{1cm}\error{cov.\Sigma\Psi}\\
&\le \frac{1}{\timescale}\sum_{k=0}^{\timescale -1}\frac{\timescale \spacescale^2 \stepsize^2}{b^2}\EE\Bigg|\rbra*{-\sum_{i=1}^{b}\Heisbatch{k}{i}\sum_{i=1}^{b}\Gradbatch{k}{i} +b\FI (\sum_{i=1}^{b}\Gradbatch{k}{i})-b \sigma_{\Sigma\Psi} }\newpmt_k\Bigg|\|\FD^2 f(\newy)\|\\
&\le \frac{1}{\timescale}\sum_{k=0}^{\timescale -1}\frac{\timescale \spacescale^2 \stepsize^2}{b^2}\sqrt{\EE\|\FD^2 f(\newy)\|^2}\\
&\qquad\qquad\times\sqrt{\EE \rbra*{\sum_{i=1}^{b}\Heisbatch{k}{i}\sum_{i=1}^{b}\Gradbatch{k}{i} -b\FI (\sum_{i=1}^{b}\Gradbatch{k}{i})+b \sigma_{\Sigma\Psi} }^2\newpmt_k^2}\\
&=\frac{1}{\timescale}\sum_{k=0}^{\timescale -1}\frac{\timescale \spacescale^2 \stepsize^2}{b^2}\sqrt{\EE\|\FD^2 f(\newy)\|^2}\\
&\qquad\qquad\times\sqrt{\EE \rbra*{\sum_{i=1}^{b}\Heisbatch{k}{i}\sum_{i=1}^{b}\Gradbatch{k}{i} -b\FI (\sum_{i=1}^{b}\Gradbatch{k}{i})+b \sigma_{\Sigma\Psi} }^2\EE\newpmt_k^2}\\
&\lesssim \frac{\timescale \spacescale^2 \stepsize^2}{b^2}\sqrt{\EE\|\FD^2 f(\newy)\|^2}\\
&\quad\times\sqrt{\rbra*{\EE \rbra*{\sum_{i=1}^{b}\Heisbatch{k}{i}\sum_{i=1}^{b}\Gradbatch{k}{i}}^2 +b^2\FI ^2\rbra*{\sum_{i=1}^{b}\Gradbatch{k}{i}}^2+b^2 \sigma_{\Sigma\Psi}^2} } \sqrt{\timescale\rbra*{\frac{h^2 \Omega}{b} + \frac{2h}{\invtemp}}}\\
&\lesssim \frac{\timescale^{3/2} \spacescale^2 \stepsize^2}{b^2}\sqrt{\EE\|\FD^2 f(\newy)\|^2}\sqrt{\rbra*{\rbra*{b^3\Omega L^2+2b^2\Omega L^2} +b^3L^2\Omega+b^2 L^2\Omega}}\sqrt{\frac{h^2 \Omega}{b} + \frac{2h}{\invtemp}}\\
&\lesssim \frac{\timescale^{3/2} \spacescale^2 \stepsize^2 \sqrt{\Omega}L \|g\|_M}{b^{1/2}}\sqrt{\frac{h^2 \Omega}{b} + \frac{2h}{\invtemp}}\sqrt{\EE\rbra*{1+\frac{1}{3}\|\newy\| +\frac{1}{3}\EE\|\newz\|^2}^2}\\
&\lesssim \frac{ D_2 \timescale^{3/2} \spacescale^2 \stepsize^2 \sqrt{\Omega}L \|g\|_M}{b^{1/2}}\sqrt{\frac{h^2 \Omega}{b} + \frac{2h}{\invtemp}},
\]
where $D_2$ is derived in \cref{D1D2}.

\subsubsection{Bounding $\error{cov.\Psi\Psi-z_G}$}\label{8.3}

First, note that
\[
&\frac{1}{2\timescale}\sum_{k=0}^{\timescale -1}\frac{\timescale \spacescale^2 \stepsize^2}{b^2}\EE^{\newy} \FD^2 f(\newy)\Bigg[\sum_{m=k+1}^{\timescale}Q(k+1,m)\rbra*{\sum_{i=1}^{b}\Gradbatch{k}{i} -\sum_{i=1}^{b} \swapGradbatch{k}{i}}\mathbb{I}_{[\frac{m}{\timescale
},\frac{m+1}{\timescale})}(t),\\
&\hphantom{~\frac{1}{2\timescale}\sum_{k=0}^{\timescale -1}\frac{\timescale \spacescale^2 \stepsize^2}{b^2}\EE^{\newy} \FD^2 f(\newy)\Bigg[}\sum_{r=k+1}^{\timescale}Q(k+1,r)\rbra*{\sum_{i=1}^{b}\Gradbatch{k}{i} -\sum_{i=1}^{b} \swapGradbatch{k}{i}}\mathbb{I}_{[\frac{r}{\timescale
},\frac{r+1}{\timescale})}(t)\Bigg]\\
&=\frac{1}{2\timescale}\sum_{k=0}^{\timescale -1}\frac{\timescale \spacescale^2 \stepsize^2}{b^2}\EE^{\newy} \rbra*{\sum_{i=1}^{b}\Gradbatch{k}{i} -\sum_{i=1}^{b} \swapGradbatch{k}{i}}\rbra*{\sum_{i=1}^{b}\Gradbatch{k}{i} -\sum_{i=1}^{b} \swapGradbatch{k}{i}}\\
&\qquad\times\FD^2 f(\newy)\Bigg[\sum_{m=k+1}^{\timescale}Q(k+1,m)\mathbb{I}_{[\frac{m}{\timescale
},\frac{m+1}{\timescale})}(t),\sum_{r=k+1}^{\timescale}Q(k+1,r)\mathbb{I}_{[\frac{r}{\timescale
},\frac{r+1}{\timescale})}(t)\Bigg]\\
&=\frac{1}{2\timescale}\sum_{k=0}^{\timescale -1}\frac{\timescale \spacescale^2 \stepsize^2}{b^2} \rbra*{\rbra*{\sum_{i=1}^{b}\Gradbatch{k}{i}}^2 +b\Omega } \\
&\qquad\times\FD^2 f(\newy)\Bigg[\sum_{m=k+1}^{\timescale}Q(k+1,m)\mathbb{I}_{[\frac{m}{\timescale
},\frac{m+1}{\timescale})}(t),\sum_{r=k+1}^{\timescale}Q(k+1,r)\mathbb{I}_{[\frac{r}{\timescale
},\frac{r+1}{\timescale})}(t)\Bigg].
\]

Therefore, can decompose the error $\error{cov.\Psi\Psi-z_G} $ into two terms:
\[
\error{cov.\Psi\Psi-z_G}=\error{cov.\Psi\Psi-z_G.1}+\error{cov.\Psi\Psi-z_G.2},
\]
where
\[\label{2b-zz_first}
&\error{cov.\Psi\Psi-z_G.1}:=\EE\Bigg\{\frac{1}{2\timescale}\sum_{k=0}^{\timescale -1}\frac{\timescale \spacescale^2 \stepsize^2}{b^2} \rbra*{\rbra*{\sum_{i=1}^{b}\Gradbatch{k}{i}}^2 -b\Omega }\\
&\quad \times\FD^2 f(\newy)\Bigg[\sum_{m=k+1}^{\timescale}Q(k+1,m)\mathbb{I}_{[\frac{m}{\timescale
},\frac{m+1}{\timescale})}(t),\sum_{r=k+1}^{\timescale}Q(k+1,r)\mathbb{I}_{[\frac{r}{\timescale
},\frac{r+1}{\timescale})}(t)\Bigg]\Bigg\}
\]
and
\[
&\error{cov.\Psi\Psi-z_G.2}\\
:=&\frac{1}{\timescale}\sum_{k=0}^{\timescale -1}\frac{\timescale \spacescale^2 \stepsize^2}{b^2} \rbra*{b\Omega }\\
&\qquad\qquad\times\EE \FD^2 f(\newy)\Bigg[\sum_{m=k+1}^{\timescale}Q(k+1,m)\mathbb{I}_{[\frac{m}{\timescale
},\frac{m+1}{\timescale})}(t),\sum_{r=k+1}^{\timescale}Q(k+1,r)\mathbb{I}_{[\frac{r}{\timescale
},\frac{r+1}{\timescale})}(t)\Bigg]\\
&\qquad -\frac{w^2 h \Omega}{2b\FI}\sum_{m,r=1}^{\timescale}\rbra*{e^{-h \FI  |m-r|}-e^{-h \FI  (m+r)}}\EE \FD^2 f(\newy)\sbra*{\mathbb{I}_{[\frac{m}{\alpha},\frac{m+1}{\alpha})},\mathbb{I}_{[\frac{r}{\alpha},\frac{r+1}{\alpha})}}\\
&= \frac{1}{\timescale}\sum_{k=0}^{\timescale -1}\frac{\timescale \spacescale^2 \stepsize^2 \Omega}{b} \sum_{m,r=k+1}^{\timescale}\EE Q(k+1,m)Q(k+1,r)\FD^2 f(\newy)\Bigg[\mathbb{I}_{[\frac{m}{\timescale
},\frac{m+1}{\timescale})}(t),\mathbb{I}_{[\frac{r}{\timescale
},\frac{r+1}{\timescale})}(t)\Bigg]\\
&\qquad -\frac{w^2 h\Omega}{2b\FI}\sum_{m,r=1}^{\timescale} \rbra*{e^{-h \FI  |m-r|}-e^{-h \FI  (m+r)}}\EE \FD^2 f(\newy)\sbra*{\mathbb{I}_{[\frac{m}{\alpha},\frac{m+1}{\alpha})},\mathbb{I}_{[\frac{r}{\alpha},\frac{r+1}{\alpha})}}.\label{2b-zz}\]

\subsubsection*{Controlling $\error{cov.\Psi\Psi-z_G.1}$}
To bound the first term, $\error{cov.\Psi\Psi-z_G.1}$ from \cref{2b-zz_first}, define 
\[
\newyt^{k\Psi} \defas \newyt&-\spacescale\sum_{m=k+1}^{\timescale}Q(k+1,m)\rbra*{\frac{\stepsize}{b}\sum_{i=1}^{b}\Gradbatch{k}{i}}\mathbb{I}_{[\frac{m}{\timescale
},\frac{m+1}{\timescale})}(t) \\
&+ \spacescale \frac{\frac{h}{b}\sum_{i=1}^{b}\Heisbatch{k}{i}}{1-\frac{h}{b}\sum_{i=1}^{b}\Heisbatch{k}{i}}\sum_{m=k+1}^{\timescale}\sum_{j=0}^{k-1}Q(j+1,m)\rbra*{\frac{h}{b}\sum_{i=1}^{b}\Gradbatch{j}{i}+\sqrt{\frac{2h}{\invtemp}}\noise_{j}}\ind{\cointer{\frac{m}{\samplesize},\frac{m+1}{\samplesize}}}.\label{58}\]
Then $\newyt^{k\Psi}$ is independent with  $\Gradbatch{k}{i},\forall i=1,\cdots,b$.

Thus using the fact that $\FD^2 f(\newy)[A_1,A_2] \le \|\FD^2 f(\newy)\|\|A_1\|\|A_2\|$, together with Cauchy-Schwarz, \cref{f_properties} and \cref{asmp:stepsize}, 
\[&\error{cov.\Psi\Psi-z_G.1}\\
&=\Bigg|\EE\frac{1}{2\timescale}\sum_{k=0}^{\timescale -1}\frac{\timescale \spacescale^2 \stepsize^{2}}{b^2}\rbra*{\rbra*{\sum_{i=1}^{b}\Gradbatch{k}{i}}^2 -b\Omega } \\
&\quad\times\rbra*{\FD^2 f(\newy)-\FD^2 f(\newy^{k\Psi})}\Bigg[\sum_{m=k+1}^{\timescale}Q(k+1,m)\mathbb{I}_{[\frac{m}{\timescale
},\frac{m+1}{\timescale})}(t),\sum_{r=k+1}^{\timescale}Q(k+1,r)\mathbb{I}_{[\frac{r}{\timescale
},\frac{r+1}{\timescale})}(t)\Bigg]\Bigg|\\
&\le \frac{1}{2\timescale}\sum_{k=0}^{\timescale -1}\EE\Bigg|\frac{\timescale \spacescale^2 \stepsize^{2}}{b^2}\rbra*{\rbra*{\sum_{i=1}^{b}\Gradbatch{k}{i}}^2 -b\Omega }\Bigg|\|\FD^2 f(\newy)-\FD^2 f(\newy^{k\Psi})\|\\
&\le\frac{1}{2\timescale}\sum_{k=0}^{\timescale -1}\frac{\timescale \spacescale^3 \stepsize^{2}}{\batchsize^2}\EE\Bigg\{\Bigg|\rbra*{\sum_{i=1}^{b}\Gradbatch{k}{i}}^2-\batchsize\Omega\Bigg|\rbra*{\frac{1}{2}+\frac{1}{3}\|\newy^{k\Psi}\|+\frac{1}{3}\EE\|\hat{Z}\|^2}\\
&\times\rbra*{\abs*{\frac{\stepsize}{b}\sum_{i=1}^{b}\Gradbatch{k}{i}}+\frac{\stepsize L}{1-\stepsize L}\sum_{j=0}^{k-1}\abs*{\frac{h}{b}\sum_{i=1}^{b}\Gradbatch{j}{i}+\sqrt{\frac{2h}{\invtemp}}\noise_{j}}}\Bigg\}\\
&\le\frac{\|g\|_M}{2\timescale}\sum_{k=0}^{\timescale -1}\frac{\timescale \spacescale^3 \stepsize^{2}}{\batchsize^2}\sqrt{\EE\rbra*{\frac{1}{2}+\frac{1}{3}\|\newy^k\|+\frac{1}{3}\EE\|\hat{Z}\|^2}^2}\Bigg\{\EE \abs*{\frac{\stepsize}{b}\rbra*{\sum_{i=1}^{b}\Gradbatch{k}{i}}^3-\stepsize\Omega\sum_{i=1}^{b}\Gradbatch{k}{i}}\\
&+\frac{\timescale\stepsize L}{1-\stepsize L}\EE\abs*{\rbra*{\sum_{i=1}^{b}\Gradbatch{k}{i}}^2-\batchsize\Omega}\sqrt{\EE\rbra*{\frac{h}{b}\sum_{i=1}^{b}\Gradbatch{1}{i}+\sqrt{\frac{2h}{\invtemp}}\noise_{1}}^2}\Bigg\}\\
&\lesssim \frac{\timescale \spacescale^3 \stepsize^{2}D_2\|g\|_M}{\batchsize^2}\\
&\quad\times\Bigg\{\frac{\stepsize}{\batchsize}\rbra*{b\EE \Gradbatch{1}{1}^4+\rbra*{2b^2-3b}\Omega^2}^{3/4} +\stepsize \batchsize^{1/2} \Omega^{3/2}+\frac{\timescale\stepsize L}{1-\stepsize L} \batchsize \Omega\sqrt{\rbra*{\frac{\Omega\stepsize^2}{\batchsize}+\frac{\stepsize}{\invtemp}}} \Bigg\}\\
&\lesssim  \frac{\timescale \spacescale^3 \stepsize^{3}D_2\|g\|_M}{\batchsize^{3/2}}\Bigg\{\rbra*{\EE \Gradbatch{1}{1}^4}^{3/4} + \Omega^{3/2}+\timescale L  \Omega\sqrt{\rbra*{\Omega\stepsize^2+\frac{\batchsize\stepsize}{\invtemp}}} \Bigg\}.\label{eq:eps_cov_first}
\]
\subsubsection*{Controlling $\error{cov.\Psi\Psi-z_G.2}$}
The analysis of the term $\error{cov.\Psi\Psi-z_G.2}$ in \cref{2b-zz} is more involved.

\textbf{Step 1.}

First we bound the difference between 
\[\frac{1}{\timescale}\sum_{k=0}^{\timescale -1}\frac{\timescale \spacescale^2 \stepsize^2 \Omega}{b} \EE\sum_{m,r=k+1}^{\timescale}  Q(k+1,m)Q(k+1,r)D^2f(\newy)\Bigg[\mathbb{I}_{[\frac{m}{\timescale
},\frac{m+1}{\timescale})}(t),\mathbb{I}_{[\frac{r}{\timescale
},\frac{r+1}{\timescale})}(t)\Bigg]\]
and an intermediate term, given by
\[\frac{1}{\timescale}\sum_{k=0}^{\timescale -1}\frac{\timescale \spacescale^2 \stepsize^2 \Omega}{b} \EE\sum_{m,r=k+1}^{\timescale} (1-h\FI )^{m-k-1}(1-h\FI )^{r-k-1}D^2f(\newy)\Bigg[\mathbb{I}_{[\frac{m}{\timescale
},\frac{m+1}{\timescale})}(t),\mathbb{I}_{[\frac{r}{\timescale
},\frac{r+1}{\timescale})}(t)\Bigg].\]

Since $ab-cd = \frac{1}{2}[(a-c)(b+d)+(b-d)(a+c)]$, notice that:
\[&\error{cov.\Psi\Psi-z_G.2.1}\\
:=&\frac{1}{\timescale}\sum_{k=0}^{\timescale -1}\frac{\timescale \spacescale^2 \stepsize^2 \Omega}{b} \EE\sum_{m,r=k+1}^{\timescale}\rbra*{Q(k+1,m)Q(k+1,r)-(1-h\FI )^{m-k-1}(1-h\FI )^{r-k-1}}\\
&\qquad \times\FD^2 f(\newy) \Bigg[\mathbb{I}_{[\frac{m}{\timescale
},\frac{m+1}{\timescale})}(t),\mathbb{I}_{[\frac{r}{\timescale
},\frac{r+1}{\timescale})}(t)\Bigg]\\
&=\frac{1}{2\timescale}\sum_{k=0}^{\timescale -1}\frac{\timescale \spacescale^2 \stepsize^2 \Omega}{b} \EE\Bigg\{\sum_{m,r=k+1}^{\timescale}\rbra*{Q(k+1,m)-(1-h\FI )^{m-k-1}}\\
&\qquad\qquad\qquad \times\rbra*{Q(k+1,r)+(1-h\FI )^{r-k-1}}\FD^2 f(\newy)  \bigg[\mathbb{I}_{[\frac{m}{\timescale
},\frac{m+1}{\timescale})}(t),\mathbb{I}_{[\frac{r}{\timescale
},\frac{r+1}{\timescale})}(t)\bigg]\Bigg\}\\
&+\frac{1}{2\timescale}\sum_{k=0}^{\timescale -1}\frac{\timescale \spacescale^2 \stepsize^2 \Omega}{b}\EE \Bigg\{\sum_{m,r=k+1}^{\timescale}\rbra*{Q(k+1,m)+(1-h\FI )^{m-k-1}}\\
&\qquad\qquad \times\rbra*{Q(k+1,r)-(1-h\FI )^{r-k-1}}\FD^2 f(\newy)  \bigg[\mathbb{I}_{[\frac{m}{\timescale
},\frac{m+1}{\timescale})}(t),\mathbb{I}_{[\frac{r}{\timescale
},\frac{r+1}{\timescale})}(t)\bigg]\Bigg\}.
\]

And since $\abs{Q(k+1,m)+(1-h\FI )^{r-k-1}}\le 2,$
\[&\error{cov.\Psi\Psi-z_G.2.1}\\
&=\frac{1}{\timescale}\sum_{k=0}^{\timescale -1}\frac{\timescale \spacescale^2 \stepsize^2 \Omega}{b} \EE\Bigg\{\sum_{m,r=k+1}^{\timescale}\rbra*{Q(k+1,m)-(1-h\FI )^{m-k-1}}\\
&\qquad\qquad \times\rbra*{Q(k+1,r)+(1-h\FI )^{r-k-1}}\FD^2 f(\newy)  \bigg[\mathbb{I}_{[\frac{m}{\timescale
},\frac{m+1}{\timescale})}(t),\mathbb{I}_{[\frac{r}{\timescale
},\frac{r+1}{\timescale})}(t)\bigg]\Bigg\}\\
&=\frac{1}{\timescale}\sum_{k=0}^{\timescale -1}\frac{\timescale \spacescale^2 \stepsize^2 \Omega}{b} \EE \FD^2 f(\newy)  \Bigg[\sum_{m=k+1}^{\timescale}\rbra*{Q(k+1,m)-(1-h\FI )^{m-k-1}}\mathbb{I}_{[\frac{m}{\timescale
},\frac{m+1}{\timescale})}(t),\\
&\hphantom{~=\frac{1}{2\timescale}\sum_{k=0}^{\timescale -1}\frac{\timescale \spacescale^2 \stepsize^2 \Omega}{b} \EE \FD^2 f(\newy)  \Bigg[}\sum_{r=k+1}^{\timescale}\rbra*{Q(k+1,r)+(1-h\FI )^{r-k-1}}\mathbb{I}_{[\frac{r}{\timescale
},\frac{r+1}{\timescale})}(t)\Bigg]\\
&\le \frac{1}{\timescale}\sum_{k=0}^{\timescale -1}\frac{\timescale \spacescale^2 \stepsize^2 \Omega}{b} \EE \sup_{m\in\{k+1,\cdots,\timescale-1\}}\abs*{2\rbra*{Q(k+1,m)-(1-h\FI )^{m-k-1}}}\|\FD^2 f(\newy)\|\\
&\le \frac{2}{\timescale}\sum_{k=0}^{\timescale -1}\frac{\timescale \spacescale^2 \stepsize^2 \Omega}{b}\sqrt{\EE\|\FD^2 f(\newy)\|^2} 
\sqrt{\EE \sup_{m\in\{k+1,\cdots,\timescale-1\}}\rbra*{Q(k+1,m)-(1-h\FI )^{m-k-1}}^2}\\
&\stackrel{ \cref{BDforQ}}\lesssim \frac{\timescale \spacescale^2 \stepsize^2 \Omega}{b}\sqrt{\EE\|\FD^2 f(\newy)\|^2} \sqrt{\frac{\timescale h^2 L^2}{b} + \frac{\timescale^2 h^4 L^4}{b}}\\
&\lesssim \frac{\timescale^{3/2} \spacescale^2 \stepsize^3 \Omega L \|g\|_M}{b^{3/2}}\sqrt{1+\timescale\stepsize^2 L^2}\sqrt{\EE\rbra*{1+\frac{1}{3}\|\newy\| +\frac{1}{3}\EE\|\newz\|^2}^2}\\
&\le \frac{ D_2\timescale^{3/2} \spacescale^2 \stepsize^3 \Omega L \|g\|_M}{b^{3/2}}\sqrt{1+\timescale\stepsize^2 L^2},\label{eq:eps_cov_third}
\]
for $D_2$ given in \cref{D1D2}.

\textbf{step 2.}

We now bound the difference between 
\[\label{qt0}\qquad&\frac{1}{\timescale}\sum_{k=0}^{\timescale -1}\frac{\timescale \spacescale^2 \stepsize^2 \Omega}{b} \EE\sum_{m,r=k+1}^{\timescale} (1-h\FI )^{m-k-1}(1-h\FI )^{r-k-1}\Bigg[\mathbb{I}_{[\frac{m}{\timescale
},\frac{m+1}{\timescale})}(t),\mathbb{I}_{[\frac{r}{\timescale
},\frac{r+1}{\timescale})}(t)\Bigg]\\
&=\frac{ \spacescale^2 \stepsize^2 \Omega}{b} \sum_{m,r=1}^{\timescale}\sum_{k=0}^{m\wedge r -1} (1-h\FI )^{m-k-1}(1-h\FI )^{r-k-1}\\
&\hspace{7cm}\times\EE D^2f(\newy)\Bigg[\mathbb{I}_{[\frac{m}{\timescale
},\frac{m+1}{\timescale})}(t),\mathbb{I}_{[\frac{r}{\timescale
},\frac{r+1}{\timescale})}(t)\Bigg]\\
&=\frac{\spacescale^2 \stepsize\Omega}{2b\FI (1-h\FI /2)}\sum_{m,r=1}^{\timescale}\rbra*{(1-h\FI )^{|m-r|}-(1-h\FI )^{m+r}}\\
&\hspace{7cm}\times\EE D^2f(\newy)\Bigg[\mathbb{I}_{[\frac{m}{\timescale
},\frac{m+1}{\timescale})}(t),\mathbb{I}_{[\frac{r}{\timescale
},\frac{r+1}{\timescale})}(t)\Bigg]\\
&=\frac{\spacescale^2 \stepsize\Omega}{2b\FI }\sum_{m,r=1}^{\timescale}\rbra*{(1-h\FI )^{|m-r|}-(1-h\FI )^{m+r}}\EE D^2f(\newy)\Bigg[\mathbb{I}_{[\frac{m}{\timescale
},\frac{m+1}{\timescale})}(t),\mathbb{I}_{[\frac{r}{\timescale
},\frac{r+1}{\timescale})}(t)\Bigg]\\
&+\frac{h\FI }{2}\frac{\spacescale^2 \stepsize\Omega}{2b\FI (1-h\FI /2)}\sum_{m,r=1}^{\timescale}\rbra*{(1-h\FI )^{|m-r|}-(1-h\FI )^{m+r}}\\
&\hspace{7cm}\times\EE D^2f(\newy)\Bigg[\mathbb{I}_{[\frac{m}{\timescale
},\frac{m+1}{\timescale})}(t),\mathbb{I}_{[\frac{r}{\timescale
},\frac{r+1}{\timescale})}(t)\Bigg]
\]
and
\[
\frac{w^2 h\Omega}{2b\FI}\sum_{m,r=1}^{\timescale}\rbra*{e^{-h \FI  |m-r|}-e^{-h \FI  (m+r)}}\EE \FD^2 f(\newy)\sbra*{\mathbb{I}_{[\frac{m}{\alpha},\frac{m+1}{\alpha})},\mathbb{I}_{[\frac{r}{\alpha},\frac{r+1}{\alpha})}}.
\]

Note that the following quantity is $\frac{\stepsize\Sigma}{2}$ times the quantity in \cref{qt0} and $0\le 1-\stepsize \FI \le 1$, so
\[&\error{cov.\Psi\Psi-z_G.2.2}\\
&:=\frac{h\FI }{2}\frac{\spacescale^2 \stepsize\Omega}{2b\FI (1-h\FI /2)}\\
&\qquad\times\sum_{m,r=1}^{\timescale}\rbra*{(1-h\FI )^{|m-r|}-(1-h\FI )^{m+r}}\EE D^2f(\newy)\Bigg[\mathbb{I}_{[\frac{m}{\timescale
},\frac{m+1}{\timescale})}(t),\mathbb{I}_{[\frac{r}{\timescale
},\frac{r+1}{\timescale})}(t)\Bigg]\\
&=\frac{h\FI }{2}\frac{1}{\timescale}\sum_{k=0}^{\timescale -1}\frac{\timescale \spacescale^2 \stepsize^2 \Omega}{b} \\
&\qquad\times\EE\sum_{m,r=k+1}^{\timescale} (1-h\FI )^{m-k-1}(1-h\FI )^{r-k-1}\FD^2 f(\newy)\Bigg[\mathbb{I}_{[\frac{m}{\timescale
},\frac{m+1}{\timescale})}(t),\mathbb{I}_{[\frac{r}{\timescale
},\frac{r+1}{\timescale})}(t)\Bigg]\\
&= \frac{h\FI }{2}\frac{1}{\timescale}\sum_{k=0}^{\timescale -1}\frac{\timescale \spacescale^2 \stepsize^2 \Omega}{b} \\
&\qquad \times\EE \FD^2 f(\newy)\Bigg[\sum_{r=k+1}^{\timescale}(1-h\FI )^{m-k-1}\mathbb{I}_{[\frac{m}{\timescale
},\frac{m+1}{\timescale})}(t),\sum_{r=k+1}^{\timescale}(1-h\FI )^{r-k-1}\mathbb{I}_{[\frac{r}{\timescale
},\frac{r+1}{\timescale})}(t)\Bigg]\\
&\le \frac{\timescale \spacescale^2 \stepsize^3 \Omega \FI }{2b}\EE\|\FD^2 f(\newy)\|\\
&\le \frac{\timescale \spacescale^2 \stepsize^3 \Omega \FI  \|g\|_M}{2b}\EE\rbra*{1+\frac{1}{3}\|\newy\| +\frac{1}{3}\EE\|\newz\|^2}\\
&\le \frac{D_2 \timescale \spacescale^2 \stepsize^3 \Omega \FI  \|g\|_M}{2b},\label{eq:step2_part1}
\]
where $D_2$ is given in \cref{D1D2}.
So the last thing to bound is 
\[\label{eq:last_thing_to_bound}
&\error{cov.\Psi\Psi-z_G.2.3}\\
:=&\Bigg|\frac{\spacescale^2\stepsize\Omega}{2 b \FI }\sum_{m,r=1}^{\timescale}\sbra*{ \rbra*{(1-h\FI )^{|m-r|}-(1-h\FI )^{m+r}}-\rbra*{e^{-h \FI  |m-r|}-e^{-h \FI  (m+r)}}}\\
&\qquad \times \EE D^2f(\newy)\Bigg[\mathbb{I}_{[\frac{m}{\timescale
},\frac{m+1}{\timescale})}(t),\mathbb{I}_{[\frac{r}{\timescale
},\frac{r+1}{\timescale})}(t)\Bigg]\Bigg|.
\]

First from Taylor expansion of $\log (1-\stepsize\FI)$,

\[ e^{-h\FI t}- \left(1-h\FI \right)^{t} &= e^{-h\FI t}\left(1-e^{-\frac{h^2\FI ^2 t}{2}-\frac{h^3\FI ^3t}{3}+\cdots}\right)\\
&= e^{-h\FI t} \frac{h^2\FI ^2 t}{2}+ r(t),\]
where $|r(t)|\le 3h^3\FI ^3 t.$

So we can write the original quantity in \cref{eq:last_thing_to_bound} as

\[
&\error{cov.\Psi\Psi-z_G.2.3}\\
=&\Bigg|\frac{\spacescale^2\stepsize\Omega}{2 b \FI }\sum_{m,r=1}^{\timescale}\left[e^{-h\FI|m-r|}\frac{h^2\FI ^2 |m-r|}{2} - e^{-h\FI(m+r)}\frac{h^2\FI ^2 (m+r)}{2} +R(m,r)\right] \\
&\qquad \times\EE \FD^2 f(\newy)\Bigg[\mathbb{I}_{[\frac{m}{\timescale
},\frac{m+1}{\timescale})}(t),\mathbb{I}_{[\frac{r}{\timescale
},\frac{r+1}{\timescale})}(t)\Bigg]\Bigg|
,\]
where $R(m,r) = r(\abs{m-r})-r(m+r)$, so $\abs{R(m,r)}\le 9\FI ^3\timescale\stepsize^3$.

To bound this term, first, note that
\[ &\error{cov.\Psi\Psi-z_G.2.3.1}\\
&:=\Bigg|\frac{\spacescale^2\stepsize\Omega}{2 b \FI }\sum_{m,r=1}^{\timescale}R(m,r) 
\EE \FD^2 f(\newy)\Bigg[\mathbb{I}_{[\frac{m}{\timescale
},\frac{m+1}{\timescale})}(t),\mathbb{I}_{[\frac{r}{\timescale
},\frac{r+1}{\timescale})}(t)\Bigg]\Bigg|\\
&=\Bigg|\frac{\spacescale^2\stepsize\Omega}{2 b \FI }\sum_{m=1}^{\timescale}
\EE \FD^2 f(\newy)\Bigg[\mathbb{I}_{[\frac{m}{\timescale
},\frac{m+1}{\timescale})}(t),\sum_{r=1}^{\timescale}R(m,r) \mathbb{I}_{[\frac{r}{\timescale
},\frac{r+1}{\timescale})}(t)\Bigg]\Bigg|\\
&\le \frac{\spacescale^2\stepsize\Omega \EE \|\FD^2 f(\newy)\|}{2 b \FI } \sum_{m=1}^{\timescale}\sup_{r \in \{1,\cdots,\timescale\}}\left|R(m,r)\right|\\
&\lesssim  \frac{\timescale^2\spacescale^2\stepsize^4\Omega \FI ^2\EE \|\FD^2 f(\newy)\|}{2 b}.\label{eq:cov_psi_psi2.3.1}
\]

Second,
\[
&\error{cov.\Psi\Psi-z_G.2.3.2}\\
&:=\Bigg|\frac{\spacescale^2\stepsize\Omega}{2 b \FI }\sum_{m,r=1}^{\timescale}e^{-h\FI(m+r)}\frac{h^2\FI ^2 (m+r)}{2} 
\EE \FD^2 f(\newy)\Bigg[\mathbb{I}_{[\frac{m}{\timescale
},\frac{m+1}{\timescale})}(t),\mathbb{I}_{[\frac{r}{\timescale
},\frac{r+1}{\timescale})}(t)\Bigg]\Bigg|\\
\le & \Bigg|\frac{\spacescale^2\stepsize\Omega}{2 b \FI }\sum_{m,r=1}^{\timescale}e^{-h\FI(m+r)}\frac{h^2\FI ^2 m}{2}  
\EE \FD^2 f(\newy)\Bigg[\mathbb{I}_{[\frac{m}{\timescale
},\frac{m+1}{\timescale})}(t),\mathbb{I}_{[\frac{r}{\timescale
},\frac{r+1}{\timescale})}(t)\Bigg]\Bigg|\\
&+ \Bigg|\frac{\spacescale^2\stepsize\Omega}{2 b \FI }\sum_{m,r=1}^{\timescale}e^{-h\FI(m+r)}\frac{h^2\FI ^2 r}{2}  
\EE \FD^2 f(\newy)\Bigg[\mathbb{I}_{[\frac{m}{\timescale
},\frac{m+1}{\timescale})}(t),\mathbb{I}_{[\frac{r}{\timescale
},\frac{r+1}{\timescale})}(t)\Bigg]\Bigg|\\
= & \Bigg|\frac{\spacescale^2\stepsize^2\Omega \FI }{4 b}  
\EE \FD^2 f(\newy)\left[\sum_{m=1}^{\timescale} hm e^{-\FI m} \mathbb{I}_{[\frac{m}{\timescale
},\frac{m+1}{\timescale})}(t),\sum_{r=1}^{\timescale} e^{-\FI hr}\mathbb{I}_{[\frac{r}{\timescale
},\frac{r+1}{\timescale})}(t)\right]\Bigg|\\
& +\Bigg|\frac{\spacescale^2\stepsize^2\Omega \FI }{4 b}   
\EE \FD^2 f(\newy)\left[\sum_{m=1}^{\timescale} e^{-\FI hm}\mathbb{I}_{[\frac{m}{\timescale
},\frac{m+1}{\timescale})}(t),\sum_{r=1}^{\timescale} hr e^{-\FI r} \mathbb{I}_{[\frac{r}{\timescale
},\frac{r+1}{\timescale})}(t)\right]\Bigg|\\
\le & \frac{\spacescale^2\stepsize^3 \timescale\Omega \FI }{ 2 b}   
\EE \|\FD^2 f(\newy)\|..\label{eq:cov_psi_psi2.3.2}
\]

Third, using $|m-r| = m+r -2( m\wedge r)$, we have 
\[
&\error{cov.\Psi\Psi-z_G.2.3.3}\\
&:=\frac{\spacescale^2\stepsize\Omega}{2 b \FI }\sum_{m,r=1}^{\timescale}e^{-h\FI|m-r|}\frac{h^2 \FI ^2 |m-r|}{2}  
\EE \FD^2 f(\newy)\Bigg[\mathbb{I}_{[\frac{m}{\timescale
},\frac{m+1}{\timescale})}(t),\mathbb{I}_{[\frac{r}{\timescale
},\frac{r+1}{\timescale})}(t)\Bigg]\\
&=\frac{\spacescale^2\stepsize\Omega}{2 b \FI }\sum_{m,r=1}^{\timescale}e^{2h\FI(m\wedge r)}\frac{h^2\FI ^2 |m-r|}{2}  
\EE \FD^2 f(\newy)\Bigg[e^{-h\FI m}\mathbb{I}_{[\frac{m}{\timescale
},\frac{m+1}{\timescale})}(t),e^{-h\FI  r}\mathbb{I}_{[\frac{r}{\timescale
},\frac{r+1}{\timescale})}(t)\Bigg]\\
&=\frac{\spacescale^2\stepsize\Omega}{2 b \FI }\sum_{m,r=1}^{\timescale}e^{2h\FI(m\wedge r)}\frac{h^2\FI ^2 (m+r)}{2}  
\EE \FD^2 f(\newy)\Bigg[e^{-h\FI m}\mathbb{I}_{[\frac{m}{\timescale
},\frac{m+1}{\timescale})}(t),e^{-h\FI  r}\mathbb{I}_{[\frac{r}{\timescale
},\frac{r+1}{\timescale})}(t)\Bigg]\\
&-\frac{\spacescale^2\stepsize\Omega}{b \FI }\sum_{m,r=1}^{\timescale}e^{2h\FI(m\wedge r)}\frac{h^2\FI ^2 (m\wedge r)}{2}  
\EE \FD^2 f(\newy)\Bigg[e^{-h\FI m}\mathbb{I}_{[\frac{m}{\timescale
},\frac{m+1}{\timescale})}(t),e^{-h\FI  r}\mathbb{I}_{[\frac{r}{\timescale
},\frac{r+1}{\timescale})}(t)\Bigg]\\
&=\frac{\spacescale^2\stepsize^2\Omega \FI }{4 b }\sum_{m,r=1}^{\timescale}e^{2h\FI(m\wedge r)}  
\EE \FD^2 f(\newy)\Bigg[hme^{-h\FI m}\mathbb{I}_{[\frac{m}{\timescale
},\frac{m+1}{\timescale})}(t),e^{-h\FI  r}\mathbb{I}_{[\frac{r}{\timescale
},\frac{r+1}{\timescale})}(t)\Bigg]\\
&+\frac{\spacescale^2\stepsize^2\Omega \FI }{4 b }\sum_{m,r=1}^{\timescale}e^{2h\FI(m\wedge r)}  
\EE \FD^2 f(\newy)\Bigg[e^{-h\FI m}\mathbb{I}_{[\frac{m}{\timescale
},\frac{m+1}{\timescale})}(t),hre^{-h\FI  r}\mathbb{I}_{[\frac{r}{\timescale
},\frac{r+1}{\timescale})}(t)\Bigg]\\
&-\frac{\spacescale^2\stepsize^3\Omega \FI }{2b }\sum_{m,r=1}^{\timescale}e^{2h\FI(m\wedge r)} (m\wedge r)  
\EE \FD^2 f(\newy)\Bigg[e^{-h\FI m}\mathbb{I}_{[\frac{m}{\timescale
},\frac{m+1}{\timescale})}(t),e^{-h\FI  r}\mathbb{I}_{[\frac{r}{\timescale
},\frac{r+1}{\timescale})}(t)\Bigg].\label{eq:cov_psi_psi2.3.3}
\]

All the three lines have similar structure.
For the first line, since $\left|e^{2h\FI k}-e^{2h\FI(k-1)} \right| \le 2h\FI  e^{2\FI hk}$ and $e^{-h\FI  m}e^{-h\FI r} \le e^{-2h\FI k}$ for $m,r \ge k,$
\[
&\Bigg|\frac{\spacescale^2\stepsize^2\Omega \FI }{4 b }\sum_{m,r=1}^{\timescale}e^{2h\FI(m\wedge r)}  
\EE \FD^2 f(\newy)\Bigg[hme^{-h\FI m}\mathbb{I}_{[\frac{m}{\timescale
},\frac{m+1}{\timescale})}(t),e^{-h\FI  r}\mathbb{I}_{[\frac{r}{\timescale
},\frac{r+1}{\timescale})}(t)\Bigg]\Bigg|\\
\le &\Bigg|\frac{\spacescale^2\stepsize^2\Omega \FI }{4 b }\sum_{m,r=1}^{\timescale}e^{2h\FI}  
\EE \FD^2 f(\newy)\Bigg[hme^{-h\FI m}\mathbb{I}_{[\frac{m}{\timescale
},\frac{m+1}{\timescale})}(t),e^{-h\FI  r}\mathbb{I}_{[\frac{r}{\timescale
},\frac{r+1}{\timescale})}(t)\Bigg]\Bigg|\\
&+ \Bigg|\frac{\spacescale^2\stepsize^2\Omega \FI }{4 b }\sum_{k=2}^{\timescale}\sum_{m,r=k}^{\timescale}\left(e^{2h\FI k}-e^{2h\FI(k-1)}\right)  
\\
&\hspace{3cm}\times\EE \FD^2 f(\newy)\Bigg[hme^{-h\FI m}\mathbb{I}_{[\frac{m}{\timescale
},\frac{m+1}{\timescale})}(t),e^{-h\FI  r}\mathbb{I}_{[\frac{r}{\timescale
},\frac{r+1}{\timescale})}(t)\Bigg]\Bigg|\\
= &\Bigg|\frac{\spacescale^2\stepsize^2\Omega \FI }{4 b }e^{2h\FI} 
\EE \FD^2 f(\newy)\Bigg[\sum_{m=1}^{\timescale}hme^{-h\FI m}\mathbb{I}_{[\frac{m}{\timescale
},\frac{m+1}{\timescale})}(t),\sum_{r=1}^{\timescale}e^{-h\FI  r}\mathbb{I}_{[\frac{r}{\timescale
},\frac{r+1}{\timescale})}(t)\Bigg]\Bigg|\\
&+ \Bigg|\frac{\spacescale^2\stepsize^2\Omega \FI }{4 b }\sum_{k=2}^{\timescale}\left(e^{2h\FI k}-e^{2h\FI(k-1)}\right)  
\\
&\hspace{3cm}\times\EE \FD^2 f(\newy)\Bigg[\sum_{m=k}^{\timescale}hme^{-h\FI m}\mathbb{I}_{[\frac{m}{\timescale
},\frac{m+1}{\timescale})}(t),\sum_{r=k}^{\timescale}e^{-h\FI  r}\mathbb{I}_{[\frac{r}{\timescale
},\frac{r+1}{\timescale})}(t)\Bigg]\Bigg|\\
\le &\frac{\spacescale^2\timescale\stepsize^3\Omega \FI }{4 b } 
\EE \|\FD^2 f(\newy)\| \rbra*{1+2h\timescale \FI }.
\]

Similarly, since $\left|ke^{2h\FI k}-(k-1)e^{2h\FI(k-1)} \right| \le \left|ke^{2h\FI k}-ke^{2h\FI(k-1)} \right| + \left|e^{2h\FI(k-1)}\right| \le (2h\timescale \FI +1) e^{2\FI hk}$ and $e^{-h\FI  m}e^{-h\FI r} \le e^{-2h\FI k}$ for $m,r \ge k,$
\[
&\frac{\spacescale^2\stepsize^3\Omega \FI }{2b }\sum_{m,r=1}^{\timescale}e^{2h\FI(m\wedge r)} (m\wedge r)  
\EE \FD^2 f(\newy)\Bigg[e^{-h\FI m}\mathbb{I}_{[\frac{m}{\timescale
},\frac{m+1}{\timescale})}(t),e^{-h\FI  r}\mathbb{I}_{[\frac{r}{\timescale
},\frac{r+1}{\timescale})}(t)\Bigg]\\
\le & \frac{\spacescale^2\stepsize^3\Omega \FI }{2b } 
\EE \|\FD^2 f(\newy)\| \rbra*{\timescale+\timescale(1+2h\timescale \FI )}.
\]

So, for $\error{cov.\Psi\Psi-z_G.2.3.3}$ given in \cref{eq:cov_psi_psi2.3.3},
\[&\error{cov.\Psi\Psi-z_G.2.3.3}
\le \frac{\spacescale^2\stepsize^3\Omega \FI }{2 b } 
\EE \|\FD^2 f(\newy)\| \rbra*{1+2\timescale(1+2h\timescale \FI )}.\label{eq:cov_psi_psi2.3.3_bound}
\]

Thus, it follows from \cref{eq:cov_psi_psi2.3.1,eq:cov_psi_psi2.3.2,eq:cov_psi_psi2.3.3,eq:cov_psi_psi2.3.3_bound} that
\[&\error{cov.\Psi\Psi-z_G.2.3}\\
\lesssim& \error{cov.\Psi\Psi-z_G.2.3.1}+\error{cov.\Psi\Psi-z_G.2.3.2}+\error{cov.\Psi\Psi-z_G.2.3.3}\\
\lesssim& \frac{  \timescale^2\spacescale^2\stepsize^4\Omega \FI ^2\EE \|\FD^2 f(\newy)\|}{ b}  + \frac{\spacescale^2\stepsize^3 \timescale\Omega \FI }{  b}   
\EE \|\FD^2 f(\newy)\|\\
&+\frac{\spacescale^2\stepsize^3\Omega \FI }{ b } 
\EE \|\FD^2 f(\newy)\| \rbra*{1+2\timescale(1+2h\timescale \FI )}\\
\lesssim&\frac{\stepsize^3 \spacescale^2 \Omega \FI  \EE \|\FD^2 f(\newy)\|}{b}\rbra*{\timescale^2\stepsize \FI +\timescale+\timescale(1+2h\timescale \FI )+1}\\
\lesssim& \frac{\stepsize^3 \spacescale^2 \Omega \FI  \|g\|_M}{2b}\rbra*{\timescale^2\stepsize \FI +\timescale+1}\EE\rbra*{1+\frac{1}{3}\|\newy\| +\frac{1}{3}\EE\|\newz\|^2}\\
\lesssim& \frac{ D_2 \stepsize^3 \spacescale^2 \Omega \FI  \|g\|_M}{2b}\rbra*{\timescale^2\stepsize \FI +\timescale},\label{eq:eps_cov_final}
\]
where $D_2$ is given in \cref{D1D2}.

Adding the terms \cref{eq:eps_cov_first,eq:eps_cov_third,eq:step2_part1,eq:eps_cov_final}, we have
\[\label{eq:cov3_bound}&\error{cov.\Psi\Psi -Z_G}\\
\le &\error{cov.\Psi\Psi -Z_G.1}+\error{cov.\Psi\Psi -Z_G.2.1}+\error{cov.\Psi\Psi -Z_G.2.2}+\error{cov.\Psi\Psi -Z_G.2.3}\\
\lesssim& \frac{\timescale \spacescale^3 \stepsize^{3}D_2\|g\|_M}{\batchsize^{3/2}}\Bigg\{\rbra*{\EE \Gradbatch{1}{1}^4}^{3/4} + \Omega^{3/2}+\timescale L  \Omega\sqrt{\rbra*{\Omega\stepsize^2+\frac{\batchsize\stepsize}{\invtemp}}} \Bigg\}\\
&+ \frac{ D_2\timescale^{3/2} \spacescale^2 \stepsize^3 \Omega L \|g\|_M}{b^{3/2}}\sqrt{1+\timescale\stepsize^2 L^2}+\frac{D_2 \timescale \spacescale^2 \stepsize^3 \Omega \FI  \|g\|_M}{2b}\\
&+\frac{ D_2 \stepsize^3 \spacescale^2 \Omega \FI  \|g\|_M}{2b}\rbra*{\timescale^2\stepsize \FI +\timescale}\\
\lesssim&\frac{\timescale \spacescale^3 \stepsize^{3}D_2\|g\|_M}{\batchsize^{3/2}}\Bigg\{\rbra*{\EE \Gradbatch{1}{1}^4}^{3/4} + \Omega^{3/2}+\timescale L  \Omega\sqrt{\rbra*{\Omega\stepsize^2+\frac{\batchsize\stepsize}{\invtemp}}} \Bigg\}\\
&+ \frac{ D_2\timescale^{3/2} \spacescale^2 \stepsize^3 \Omega L \|g\|_M}{b^{3/2}}\sqrt{1+\timescale\stepsize^2 L^2}+\frac{D_2 \timescale \spacescale^2 \stepsize^3 \Omega \FI  \|g\|_M}{2b}\rbra*{\timescale\stepsize \FI +1}.
\]

\subsubsection{Bounding $\error{cov.\Sigma\noise}$} \label{app:bounding_err_cov_sig_noise}
First, note that
\[
&\frac{1}{\timescale}\sum_{k=0}^{\timescale -1}\frac{\sqrt{2}\timescale \spacescale^2 \stepsize^{3/2}}{b\invtemp^{1/2}}\EE^{\newy} \FD^2 f(\newy)\Bigg[\sum_{m=k+1}^{\timescale}Q(k+1,m)\rbra*{\noise_{k}-\xi'_{k}}\mathbb{I}_{[\frac{m}{\timescale
},\frac{m+1}{\timescale})}(t),\\
&\hspace{3cm}\sum_{r=k+1}^{\timescale}Q(k+1,r)\rbra*{-\sum_{i=1}^{b}\Heisbatch{k}{i} +\sum_{i=1}^{b}\swapHeisbatch{k}{i} }\newpmt_k\mathbb{I}_{[\frac{r}{\timescale
},\frac{r+1}{\timescale})}(t)\Bigg]\\
&=\frac{1}{\timescale}\sum_{k=0}^{\timescale -1}\frac{\sqrt{2}\timescale \spacescale^2 \stepsize^{3/2}}{b\invtemp^{1/2}}\EE^{\newy} \rbra*{\noise_{k}-\xi'_{k}}\rbra*{-\sum_{i=1}^{b}\Heisbatch{k}{i} +\sum_{i=1}^{b}\swapHeisbatch{k}{i} }\newpmt_k\\
&\qquad\times  \FD^2 f(\newy)\Bigg[\sum_{m=k+1}^{\timescale}Q(k+1,m)\mathbb{I}_{[\frac{m}{\timescale
},\frac{m+1}{\timescale})}(t),\sum_{r=k+1}^{\timescale}Q(k+1,r)\mathbb{I}_{[\frac{r}{\timescale
},\frac{r+1}{\timescale})}(t)\Bigg]\\
&=\frac{1}{\timescale}\sum_{k=0}^{\timescale -1}\frac{\sqrt{2}\timescale \spacescale^2 \stepsize^{3/2}}{b\invtemp^{1/2}} \noise_k\rbra*{-\sum_{i=1}^{b}\Heisbatch{k}{i} +b\FI }\newpmt_k \\
&\qquad\times \FD^2 f(\newy)\Bigg[\sum_{m=k+1}^{\timescale}Q(k+1,m)\mathbb{I}_{[\frac{m}{\timescale
},\frac{m+1}{\timescale})}(t),\sum_{r=k+1}^{\timescale}Q(k+1,r)\mathbb{I}_{[\frac{r}{\timescale
},\frac{r+1}{\timescale})}(t)\Bigg]
.\]

Since $Q(i,j)\le 1,\forall i,j$, using \cref{BDforM},
\[\label{eq:cov4_bound}
\error{cov.\Sigma\noise}&=\Bigg|\EE\frac{1}{\timescale}\sum_{k=0}^{\timescale -1}\frac{\sqrt{2}\timescale \spacescale^2 \stepsize^{3/2}}{b\invtemp^{1/2}}\noise_k\rbra*{-\sum_{i=1}^{b}\Heisbatch{k}{i} +b\FI }\newpmt_k \\
&\qquad \times\FD^2 f(\newy)\Bigg[\sum_{m=k+1}^{\timescale}Q(k+1,m)\mathbb{I}_{[\frac{m}{\timescale
},\frac{m+1}{\timescale})}(t),\sum_{r=k+1}^{\timescale}Q(k+1,r)\mathbb{I}_{[\frac{r}{\timescale
},\frac{r+1}{\timescale})}(t)\Bigg]\Bigg|\\
&\le \frac{1}{\timescale}\sum_{k=0}^{\timescale -1}\frac{\sqrt{2}\timescale \spacescale^2 \stepsize^{3/2}}{b\invtemp^{1/2}}\EE\Bigg|\noise_k\rbra*{-\sum_{i=1}^{b}\Heisbatch{k}{i} +b\FI }\newpmt_k\Bigg|\|\FD^2 f(\newy)\|\\
&\le \frac{1}{\timescale}\sum_{k=0}^{\timescale -1}\frac{\sqrt{2}\timescale \spacescale^2 \stepsize^{3/2}}{b\invtemp^{1/2}}\sqrt{\EE\|\FD^2 f(\newy)\|^2}\sqrt{\EE \noise_k^2\rbra*{\sum_{i=1}^{b}\Heisbatch{k}{i} -b\FI }^2\newpmt_k^2}\\
&=\sum_{k=0}^{\timescale-1}\frac{\sqrt{2}\spacescale^2 \stepsize^{3/2}}{b\invtemp^{1/2}}\sqrt{\EE\|\FD^2 f(\newy)\|^2}\sqrt{\EE \rbra*{\sum_{i=1}^{b}\Heisbatch{k}{i} -b\FI }^2\EE\newpmt_k^2}\\
&\le \frac{\sqrt{2}\timescale^{3/2} \spacescale^2 \stepsize^{3/2}L \|g\|_M}{b^{1/2}\invtemp^{1/2}}\sqrt{\rbra*{\frac{h^2 \Omega}{b} + \frac{2h}{\invtemp}}}\sqrt{\EE\rbra*{1+\frac{1}{3}\|\newy\| +\frac{1}{3}\EE\|\newz\|^2}^2}\\
&\le \frac{D_2\sqrt{2}\timescale^{3/2} \spacescale^2 \stepsize^{3/2}L \|g\|_M}{b^{1/2}\invtemp^{1/2}}\sqrt{\rbra*{\frac{h^2 \Omega}{b} + \frac{2h}{\invtemp}}},
\]
where $D_2$ is given in \cref{D1D2}.

\subsubsection{Bounding $\error{cov.\Psi\noise}$} \label{bounding_error_cov_psi_noise}

First, note that
\[
&\frac{1}{\timescale}\sum_{k=0}^{\timescale -1}\frac{\sqrt{2}\timescale \spacescale^2 \stepsize^{3/2}}{b\invtemp^{1/2}}\EE^{\newy} \FD^2 f(\newy)\Bigg[\sum_{m=k+1}^{\timescale}Q(k+1,m)\rbra*{\noise_{k}-\xi'_{k}}\mathbb{I}_{[\frac{m}{\timescale
},\frac{m+1}{\timescale})}(t),\\
&\hphantom{~\frac{1}{\timescale}\sum_{k=0}^{\timescale -1}\frac{\sqrt{2}\timescale \spacescale^2 \stepsize^{3/2}}{b\invtemp^{1/2}}\EE^{\newy} \FD^2 f(\newy)\Bigg[}\sum_{r=k+1}^{\timescale}Q(k+1,r)\rbra*{\sum_{i=1}^{b}\Gradbatch{k}{i} -\sum_{i=1}^{b} \swapGradbatch{k}{i}}\mathbb{I}_{[\frac{r}{\timescale
},\frac{r+1}{\timescale})}(t)\Bigg]\\
&=\frac{1}{\timescale}\sum_{k=0}^{\timescale -1}\frac{\sqrt{2}\timescale \spacescale^2 \stepsize^{3/2}}{b\invtemp^{1/2}}\EE^{\newy} \rbra*{\noise_{k}-\xi'_{k}}\rbra*{\sum_{i=1}^{b}\Gradbatch{k}{i} -\sum_{i=1}^{b} \swapGradbatch{k}{i}} \\
&\qquad\times \FD^2 f(\newy)\Bigg[\sum_{m=k+1}^{\timescale}Q(k+1,m)\mathbb{I}_{[\frac{m}{\timescale
},\frac{m+1}{\timescale})}(t),\sum_{r=k+1}^{\timescale}Q(k+1,r)\mathbb{I}_{[\frac{r}{\timescale
},\frac{r+1}{\timescale})}(t)\Bigg]\\
&=\frac{1}{\timescale}\sum_{k=0}^{\timescale -1}\frac{\sqrt{2}\timescale \spacescale^2 \stepsize^{3/2}}{b\invtemp^{1/2}} \noise_k\\
&\quad\times\sum_{i=1}^{b}\Gradbatch{k}{i} \FD^2 f(\newy)\Bigg[\sum_{m=k+1}^{\timescale}Q(k+1,m)\mathbb{I}_{[\frac{m}{\timescale
},\frac{m+1}{\timescale})}(t),\sum_{r=k+1}^{\timescale}Q(k+1,r)\mathbb{I}_{[\frac{r}{\timescale
},\frac{r+1}{\timescale})}(t)\Bigg]
.\]

Here we define 
\[
\newyt^{k} \defas \newyt&-\spacescale\sum_{m=k+1}^{\timescale}Q(k+1,m)\rbra*{\frac{\stepsize}{b}\sum_{i=1}^{b}\Gradbatch{k}{i}+\sqrt{\frac{2\stepsize}{\invtemp}}\noise_{k}}\mathbb{I}_{[\frac{m}{\timescale
},\frac{m+1}{\timescale})}(t)\\
&+ \spacescale \frac{\frac{h}{b}\sum_{i=1}^{b}\Heisbatch{k}{i}}{1-\frac{h}{b}\sum_{i=1}^{b}\Heisbatch{k}{i}}\sum_{j=0}^{k-1}\sum_{m=k+1}^{\timescale}Q(j+1,m)\rbra*{\frac{h}{b}\sum_{i=1}^{b}\Gradbatch{j}{i}+\sqrt{\frac{2h}{\invtemp}}\noise_{j}}\ind{\cointer{\frac{m}{\samplesize},\frac{m+1}{\samplesize}}}
\label{newyk}\]
Then $\newyt^{k}$ is independent with $\noise_k$ and $\Gradbatch{k}{i},\forall i=1,\cdots,b$.

Thus, using $f$'s properties,

\[\label{eq:cov5_bound}
\error{cov.\Psi\noise}&=\Bigg|\EE\frac{1}{\timescale}\sum_{k=0}^{\timescale -1}\frac{\sqrt{2}\timescale \spacescale^2 \stepsize^{3/2}}{b\invtemp^{1/2}}\noise_k\sum_{i=1}^{b}\Gradbatch{k}{i} \\
&\quad \times \FD^2 f(\newy)\Bigg[\sum_{m=k+1}^{\timescale}Q(k+1,m)\mathbb{I}_{[\frac{m}{\timescale
},\frac{m+1}{\timescale})}(t),\sum_{r=k+1}^{\timescale}Q(k+1,r)\mathbb{I}_{[\frac{r}{\timescale
},\frac{r+1}{\timescale})}(t)\Bigg]\Bigg|\\
&=\Bigg|\EE\frac{1}{\timescale}\sum_{k=0}^{\timescale -1}\frac{\sqrt{2}\timescale \spacescale^2 \stepsize^{3/2}}{b\invtemp^{1/2}}\noise_k\sum_{i=1}^{b}\Gradbatch{k}{i} \\
&\quad \times\rbra*{\FD^2 f(\newy)-\FD^2 f(\newy^{k})}\Bigg[\sum_{m=k+1}^{\timescale}Q(k+1,m)\mathbb{I}_{[\frac{m}{\timescale
},\frac{m+1}{\timescale})}(t),\\
&\hspace{5cm}\sum_{r=k+1}^{\timescale}Q(k+1,r)\mathbb{I}_{[\frac{r}{\timescale
},\frac{r+1}{\timescale})}(t)\Bigg]\Bigg|\\
&\le \frac{1}{\timescale}\sum_{k=0}^{\timescale -1}\frac{\sqrt{2}\timescale \spacescale^2 \stepsize^{3/2}}{b\invtemp^{1/2}}\EE\Bigg|\noise_k\sum_{i=1}^{b}\Gradbatch{k}{i}\Bigg|\|\FD^2 f(\newy)-\FD^2 f(\newy^{k})\|\\
&\le\frac{1}{\timescale}\sum_{k=0}^{\timescale -1}\frac{\sqrt{2}\timescale \spacescale^3 \stepsize^{3/2}}{b\invtemp^{1/2}}\EE\Bigg\{\Bigg|\noise_k\sum_{i=1}^{b}\Gradbatch{k}{i}\Bigg|\rbra*{\frac{1}{2}+\frac{1}{3}\|\newy^k\|+\frac{1}{3}\EE\|\hat{Z}\|^2}\\
&\times\rbra*{\abs*{\frac{\stepsize}{b}\sum_{i=1}^{b}\Gradbatch{k}{i}+\sqrt{\frac{2\stepsize}{\invtemp}}\noise_{k}}+\frac{\stepsize L}{1-\stepsize L}\sum_{j=0}^{k-1}\abs*{\frac{h}{b}\sum_{i=1}^{b}\Gradbatch{j}{i}+\sqrt{\frac{2h}{\invtemp}}\noise_{j}}}\Bigg\}\\
&\le\frac{\|g\|_M}{\timescale}\sum_{k=0}^{\timescale -1}\frac{\sqrt{2}\timescale \spacescale^3 \stepsize^{3/2}}{b\invtemp^{1/2}}\sqrt{\EE\rbra*{\frac{1}{2}+\frac{1}{3}\|\newy^k\|+\frac{1}{3}\EE\|\hat{Z}\|^2}^2}\\
&\times\Bigg\{\EE \abs*{\frac{\stepsize}{b}\noise_k\rbra*{\sum_{i=1}^{b}\Gradbatch{k}{i}}^2+\sqrt{\frac{2\stepsize}{\invtemp}}\noise_{k}^2\sum_{i=1}^{b}\Gradbatch{k}{i}}\\
&+\frac{\timescale\stepsize L}{1-\stepsize L}\EE\abs*{\noise_k\sum_{i=1}^{b}\Gradbatch{k}{i}}\sqrt{\EE\rbra*{\frac{h}{b}\sum_{i=1}^{b}\Gradbatch{1}{i}+\sqrt{\frac{2h}{\invtemp}}\noise_{1}}^2}\Bigg\}\\
&\lesssim \frac{\timescale \spacescale^3 \stepsize^{3/2}D_2\|g\|_M}{b\invtemp^{1/2}}\Bigg\{\stepsize\Omega +\sqrt{\frac{\stepsize \batchsize \Omega}{\invtemp}}+\frac{\timescale\stepsize L}{1-\stepsize L} \sqrt{\batchsize \Omega\rbra*{\frac{\Omega\stepsize^2}{\batchsize}+\frac{\stepsize}{\batchsize}}} \Bigg\}\\
&\lesssim \frac{\timescale \spacescale^3 \stepsize^{3/2}\sqrt{\Omega}D_2\|g\|_M}{\batchsize^{1/2}\invtemp^{1/2}}\rbra*{\frac{\stepsize\sqrt{\Omega}}{\batchsize^{1/2}} +\sqrt{\frac{\stepsize}{\invtemp}}}\rbra*{1+\timescale\stepsize L}
.
\]

\subsubsection{Bounding $\error{cov.\noise\noise-z_\noise}$}\label{bounding_error_cov_noise_noisez_noise}

First,
\[
&\frac{1}{2\timescale}\sum_{k=0}^{\timescale -1}\frac{2\timescale \spacescale^2 \stepsize}{\invtemp}\EE^{\newy} \FD^2 f(\newy)\Bigg[\sum_{m=k+1}^{\timescale}Q(k+1,m)\rbra*{\noise_{k}-\xi'_{k}}\mathbb{I}_{[\frac{m}{\timescale
},\frac{m+1}{\timescale})}(t),\\
&\hspace{5cm}\sum_{r=k+1}^{\timescale}Q(k+1,r)\rbra*{\noise_{k}-\xi'_{k}}\mathbb{I}_{[\frac{r}{\timescale
},\frac{r+1}{\timescale})}(t)\Bigg]\\
&=\frac{\spacescale^2 \stepsize}{\invtemp}\sum_{k=0}^{\timescale -1}\EE^{\newy} \rbra*{\noise_{k}-\xi'_{k}}^2 \FD^2 f(\newy)\Bigg[\sum_{m=k+1}^{\timescale}Q(k+1,m)\mathbb{I}_{[\frac{m}{\timescale
},\frac{m+1}{\timescale})}(t),\\
&\hspace{5cm}\sum_{r=k+1}^{\timescale}Q(k+1,r)\mathbb{I}_{[\frac{r}{\timescale
},\frac{r+1}{\timescale})}(t)\Bigg]\\
&=\frac{\spacescale^2 \stepsize}{\invtemp}\sum_{k=0}^{\timescale -1}\rbra*{\noise_{k}^2+1}\FD^2 f(\newy) \Bigg[\sum_{m=k+1}^{\timescale}Q(k+1,m)\mathbb{I}_{[\frac{m}{\timescale
},\frac{m+1}{\timescale})}(t),\sum_{r=k+1}^{\timescale}Q(k+1,r)\mathbb{I}_{[\frac{r}{\timescale
},\frac{r+1}{\timescale})}(t)\Bigg]\\
&=\frac{2\spacescale^2 \stepsize}{\invtemp}\sum_{k=0}^{\timescale -1}\FD^2 f(\newy) \Bigg[\sum_{m=k+1}^{\timescale}Q(k+1,m)\mathbb{I}_{[\frac{m}{\timescale
},\frac{m+1}{\timescale})}(t),\sum_{r=k+1}^{\timescale}Q(k+1,r)\mathbb{I}_{[\frac{r}{\timescale
},\frac{r+1}{\timescale})}(t)\Bigg]\\
&+\frac{\spacescale^2 \stepsize}{\invtemp}\sum_{k=0}^{\timescale -1}\rbra*{\noise_{k}^2-1}\FD^2 f(\newy) \Bigg[\sum_{m=k+1}^{\timescale}Q(k+1,m)\mathbb{I}_{[\frac{m}{\timescale
},\frac{m+1}{\timescale})}(t),\sum_{r=k+1}^{\timescale}Q(k+1,r)\mathbb{I}_{[\frac{r}{\timescale
},\frac{r+1}{\timescale})}(t)\Bigg].
\]
For the second line, define 
\[\label{newyknoise}
\newyt^{k\noise} \defas \newyt-\spacescale\sum_{m=k+1}^{\timescale}Q(k+1,m)\sqrt{\frac{2\stepsize}{\invtemp}}\noise_{k}\mathbb{I}_{[\frac{m}{\timescale
},\frac{m+1}{\timescale})}(t).\]
Then $\newyt^{k\noise}$ is independent with  $\noise_{k}$. Thus using $f$'s properties,

\[
&\Bigg|\EE\frac{\spacescale^2 \stepsize}{\invtemp}\sum_{k=0}^{\timescale -1}\rbra*{\noise_{k}^2-1}\FD^2 f(\newy) \Bigg[\sum_{m=k+1}^{\timescale}Q(k+1,m)\mathbb{I}_{[\frac{m}{\timescale
},\frac{m+1}{\timescale})}(t),\sum_{r=k+1}^{\timescale}Q(k+1,r)\mathbb{I}_{[\frac{r}{\timescale
},\frac{r+1}{\timescale})}(t)\Bigg]\Bigg|\\
&=\Bigg|\EE\frac{\spacescale^2 \stepsize}{\invtemp}\sum_{k=0}^{\timescale -1}\rbra*{\noise_{k}^2-1}\rbra*{\FD^2 f(\newy)-\FD^2 f(\newy^{k\noise})}\Bigg[\sum_{m=k+1}^{\timescale}Q(k+1,m)\mathbb{I}_{[\frac{m}{\timescale
},\frac{m+1}{\timescale})}(t),\\
&\hspace{8cm}\sum_{r=k+1}^{\timescale}Q(k+1,r)\mathbb{I}_{[\frac{r}{\timescale
},\frac{r+1}{\timescale})}(t)\Bigg]\Bigg|\\
&\le \frac{\spacescale^2 \stepsize}{\invtemp}\sum_{k=0}^{\timescale -1}\EE\abs*{\noise_{k}^2-1}\|\FD^2 f(\newy)-\FD^2 f(\newy^{k\noise})\|\\
&\le \frac{ \spacescale^2 \stepsize}{\invtemp}\sum_{k=0}^{\timescale -1}\sqrt{\EE\|\FD^2 f(\newy)-\FD^2 f(\newy^{k\noise})\|^2}\sqrt{\EE \rbra*{\noise_{k}^2-1}^2}\\
&\lesssim \frac{\spacescale^3 \stepsize^{3/2} \|g\|_M}{\invtemp^{3/2}}\sum_{k=0}^{\timescale -1}\sqrt{\EE \rbra*{\frac{1}{2}+\frac{1}{3}\|\newy^{k\noise}\|+\frac{1}{3}\EE\|\hat{Z}\|^2}^2}\\
&\lesssim \frac{D_2\timescale \spacescale^3 \stepsize^{3/2} \|g\|_M}{\invtemp^{3/2}},\label{eq:cov_xi_xi1}
\] 
where $D_2$ is given in \cref{D1D2}.
Then we can bound the second term with the results of \cref{2b-zz,eq:eps_cov_third,eq:step2_part1,eq:eps_cov_final}.
\[&\EE\frac{2\spacescale^2 \stepsize}{\invtemp}\sum_{k=0}^{\timescale -1}\FD^2 f(\newy) \Bigg[\sum_{m=k+1}^{\timescale}Q(k+1,m)\mathbb {I}_{[\frac{m}{\timescale
},\frac{m+1}{\timescale})}(t),\sum_{r=k+1}^{\timescale}Q(k+1,r)\mathbb{I}_{[\frac{r}{\timescale
},\frac{r+1}{\timescale})}(t)\Bigg]\\
&-\frac{\spacescale^2} {\invtemp \FI}\sum_{m,r=1}^{\timescale}\rbra*{e^{-h \FI  |m-r|}-e^{-h \FI  (m+r)}}\EE \FD^2 f(\newy)\sbra*{\mathbb{I}_{[\frac{m}{\alpha},\frac{m+1}{\alpha})},\mathbb{I}_{[\frac{r}{\alpha},\frac{r+1}{\alpha})}}\\
\lesssim& \frac{ D_2\timescale^{3/2} \spacescale^2 \stepsize^2 L \|g\|_M}{b^{1/2}\invtemp}\sqrt{1+\timescale\stepsize^2 L^2}+\frac{D_2 \timescale \spacescale^2 \stepsize^2  \FI  \|g\|_M}{\invtemp}\rbra*{\timescale\stepsize \FI +1}.\label{eq:cov_xi_xi2}
\]

Thus, summing up the bounds from \cref{eq:cov_xi_xi1,eq:cov_xi_xi2} we obtain
\[\label{eq:cov6_bound}
&\error{cov.\noise\noise-Z_\noise}\\
\lesssim &  \frac{D_2\timescale \spacescale^3 \stepsize^{3/2} \|g\|_M}{\invtemp^{3/2}}+\frac{ D_2\timescale^{3/2} \spacescale^2 \stepsize^2 L \|g\|_M}{b^{1/2}\invtemp}\sqrt{1+\timescale\stepsize^2 L^2}+\frac{D_2 \timescale \spacescale^2 \stepsize^2  \FI  \|g\|_M}{\invtemp}\rbra*{\timescale\stepsize \FI +1}.\]

\subsubsection{Final bound on $\error{cov}$}
The final bound on $\error{cov}$ in \cref{lemma_error_cov} follows from \cref{eq:quad_decompos,eq:cov1_bound,eq:cov2_bound,eq:cov3_bound,eq:cov4_bound,eq:cov5_bound,eq:cov6_bound}.
\section{Additional technical lemmas}
\begin{lemma}\label{D1D2}
Let $\newy^{k},\newy^{k\Psi},\newy^{k\noise} $ be given by \cref{58,newyk,newyknoise}. Then,
\[
&\sqrt{\EE\rbra*{1+\frac{2}{3}\|\newy\|^2 +\frac{4}{3}\EE\|\newz\|^2}^2}\lesssim D_1\qquad\text{and}\\
&\max\Bigg\{\sqrt{\EE\sbra*{\rbra*{1+\frac{1}{3}\|\newy\| +\frac{1}{3}\EE\|\newz\|^2}^2}}, \sqrt{\EE \rbra*{\frac{1}{2}+\frac{1}{3}\|\newy^{k\Psi}\|+\frac{1}{3}\EE\|\newz\|^2}^2}\\
&\qquad\sqrt{\EE\sbra*{\rbra*{\frac{1}{2}+\frac{1}{3}\|\newy^k\|+\frac{1}{3}\EE\|\newz\|^2}^2}},\sqrt{\EE \sbra*{\rbra*{\frac{1}{2}+\frac{1}{3}\|\newy^{k\noise}\|+\frac{1}{3}\EE\|\newz\|^2}^2}}\Bigg\}\lesssim D_2,
\]
where
\[
D_1:=&\Bigg\{1 +w^4\frac{\timescale^2 \stepsize^4}{\batchsize^2}\sbra*{\frac{\timescale\stepsize^2L^2}{b}E_1+\timescale^2\stepsize^4L^4E_1+\frac{\EE \Gradbatch{1}{1}^4}{b}+\Omega^2+\Omega}\\
&+w^4\frac{\timescale^2\stepsize^2}{\invtemp^2}\sbra*{1+\frac{\timescale\stepsize^2L^2}{\batchsize}+\timescale^2\stepsize^4L^4}+w^4\rbra*{\frac{1}{\batchsize}+\timescale \stepsize^2L^2}\sqrt{\Omega}(1+L^3) \frac{\timescale^3h^{11/2}L^2}{\invtemp^{3/2}}\\
&+ \frac{1}{\FI^2 }\rbra*{\frac{w^2 h}{b} \Omega  + \frac{ w^2 }{\invtemp}}^2\log^2(1+\timescale\stepsize \FI )\Bigg\}^{1/2}\\
D_2:=&\Bigg\{1+w^2\frac{\timescale \stepsize^2}{\batchsize}\sbra*{\sqrt{\Omega}L(1+L^3)\frac{\timescale^{1/2}\stepsize^{3/4}}{\invtemp^{3/4}}+L^2\Omega\timescale\stepsize^2+\Omega}+w^2\frac{\timescale\stepsize}{\invtemp}\sbra*{\frac{\timescale^{1/2}\stepsize L}{\batchsize^{1/2}}+1+L^2\timescale\stepsize^2}\\
&\quad+w^2\frac{\timescale^{3/2}\stepsize^3L}{\batchsize^{3/2}}\sqrt{E_1}+w^2 \rbra*{\frac{\Omega\stepsize^2} {b}+\frac{\stepsize}{\invtemp}}\rbra*{1+\frac{\timescale^2 h^2 L^2}{(1-L\stepsize)^2}}\\
&\quad  +\frac{1}{\FI ^2}\rbra*{\frac{w^2 h}{b} \Omega  + \frac{ w^2 }{\invtemp}}^2\log ^2(1+\timescale\stepsize \FI )\Bigg\}^{1/2}.
\]
and 
\[
E_1 &:= \EE \Gradbatch{1}{1}^4+\Omega^2+\sqrt{\Omega}(1+L^3)\sbra*{\rbra*{\EE \Gradbatch{1}{1}^4+\Omega^2}^{3/4}+1}+\Omega L(1+L^4)\sbra*{\Omega\frac{\timescale h^2}{b}+\frac{\timescale h}{\invtemp}}.
\]
\end{lemma}
\begin{proof}
From \cite[Thoerem 2.2]{jia2020moderate} we know that for some constant $C_z>0$ 
\[\label{eq:bound_new_z}
\EE\|\newz\|^2 \le \frac{C_z}{\timescale\stepsize \FI }\rbra*{\frac{w^2 \alpha h^2}{b} \Omega  + \frac{ w^2 \alpha h}{\invtemp}}\log(1+\timescale\stepsize \FI )=\frac{C_z}{\FI }\rbra*{\frac{w^2 h}{b} \Omega  + \frac{ w^2 }{\invtemp}}\log(1+\timescale\stepsize \FI ).\]

It follows from \cref{BDforSM} that
\[
\EE\|\newy\|^2& = w^2 \EE \max_{k\in\{0,\cdots,\timescale\}} \newpmt_k^2\\
&\lesssim w^2\frac{\timescale \stepsize^2}{\batchsize}\sbra*{\frac{E_1^{1/2}\timescale^{1/2}\stepsize L}{\batchsize^{1/2}}+\sqrt{\Omega}L(1+L^3)\frac{\timescale^{1/2}\stepsize^{3/4}}{\invtemp^{3/4}}+L^2\Omega\timescale\stepsize^2+\Omega}\\
&+w^2\frac{\timescale\stepsize}{\invtemp}\sbra*{\frac{\timescale^{1/2}\stepsize L}{\batchsize^{1/2}}+1+L^2\timescale\stepsize^2};\label{eq:bound_new_y_2}\\
\EE\|\newy\|^4& = w^4 \EE \max_{k\in\{0,\cdots,\timescale\}} \newpmt_k^4\\
& \hspace{-0.5cm}\lesssim w^4\frac{\timescale^2 \stepsize^4}{\batchsize^2}\sbra*{\frac{\timescale\stepsize^2L^2}{b}E_1+\timescale^2\stepsize^4L^4E_1+\frac{\EE \Gradbatch{1}{1}^4}{b}+\Omega^2+\Omega}\\
&+w^4\frac{\timescale^2\stepsize^2}{\invtemp^2}\sbra*{1+\frac{\timescale\stepsize^2L^2}{\batchsize}+\timescale^2\stepsize^4L^4}+w^4\rbra*{\frac{1}{\batchsize}+\timescale \stepsize^2L^2}\sqrt{\Omega}(1+L^3) \frac{\timescale^3h^{11/2}L^2}{\invtemp^{3/2}},\label{eq:bound_new_y_4}
\]
Moreover, given \cref{58,newyk,newyknoise}, since $Q(i,j)\le 1$, $\EE\abs{\frac{\frac{h}{b}\sum_{i=1}^{b}\Heisbatch{k}{i}}{1-\frac{h}{b}\sum_{i=1}^{b}\Heisbatch{k}{i}}}^2\le \frac{\stepsize^2 L^2}{(1-L\stepsize)^2}$, $\EE\rbra*{\frac{h}{b}\sum_{i=1}^{b}\Gradbatch{j}{i}+\sqrt{\frac{2h}{\invtemp}}\noise_{j}}^2=\frac{\Omega\stepsize^2} {b}+\frac{2\stepsize}{\invtemp} $,
\[\max\{\EE\|\newy^{k}\|^2,\EE\|\newy^{k\Psi}\|^2,\EE\|\newy^{k\noise}\|^2\}\le 2\EE\|\newy\|^2 + 4w^2\rbra*{\frac{\Omega\stepsize^2} {b}+\frac{2\stepsize}{\invtemp}}\rbra*{1+\frac{\timescale^2 h^2 L^2}{(1-L\stepsize)^2}}\label{eq:max_yk_bound}
\]

The following upper bound follows from \cref{eq:max_yk_bound,eq:bound_new_y_2,eq:bound_new_z}:
\[
&\max\Bigg\{\EE\sbra*{\rbra*{1+\frac{1}{3}\|\newy\| +\frac{1}{3}\EE\|\newz\|^2}^2}, \EE \sbra*{\rbra*{\frac{1}{2}+\frac{1}{3}\|\newy^{k\Psi}\|+\frac{1}{3}\EE\|\newz\|^2}^2},\\
&\qquad\EE\sbra*{\rbra*{\frac{1}{2}+\frac{1}{3}\|\newy^k\|+\frac{1}{3}\EE\|\newz\|^2}^2},\EE \sbra*{\rbra*{\frac{1}{2}+\frac{1}{3}\|\newy^{k\noise}\|+\frac{1}{3}\EE\|\newz\|^2}^2}\Bigg\}\\
\lesssim&1+\EE\|\newy\|^2 +w^2 \rbra*{\frac{\Omega\stepsize^2} {b}+\frac{\stepsize}{\invtemp}}\rbra*{1+\frac{\timescale^2 h^2 L^2}{(1-L\stepsize)^2}} +\frac{1}{\FI ^2}\rbra*{\frac{w^2 h}{b} \Omega  + \frac{ w^2 }{\invtemp}}^2\log ^2(1+\timescale\stepsize \FI )\\
\lesssim& 1+w^2\frac{\timescale \stepsize^2}{\batchsize}\sbra*{\sqrt{\Omega}L(1+L^3)\frac{\timescale^{1/2}\stepsize^{3/4}}{\invtemp^{3/4}}+L^2\Omega\timescale\stepsize^2+\Omega}+w^2\frac{\timescale\stepsize}{\invtemp}\sbra*{\frac{\timescale^{1/2}\stepsize L}{\batchsize^{1/2}}+1+L^2\timescale\stepsize^2}\\
&\quad+w^2\frac{\timescale^{3/2}\stepsize^3L}{\batchsize^{3/2}}\sqrt{E_1}+w^2 \rbra*{\frac{\Omega\stepsize^2} {b}+\frac{\stepsize}{\invtemp}}\rbra*{1+\frac{\timescale^2 h^2 L^2}{(1-L\stepsize)^2}}\\
&\quad  +\frac{1}{\FI ^2}\rbra*{\frac{w^2 h}{b} \Omega  + \frac{ w^2 }{\invtemp}}^2\log ^2(1+\timescale\stepsize \FI )=:D_2^2.
\]
Moreover, the following upper bound follows from \cref{eq:bound_new_y_4,eq:bound_new_z}:
\[&\EE\rbra*{1+\frac{2}{3}\|\newy\|^2 +\frac{4}{3}\EE\|\newz\|^2}^2\\
\leq &3+\frac{4}{3}\EE\|\newy\|^4+\frac{16}{3}\rbra*{\EE\|\newz\|^2}^2\\
\lesssim & 1 +w^4\frac{\timescale^2 \stepsize^4}{\batchsize^2}\sbra*{\frac{\timescale\stepsize^2L^2}{b}E_1+\timescale^2\stepsize^4L^4E_1+\frac{\EE \Gradbatch{1}{1}^4}{b}+\Omega^2+\Omega}\\
&+w^4\frac{\timescale^2\stepsize^2}{\invtemp^2}\sbra*{1+\frac{\timescale\stepsize^2L^2}{\batchsize}+\timescale^2\stepsize^4L^4}+w^4\rbra*{\frac{1}{\batchsize}+\timescale \stepsize^2L^2}\sqrt{\Omega}(1+L^3) \frac{\timescale^3h^{11/2}L^2}{\invtemp^{3/2}}\\
&+ \frac{1}{\FI^2 }\rbra*{\frac{w^2 h}{b} \Omega  + \frac{ w^2 }{\invtemp}}^2\log^2(1+\timescale\stepsize \FI )=:D_1^2.
\]

\end{proof}
\begin{lemma}
\label{BDforQ}
Recall the notation of \cref{eq:Q_def}.
For any $k \ge -1$, under \cref{asmp:curvature,asmp:stepsize}, 
\[
\EE \sup_{m\in \{k+1,\cdots,\timescale\}} \rbra*{Q(k+1,m)-\EE Q(k+1,m)}^2\lesssim \frac{\timescale h^2 L^2}{b} + \frac{\timescale^2 h^4 L^4}{b}.
\]
\end{lemma}
\begin{proof}
Let
\[
T_m \defas  Q(k+1,m)-s_m,\qquad\text{where}\qquad
s_m =\left(1-h\FI \right)^{m-k-1}.
\]

Then \[
T_m &= \left(T_{m-1}+s_{m-1}\right)\left(1-\frac{h}{b}\sum_{i=1}^{b}\Heisbatch{m-1}{i}\right) -s_{m-1}\left(1-h\FI \right)\\
&= T_{m-1}-h\left[ \frac{1}{b}\sum_{i=1}^{b}\Heisbatch{m-1}{i}T_{m-1}+s_{m-1}\left(\frac{1}{b}\sum_{i=1}^{b}\Heisbatch{m-1}{i}-\FI\right)\right]
.\]

So
\[
&\EE \sup_{m=k+1,\cdots,\timescale} T_m^2\\
=&\EE \sup_{m=k+1,\cdots,\timescale} \Bigg[ T^2_{m-1} -2hT_{m-1}\left(\frac{1}{b}\sum_{i=1}^{b}\Heisbatch{m-1}{i}T_{m-1}+s_{m-1}\left(\frac{1}{b}\sum_{i=1}^{b}\Heisbatch{m-1}{i}-\FI\right)\right) \\
&+h^2\left(\frac{1}{b}\sum_{i=1}^{b}\Heisbatch{m-1}{i}T_{m-1}+s_{m-1}\left(\frac{1}{b}\sum_{i=1}^{b}\Heisbatch{m-1}{i}-\FI\right)\right)^2\Bigg]\\
\le &\EE \sup_{m=k+1,\cdots,\timescale} \Bigg[ T^2_{m-1} +2hT_{m-1}\Bigg(\left(\FI-\frac{1}{b}\sum_{i=1}^{b}\Heisbatch{m-1}{i}\right)T_{m-1}\\
&\hspace{5cm}-s_{m-1}\left(\frac{1}{b}\sum_{i=1}^{b}\Heisbatch{m-1}{i}-\FI\right)\Bigg) \\
&+h^2\left(\frac{1}{b}\sum_{i=1}^{b}\Heisbatch{m-1}{i}T_{m-1}+s_{m-1}\left(\frac{1}{b}\sum_{i=1}^{b}\Heisbatch{m-1}{i}-\FI\right)\right)^2\Bigg]\\
\vdots \\
\le &\EE \sup_{m=k+1,\cdots,\timescale} \Bigg[ T^2_{k+1} +\sum_{j=k+1}^{m-1}2hT_{j}\left(\left(\FI-\frac{1}{b}\sum_{i=1}^{b}\Heisbatch{j}{i}\right)T_{j}-s_{j}\left(\frac{1}{b}\sum_{i=1}^{b}\Heisbatch{j}{i}-\FI\right)\right) \\
&+\sum_{j=k+1}^{m-1}h^2\left(\frac{1}{b}\sum_{i=1}^{b}\Heisbatch{j}{i}T_{j}+s_{j}\left(\frac{1}{b}\sum_{i=1}^{b}\Heisbatch{j}{i}-\FI\right)\right)^2\Bigg]\\
\le&   2h\EE \sbra*{\sup_{m=k+1,\cdots,\timescale}\sum_{j=k+1}^{m-1}T_{j}\left(\left(\FI-\frac{1}{b}\sum_{i=1}^{b}\Heisbatch{j}{i}\right)T_{j}-s_{j}\left(\frac{1}{b}\sum_{i=1}^{b}\Heisbatch{j}{i}-\FI\right)\right)} \\
&\qquad+h^2\EE \sbra*{\sum_{j=k+1}^{\timescale-1}\left(\frac{1}{b}\sum_{i=1}^{b}\Heisbatch{j}{i}T_{j}+s_{j}\left(\frac{1}{b}\sum_{i=1}^{b}\Heisbatch{j}{i}-\FI\right)\right)^2}+\EE T_{k+1}^2.\label{eq:t_m_bound1}
\]

Note that \[\Bigg\{ T_j\left(\left(\FI-\frac{1}{b}\sum_{i=1}^{b}\Heisbatch{j}{i}\right)T_{j}-s_{j}\left(\frac{1}{b}\sum_{i=1}^{b}\Heisbatch{j}{i}-\FI\right)\right) \Bigg\}_{j=0}^{\timescale-1}\] is a martingale difference sequence. Because $\left(\FI-\frac{1}{b}\sum_{i=1}^{b}\Heisbatch{j}{i}\right)$ are mean 0 and independent of $\cbra*{T_i}_{i=0}^{j}$.
So $m\mapsto\sum_{j=k+1}^{m-1}T_{j}\left(\left(\FI-\frac{1}{b}\sum_{i=1}^{b}\Heisbatch{j}{i}\right)T_{j}-s_{j}\left(\frac{1}{b}\sum_{i=1}^{b}\Heisbatch{j}{i}-\FI\right)\right)$ is a martingale.

Thus by Burkholder--Davis--Gundy inequality, using the fact that $|T_j+s_j|\le 1$
\[
&\EE \sup_{m=k+1,\cdots,\timescale-1}\left|\sum_{j=k+1}^{m-1}T_{j}\left(\left(\FI  -\frac{1}{b}\sum_{i=1}^{b}\Heisbatch{j}{i}\right)T_{j}-s_{j}\left(\frac{1}{b}\sum_{i=1}^{b}\Heisbatch{j}{i}-\FI\right)\right) \right|\\
&\lesssim  \EE \left\{\left(\sum_{j=k+1}^{\timescale-1}T_{j}^2 \rbra*{\FI -\frac{1}{b}\sum_{i=1}^{b}\Heisbatch{j}{i}}^2(T_j +s_j)^2\right)^\frac{1}{2}\right\}\\
&\lesssim  \left(\EE \left\{\sum_{i=0}^{\timescale-1}T_{j}^2 \rbra*{\FI -\frac{1}{b}\sum_{i=1}^{b}\Heisbatch{j}{i}}^2(T_j +s_j)^2\right\}\right)^\frac{1}{2}\\
&\lesssim \frac{ L}{b^{1/2}}\left(\sum_{i=0}^{\timescale-1}\EE T_{j}^2\right)^\frac{1}{2}.\label{eq:t_m_bound2}
\]

And since for $a,b,a>b, |a|\le 1,|b|\le 1, x\in\mathbb{N},$ we have $a^x-b^x \le x(a-b)$,
\[\EE T_j^2 &= \EE Q^2(k+1,j) -s_j^2\\
& = \rbra*{1-2h\FI +h^2\rbra*{\frac{\EE\sigma_{I(1,1)}^2-\FI ^2}{b}+\FI ^2}}^{j-k-1} -\rbra*{1-2h\FI +h^2 \FI ^2}^{j-k-1} \\
& \le \frac{(j-k-1) \stepsize^2}{b}\rbra*{\EE\sigma_{I(1,1)}^2-\FI ^2}\\
&\le \frac{\timescale \stepsize^2L^2}{b} \label{eq:t_m_bound3}
\]
Moreover,
\[
&\EE \sbra*{\sum_{j=k+1}^{\timescale-1}\left(\frac{1}{b}\sum_{i=1}^{b}\Heisbatch{j}{i}T_{j}+s_{j}\left(\frac{1}{b}\sum_{i=1}^{b}\Heisbatch{j}{i}-\FI\right)\right)^2}\\
&\le 2\EE\sum_{j=k+1}^{\timescale-1}\sbra*{\frac{\rbra*{\sum_{i=1}^{b}\Heisbatch{j}{i}}^2 T_j^2}{b^2}+s_j^2\left(\frac{1}{b}\sum_{i=1}^{b}\Heisbatch{j}{i}-\FI\right)^2}\\
&\le 2L^2 \sum_{j=0}^{\timescale-1} \EE T_j^2 +2\timescale \frac{L^2}{b}.\label{eq:t_m_bound4}
\]
Thus, it follows from \cref{eq:t_m_bound1,eq:t_m_bound2,eq:t_m_bound3,eq:t_m_bound4} that 
\[
\EE \sup_{m=k+1,\cdots,\timescale} T_m^2 \lesssim &h \frac{ L}{b^{1/2}} \frac{\timescale \stepsize L}{b^{1/2}}+ h^2 \rbra*{\frac{\timescale^2 h^2 L^4}{b}+\frac{\timescale L^2}{b}} + \frac{\timescale h^2 L^2}{b} = \frac{3\timescale h^2 L^2}{b} + \frac{\timescale^2 h^4 L^4}{b}.
\]
\end{proof}
\begin{lemma}\label{Mart}
For $m=1,\cdots,\timescale$, let $E_m \defas \sum_{k=0}^{m-1}Q(k+1,m)\rbra*{b\FI  -\sum_{i=1}^{b}\Heisbatch{k}{i}}\newpmt_k$, then under \cref{asmp:curvature,asmp:stepsize},
\[
\EE \sup_{m \in \{1,\cdots,\timescale\}} E_m^2 \lesssim bL^2\sum_{j=0}^{\timescale}\EE\newpmt_j^2+\timescale \stepsize^2 b L^4  \sum_{j=0}^{\timescale}\sum_{k=0}^{j-1}  \EE\newpmt_k^2.\]
\end{lemma}
\begin{proof}
Notice that 
\[E_{m+1} &= \rbra*{1-\frac{h}{b}\sum_{i=1}^{b}\Heisbatch{m}{i}} E_m +\rbra*{b\FI  -\sum_{i=1}^{b}\Heisbatch{m}{i}}\newpmt_m\\
&= E_m + \rbra*{b\FI  -\sum_{i=1}^{b}\Heisbatch{m}{i}}\newpmt_m - \rbra*{\frac{h}{b}\sum_{i=1}^{b}\Heisbatch{m}{i}}E_m \\
& \vdots \\
&= E_0 +\sum_{j=0}^{m}\rbra*{b\FI  -\sum_{i=1}^{b}\Heisbatch{j}{i}}\newpmt_j -\sum_{j=0}^{m}\rbra*{\rbra*{\frac{h}{b}\sum_{i=1}^{b}\Heisbatch{m}{i}}E_m}\\
&= \sum_{j=0}^{m}\rbra*{b\FI  -\sum_{i=1}^{b}\Heisbatch{j}{i}}\newpmt_j -\sum_{j=0}^{m}\rbra*{\frac{h}{b}\sum_{i=1}^{b}\Heisbatch{m}{i}}E_m,
\]
where $E_0:=0$. Thus 
\[
\EE \sup_{m \in \{1,\cdots,\timescale\}} E_m^2 \le& 2\EE \sup_{m \in \{1,\cdots,\timescale\}}  \cbra*{\sum_{j=0}^{m}\rbra*{b\FI  -\sum_{i=1}^{b}\Heisbatch{j}{i}}\newpmt_j}^2\\
&+ 2\EE \sup_{m \in \{1,\cdots,\timescale\}} \cbra*{\sum_{j=0}^{m}\rbra*{\frac{h}{b}\sum_{i=1}^{b}\Heisbatch{j}{i}}E_j}^2.
\]

For the first term, notice that $\rbra*{b\FI  -\sum_{i=1}^{b}\Heisbatch{j}{i}}\newpmt_j$ is a martingale difference sequence. So $\sum_{j=0}^{m}\rbra*{b\FI  -\sum_{i=1}^{b}\Heisbatch{j}{i}}\newpmt_j$ is a martingale. Using Doob's inequality,
\[
&\EE \sup_{m \in \{1,\cdots,\timescale\}}  \cbra*{\sum_{j=0}^{m}\rbra*{b\FI  -\sum_{i=1}^{b}\Heisbatch{j}{i}}\newpmt_j}^2\\
&\le 4\EE \cbra*{\sbra*{\sum_{j=0}^{\timescale}\rbra*{b\FI  -\sum_{i=1}^{b}\Heisbatch{j}{i}}\newpmt_j}^2}\\
&= 4\EE\cbra*{ \sum_{j=0}^{\timescale}\rbra*{b\FI  -\sum_{i=1}^{b}\Heisbatch{j}{i}}^2\newpmt_j^2}\\
& = 4b\rbra*{\EE\rbra*{\Heisbatch{1}{1}-\FI} ^2}\sum_{j=0}^{\timescale}\EE\newpmt_j^2
\le 4bL^2\sum_{j=0}^{\timescale}\EE\newpmt_j^2.
\]

For the second term, since 
\[\EE E_j^2 &= \EE \sbra*{\rbra*{\sum_{k=0}^{j-1}Q(k+1,j)\rbra*{b\FI  -\sum_{i=1}^{b}\Heisbatch{k}{i}}\newpmt_k }^2}\\
&=\EE \sum_{k=0}^{j-1}Q(k+1,j)^2\rbra*{b\FI  -\sum_{i=1}^{b}\Heisbatch{k}{i}}^2\newpmt_k^2 \\
&\le   b\rbra*{\EE\rbra*{\Heisbatch{j}{i}-\FI} ^2}\sum_{k=0}^{j-1}\EE\newpmt_k^2\\
&\le   bL^2\sum_{k=0}^{j-1}\EE\newpmt_k^2,\]
we have

\[
&\EE \sup_{m \in \{1,\cdots,\timescale\}} \cbra*{\sum_{j=0}^{m}\rbra*{\frac{h}{b}\sum_{i=1}^{b}\Heisbatch{j}{i}}E_j}^2\\
&\le \EE \sup_{m \in \{1,\cdots,\timescale\}} \frac{\timescale\stepsize^2}{b^2} \cbra*{\sum_{j=0}^{m}\rbra*{\sum_{i=1}^{b}\Heisbatch{j}{i}}^2 E_j^2}\\
& \le \EE \frac{\timescale\stepsize^2}{b^2}\cbra*{\sum_{j=0}^{\timescale}\rbra*{\sum_{i=1}^{b}\Heisbatch{j}{i}}^2 E_j^2}\\
& \le \timescale \stepsize^2 L^2 \sum_{j=0}^{\timescale}\EE E_j^2\\
& \le  \timescale \stepsize^2 b L^4  \sum_{j=0}^{\timescale}\sum_{k=0}^{j-1}  \EE\newpmt_k^2.
\]
\end{proof}

\bibliographystyle{imsart-nameyear} %
\bibliography{biblio}

\end{document}